\documentclass[twocolumn]{svjour3}
\usepackage{natbib}

\usepackage{booktabs}

\usepackage{amsmath,graphicx,amssymb,url,amsfonts}
\usepackage{algpseudocode}

\usepackage{multirow}
\usepackage{mdframed}
\usepackage{url}
\usepackage{enumitem}
\usepackage[makeroom]{cancel}
\usepackage{dblfloatfix}
\usepackage{array}
\usepackage{epstopdf}
\usepackage{float}
\usepackage{subfig}
\usepackage{breqn}
\usepackage{xcolor}
\usepackage{tabularx}
\usepackage[printonlyused,withpage]{acronym}
\usepackage{balance}
\usepackage{math}
\usepackage[lined,boxed,commentsnumbered,ruled,norelsize]{algorithm2e}

\newcommand{\argmin}{\operatornamewithlimits{argmin}}

\usepackage[pagebackref=true,breaklinks=true,letterpaper=true,colorlinks,bookmarks=false]{hyperref}
  \usepackage{diagbox}
\acrodef{ZNCC}{zero-mean normalized cross correlation}
\acrodef{ECM}{ex\-pec\-ta\-tion con\-di\-ti\-onal maximization}
\acrodef{STOI}{short-time objective intelligibility}
\acrodef{SE}{speech enhancement}
\acrodef{STFT}{short-time Fourier transform}
\acrodef{PSD}{power spectral density}
\acrodef{NMF}{nonnegative matrix factorization}
\acrodef{AV}{audio-visual}
\acrodef{DNN}{deep neural network}
\acrodefplural{DNN}[DNNs]{deep neural networks}
\acrodef{LDS}{linear dynamical system}
\acrodefplural{LDS}[LDSs]{linear dynamical systems}
\acrodef{DL-LDS}{\textit{doubly-latent} LDS}
\acrodef{VAE}{variational auto-encoder}
\acrodefplural{VAE}[VAEs]{variational auto-encoders}
\acrodef{CVAE}{conditional variational auto-encoder}
\acrodefplural{CVAE}[CVAEs]{conditional variational auto-encoders}
\acrodef{A-VAE}{audio VAE}
\acrodef{V-VAE}{visual VAE}
\acrodef{AV-CVAE}{audio-visual CVAE}
\acrodef{ROI}{region of interest}
\acrodef{MCMC}{Markov Chain Monte Carlo}
\acrodef{EM}{expectation-maximization}
\acrodef{VEM}{variational expectation-maximization}
\acrodef{MCEM}{Monte Carlo expectation-maximization}
\acrodef{TF}{time frequency}
\acrodef{ELBO}{evidence lower bound}
\acrodef{ROI}{region of interest}
\acrodef{LR}{Living Room}
\acrodef{SDR}{signal-to-distortion ratio}
\acrodef{PESQ}{perceptual evaluation of speech quality}
\acrodef{ASR}{automatic speech recognition}
\acrodef{ASE}{audio speech enhancement}
\acrodef{VSE}{visual speech enhancement}
\acrodef{AVSE}{audio-visual speech enhancement}
\acrodef{SNR}{signal-to-noise ratio}
\acrodefplural{SNRs}{signal-to-noise ratios}
\acrodef{LSTM}{long short-term memory}
\acrodef{HMM}{hidden Markov model}
\acrodef{SwVAE}{switching variational auto-encoder}
\acrodef{GAN}{generative adversarial network}
\acrodefplural{GAN}[GANs]{generative adversarial networks}
\acrodef{pdf}{probability distribution function}
\acrodef{3DMM}{3D morphable model}
\acrodef{FF}{face frontalization}
\acrodef{DFF}{dynamic face frontalization}
\acrodef{3DFA}{3D face alignment}
\acrodef{IWR}{isolated word recognition}
\usepackage{hyphenat}
\title{Expression-preserving face frontalization improves visually  assisted speech processing\thanks{This work has been partially supported by the H2020 SPRING project \#871245 and by the Multidisciplinary Institute of Artificial Intelligence (MIAI) ANR-19-P3IA-0003.}}
\author{Zhiqi Kang \and Mostafa Sadeghi \and Radu Horaud  
\and Xavier Alameda-Pineda}
\authorrunning{Z. Kang, M. Sadeghi, R. Horaud, 
and X. Alameda-PIneda}
\institute{
Z. Kang, R. Horaud, X. Alameda-Pineda \at
Inria Grenoble \& Universit\'e Grenoble Alpes, France
\and
M. Sadeghi \at
Inria Nancy Grand-Est\\
Villers-l\`es-Nancy, France
}

\usepackage[final]{review}
\begin{document}

\maketitle

\begin{abstract}
Face frontalization consists of synthesizing a frontal view from a profile one. This paper proposes a frontalization method that preserves non-rigid facial deformations, i.e. facial expressions. It is shown that expression-preserving frontalization boosts the performance of visually assisted speech processing. The method alternates between the estimation of (i)~the rigid transformation (scale, rotation, and translation) and (ii)~the non-rigid deformation between an arbitrarily-viewed face and a face model. The method has two important merits: it can deal with non-Gaussian errors in the data and it incorporates a dynamical face deformation model. For that purpose, we use the Student's t-distribution in combination with a Bayesian filter in order to account for both rigid head motions and time-varying facial deformations, e.g. caused by speech production. 
The zero-mean normalized cross-correlation (ZNCC) score is used to evaluate the ability of the method to preserve facial expressions. 
The method is thoroughly evaluated and compared with several state of the art methods, either based on traditional geometric models or on deep learning. Moreover, we show that the method, when incorporated into speech processing pipelines, improves word recognition rates and speech intelligibility scores by a considerable margin.\footnote{Supplemental material is accessible at \url{https://team.inria.fr/robotlearn/research/facefrontalization}.}
\end{abstract}




\section{Introduction}
\label{sec:introduction}
Face frontalization is the problem of synthesizing a fron\-tal view of a face from an arbitrarily viewed one. \addnote[ff-expression]{2} {Recent research has shown that face frontalization consistently boosts the performance of face recognition, e.g. \cite{yim2015rotating,Zhu_2015_CVPR,banerjee2018frontalize,zhao2018towards,zhou2018gridface,zhou2020rotate}. 
It is worth noticing that face recognition requires \textit{expression-free} face frontalization (which is also referred to as face normalization). In contrast, other applications, such as facial expression recognition, e.g. \cite{pei2020monocular} and visual speech processing, e.g. \cite{fernandez2018survey,adeel2019lip,martinez2020lipreading,cheng2020towards}, require \textit{expression-preserving} face frontalization. In this paper we present a novel face frontalization methodology that combines robust statistical inference with a dynamic model. We show that the proposed algorithms improve the performance of visual speech by a considerable margin.}

It has long been established that visual perception plays a primordial role in speech communication. In particular, vision provides an alternative representation of some of the information that is present in the audio, with the advantage that it is affected neither by acoustic noise nor by competing audio sources. The most prominent visual features used in human-to-human, human-to-computer and human-to-robot interactions are facial movements. Facial movements are a combination of rigid head movements and non-rigid facial deformations. On one side, head movements play linguistic functions as they mark the structure of the ongoing discourse and are used to regulate interaction \cite{mcclave2000linguistic}. On the other side, lip and jaw movements are generated by facial muscles which, in turn, are controlled by speech production -- they are correlated with phonemes and with word pronunciation \cite{schultz2017biosignal}. Hence visual information plays a fundamental function both in speech recognition and in speech intelligibility. 

In particular, \ac{ASR} and \ac{SE} play crucial  yet complimentary roles in speech communication systems. \ac{SE} aims to improve the quality of noisy speech signals to be used by \ac{ASR}. It is well established that \ac{ASE} is severely limited in adverse acoustic situations, e.g. background noise. Multimodal speech enhancement, and in particular \ac{AVSE} aims at incorporating the complimentary information available with visual information. Lip reading plays a similar role in \ac{ASR}. 

\ac{AVSE} has received a lot of attention in the recent past, mainly because of the advent of \acp{DNN} which have considerably boosted their performance \cite{michelsanti2021overview}. Nevertheless, the vast majority of existing methods assume clean visual information -- they take as input lip regions that are cropped from frontal and steady face images, 
\cite{hou2018audio,sadeghi2020audio,adeel2021lip}.  Currently there are no \ac{DNN} architectures able to mitigate the effect of rigid head motions that are inherently present in speech communication. Moreover, the vast majority of existing datasets for training and testing \ac{AVSE} are recorded in constrained conditions -- the participants were instructed to avoid head movements and to face the camera \cite{Abde17,anina2015ouluvs2}. As for lip reading \cite{fernandez2018survey}, although there were some attempts to deal with in the wild datasets, the current state of the art is limited to the task of \ac{IWR} \cite{chung2016lip,ma2021towards}.
Not surprisingly, the performance of existing methods rapidly degrades in the presence of noisy visual information. Therefore, although these methods have profited from state of the art deep-learning models, they are ineffective in realistic conversational scenarios. 

\begin{figure}[t]
\begin{tabular}{cccccc}
\includegraphics[trim = 21cm 3cm 19cm 7cm,clip,keepaspectratio=true,width=0.15\columnwidth]{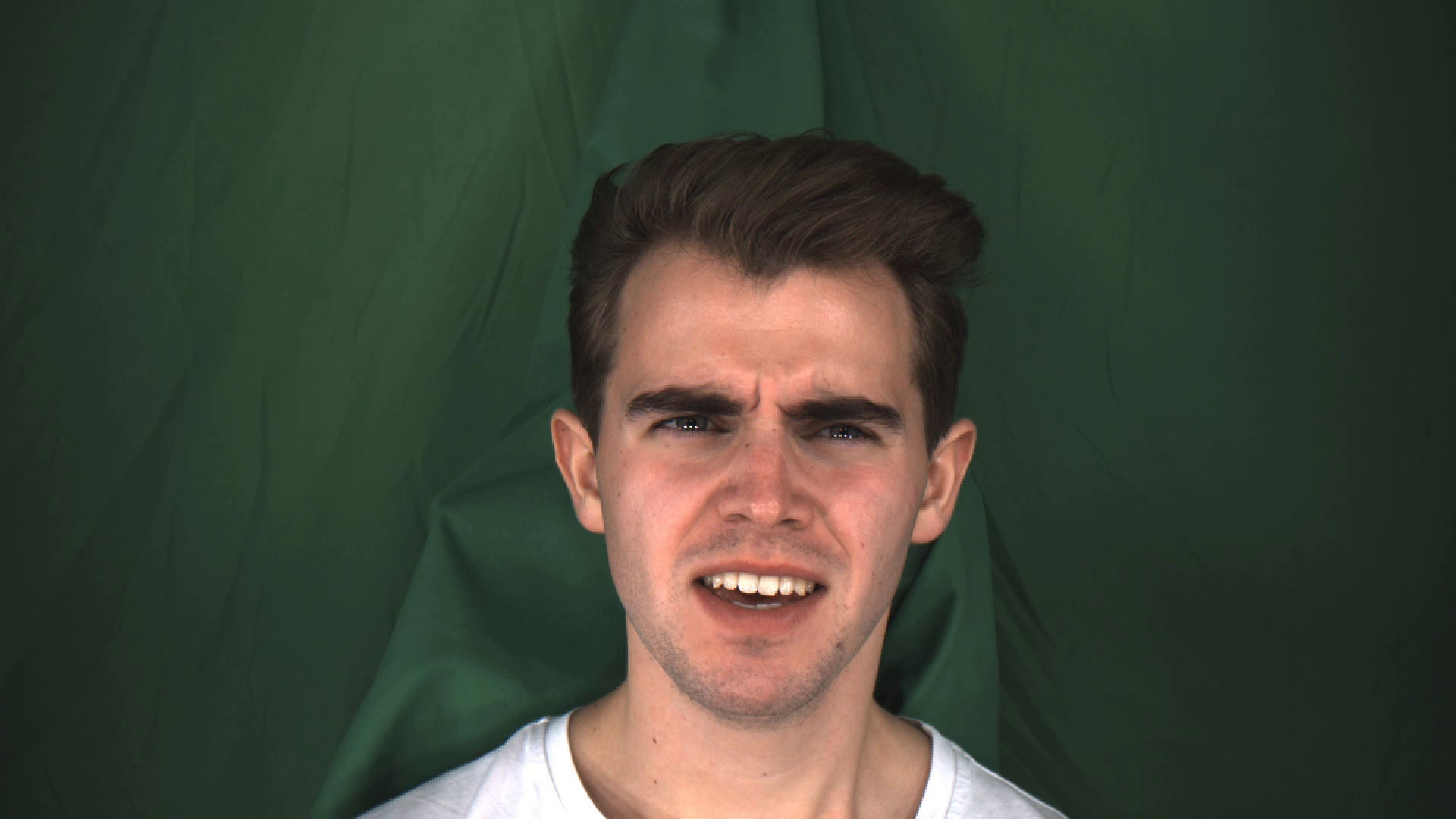}
&\includegraphics[trim = 21cm 3cm 19cm 7cm,clip,keepaspectratio=true,width=0.15\columnwidth]{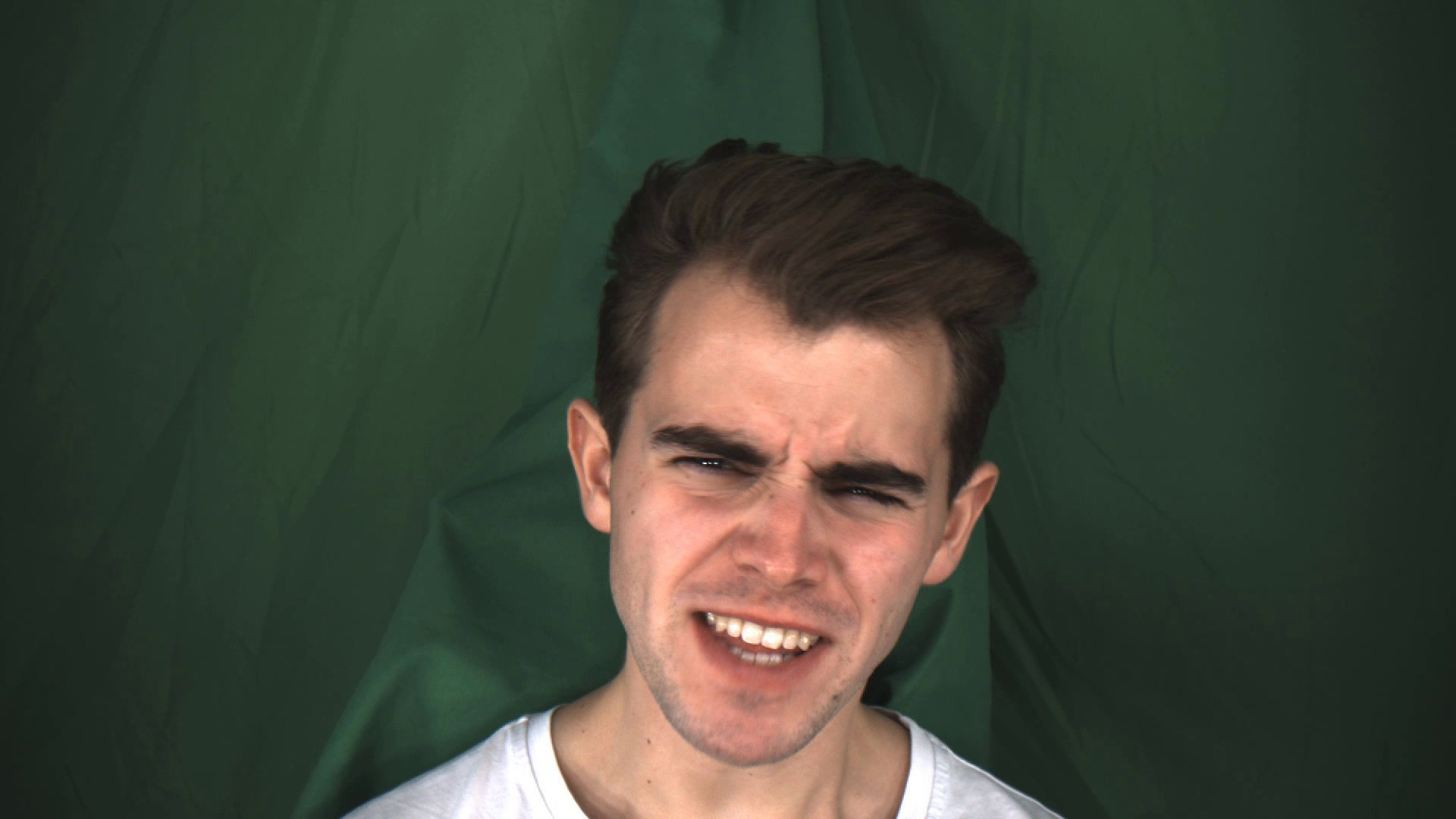}
&\includegraphics[trim = 21cm 3cm 19cm 7cm,clip,keepaspectratio=true,width=0.15\columnwidth]{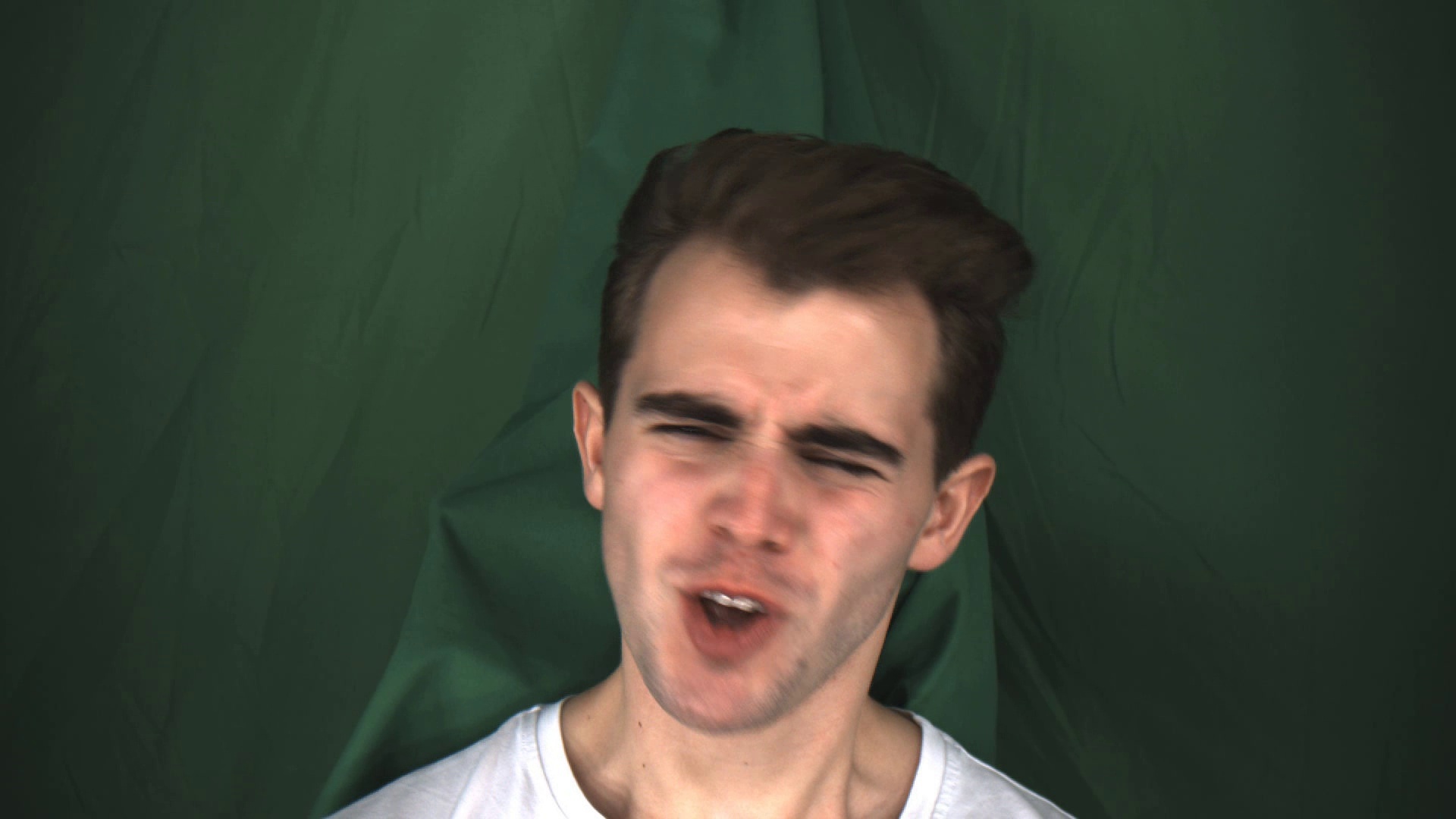}
&\includegraphics[trim = 21cm 3cm 19cm 7cm,clip,keepaspectratio=true,width=0.15\columnwidth]{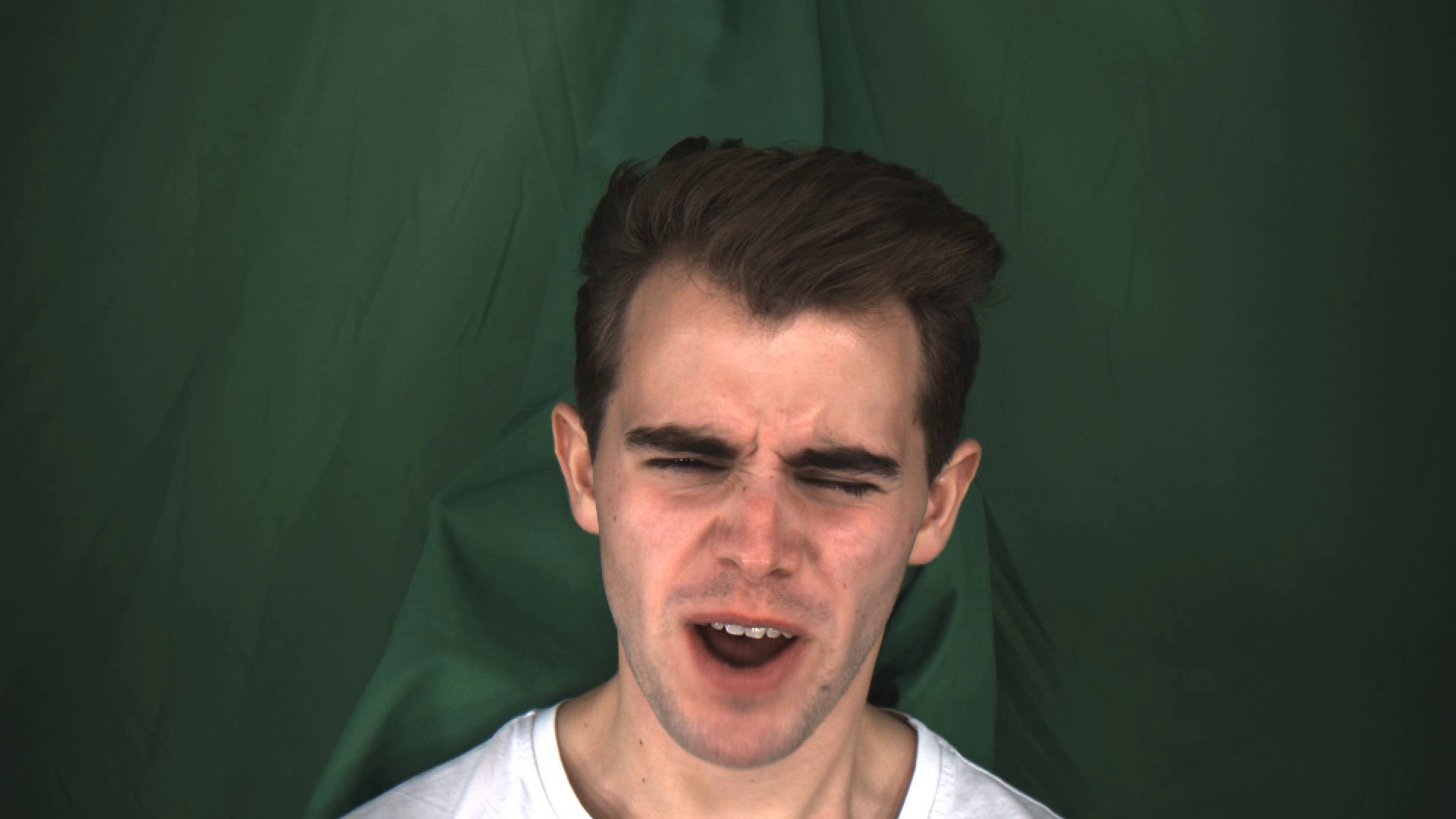}
&\includegraphics[trim = 21cm 3cm 19cm 7cm,clip,keepaspectratio=true,width=0.15\columnwidth]{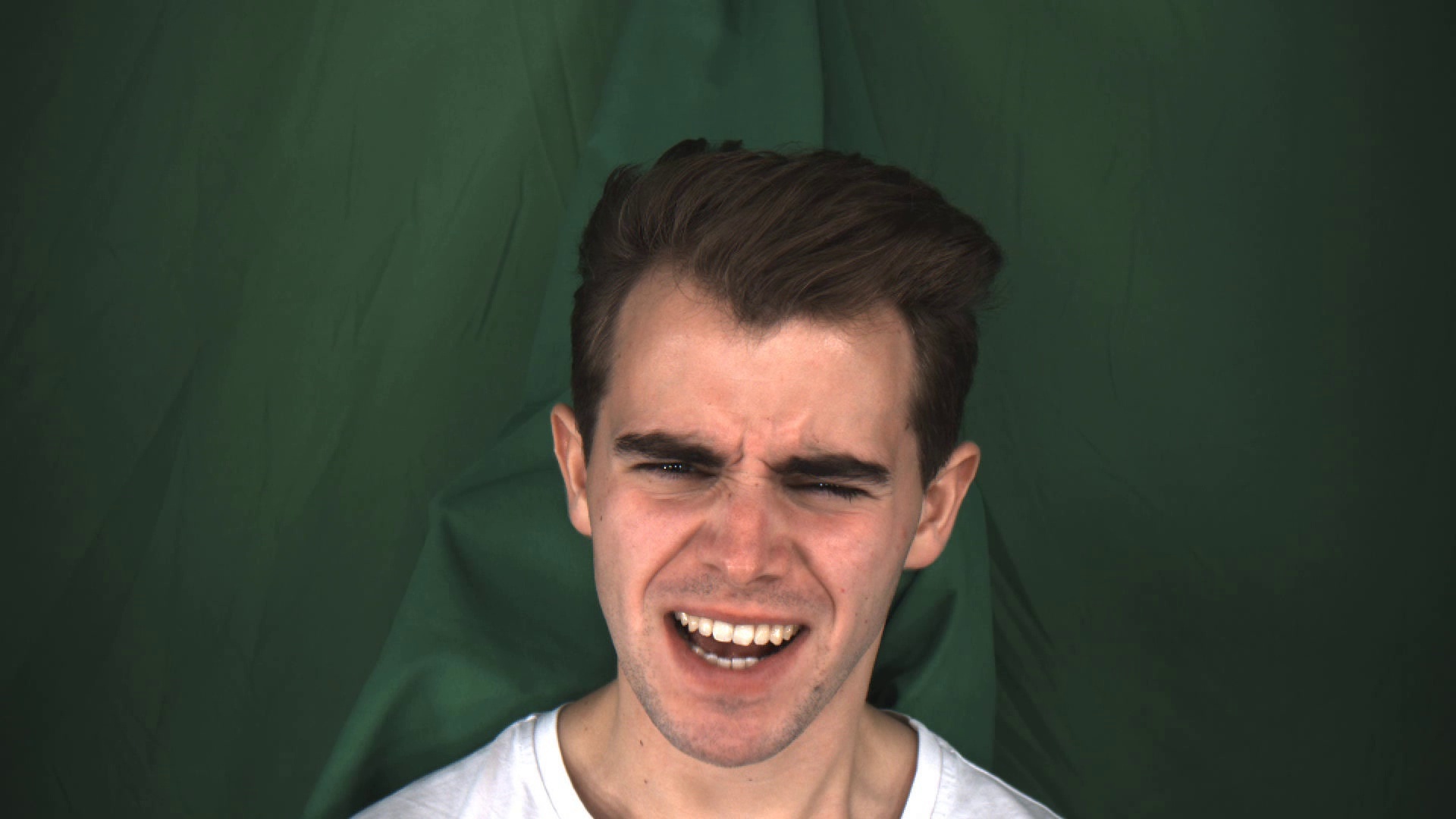}
&\includegraphics[trim = 21cm 3cm 19cm 7cm,clip,keepaspectratio=true,width=0.15\columnwidth]{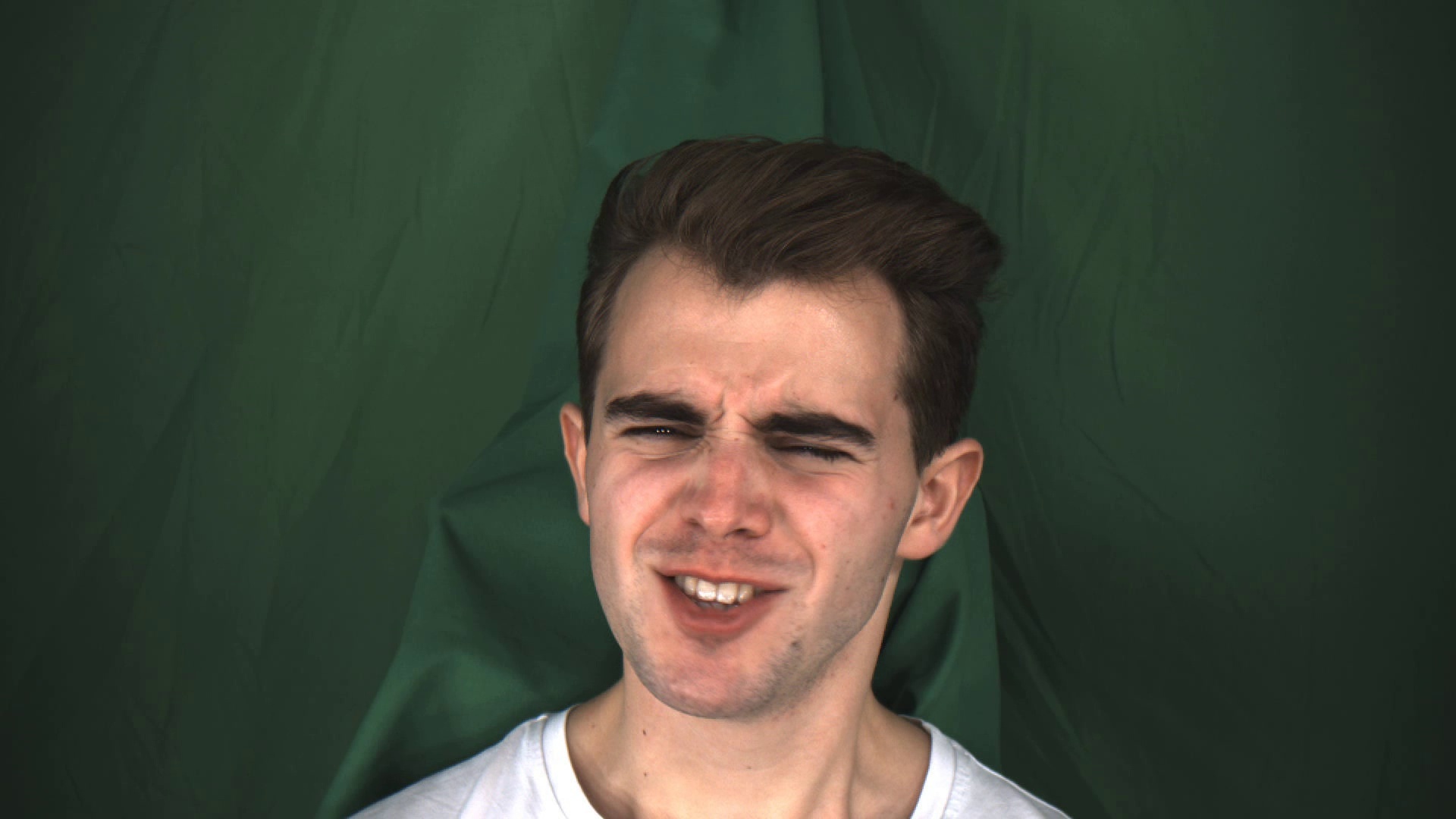}\\
\includegraphics[width=0.15\columnwidth]{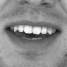}
&\includegraphics[width=0.15\columnwidth]{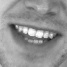}
&\includegraphics[width=0.15\columnwidth]{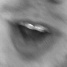}
&\includegraphics[width=0.15\columnwidth]{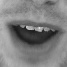}
&\includegraphics[width=0.15\columnwidth]{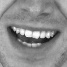}
&\includegraphics[width=0.15\columnwidth]{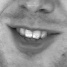}\\
\includegraphics[width=0.15\columnwidth]{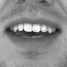}
&\includegraphics[width=0.15\columnwidth]{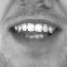}
&\includegraphics[width=0.15\columnwidth]{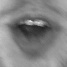}
&\includegraphics[width=0.15\columnwidth]{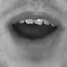}
&\includegraphics[width=0.15\columnwidth]{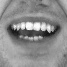}
&\includegraphics[width=0.15\columnwidth]{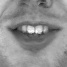}
\end{tabular}
\caption{\label{fig:lips} An example of applying expression-preserving face frontalization to a person that utters speech. Top: input images; Middle: lip regions before removing head movements; Bottom: lip regions after removing head movements.}
\end{figure}

\begin{figure}[t]
\centering
\subfloat[Vertical lip motion]{\includegraphics[keepaspectratio=true,width=0.49\columnwidth]{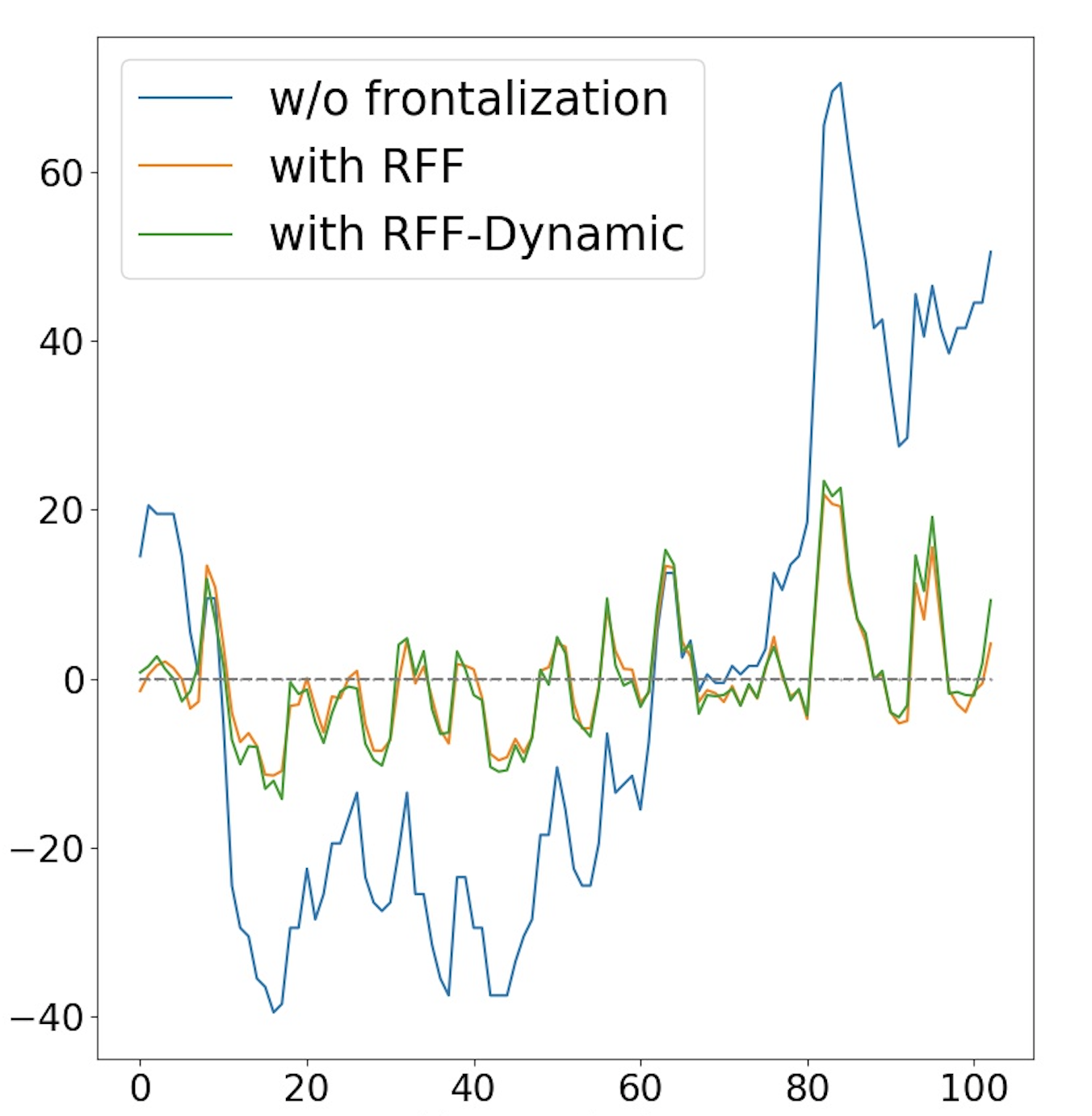}}
\subfloat[Horizontal lip motion]{\includegraphics[keepaspectratio=true,width=0.48\columnwidth]{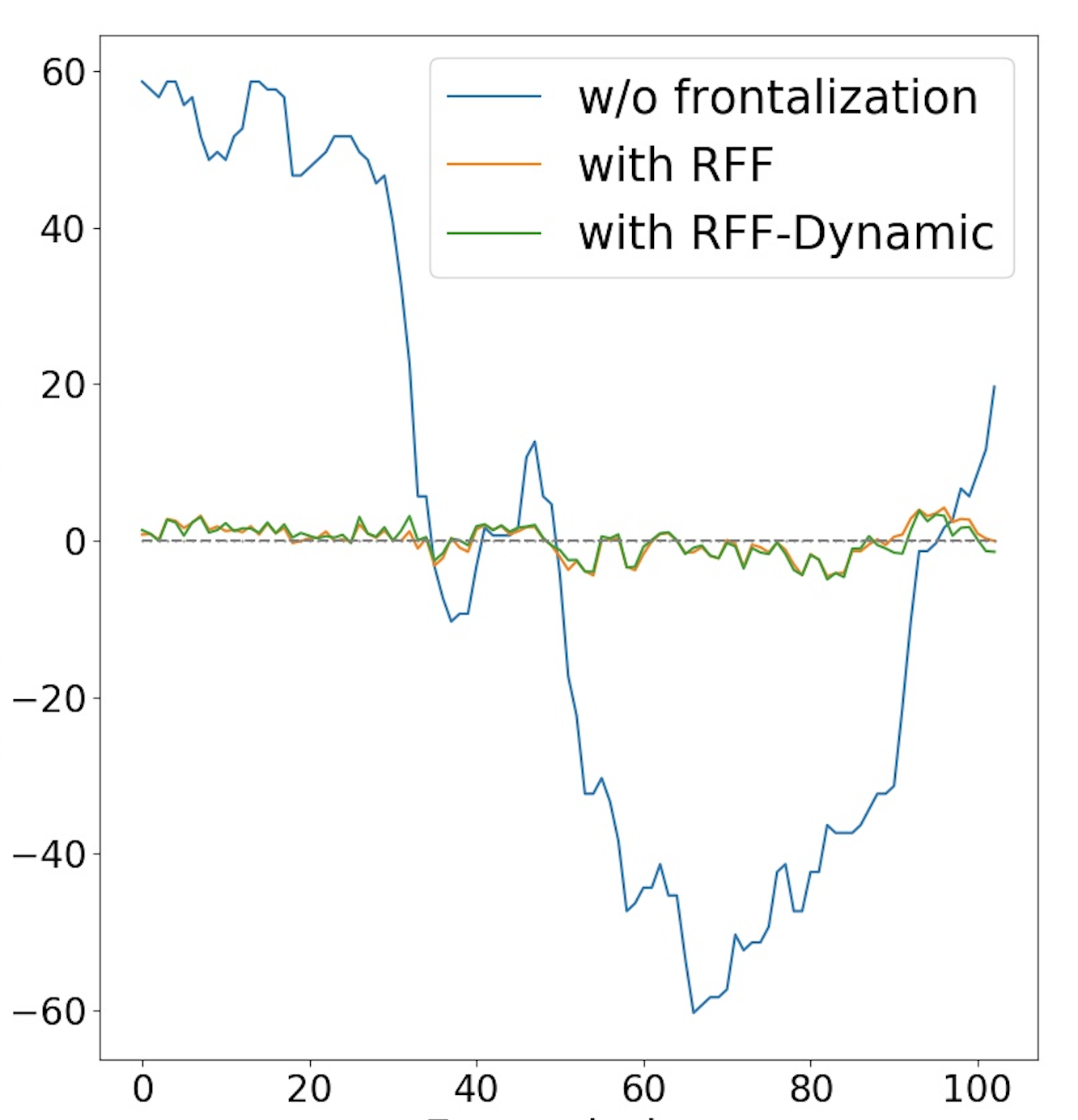}}
\caption{\label{fig:displacement} Lip motion without frontalization (blue), with robust frontalization (orange) \cite{kang2021robust}, and with robust-dynamic frontalization (green), proposed in this paper. The curves correspond to the example of Figure~\ref{fig:lips}. The curves correspond to image-plane displacements in pixels as a function of the frame number and they show the motion of the landmark located at the center of the upper lip, for the example shown on Figure~\ref{fig:lips}.  }
\end{figure}

In this paper we are interested in investigating vision-assisted speech processing methods that are robust with respect to noisy lip movements caused by head motions, e.g. Figure~\ref{fig:lips} and Figure~\ref{fig:displacement}. We propose to incorporate \ac{FF} into visual and audio-visual speech processing \ac{DNN} pipelines. 
Visual speech processing necessitates a \ac{FF} method that guarantees that non-rigid facial deformations are preserved. Moreover and unlike \ac{FF} for face recognition from a single image, \ac{FF} for visual speech analysis must incorporate a dynamic model in order to capture the temporal nature of lip movements.
We address these challenging problems on the following grounds: (i)~the image warping needed by \ac{FF} should be guided by a rigid transformation, (ii)~the estimation of this transformation should be robust with respect to non-rigid deformations, and (iii)~a dynamic face deformation model is needed in order to characterize the temporal behaviour of lip and jaw movements associated with speech. 

\addnote[main-contribution]{1}{
The rationale of the proposed method is to decouple rigid head pose from non-rigid facial deformations. These two pieces of information are encoded in the observed 3D landmarks while they are affected by two types of errors: small detection errors that are indistinguishable from non-rigid deformations, and large localization errors that might strongly biais the results. 
For these reasons, the estimation problem at hand is cast into the problem of robust statistical inference. Head pose and non-rigid deformation are coupled with the \textit{generalized Student's t-dis\-tri\-bution} \cite{forbes2014new} -- a heavy tailed \ac{pdf} that is able to deal both with Gaussian inliers and with non-Gaussian outliers. The associated \ac{ECM} procedure alternates between (i)~the evaluation of the posterior distributions of weights associated with observed facial landmarks, (ii)~the estimation of  the rigid head-pose parameters (scale, rotation and translation) and (iii)~the estimation of the parameters of a deformable face model, e.g. \cite{blanz1999morphable}, i.e. Figure~\ref{fig:graphicalmodel}-(a). The landmark weights just mentioned have a ponderable role: the higher is the weight, the more reliable is the landmark.
}

\addnote[tmm]{1}{
In the past the Student's t-mixture model (TMM) \cite{peel2000robust} was used for the task of robust non-rigid registration of multiple point sets \cite{zhou2014robust,ravikumar2018group}. These methods jointly register the points and estimate the rigid transformations that allow to optimally align the sets, on the premise that a majority of points in the sets are in rigid correspondence. In the case of landmark-based \ac{FF} both rigid and non-rigid alignment are needed while it is not necessary to perform registration, hence a single \ac{pdf}, and not a mixture, is sufficient. 
}


We also propose a dynamical extension. The frontalized landmarks are treated as observations of  a \ac{LDS}. Unlike a standard \ac{LDS}, the proposed one is equipped with two sequences of latent variables governed by two interconnected linear-Gaussian dynamical regimes, i.e. Figure~\ref{fig:graphicalmodel}-(b). The two latent variables correspond to the 3D vertex coordinates and to the low-dimensional face embedding of a \ac{3DMM}, respectively. The 3D vertices at the current frame are stochastically generated from the 3D vertices at the previous frame. Similarly, the current shape embedding is stochastically generated from the previous embedding. At each frame, the vertices are reconstructed from the shape embedding. In turn, the vertices stochastically generate the frontalized landmarks. We provide a formal derivation of the proposed \textit{doubly latent} \ac{LDS} and we show that it can be reformulated as a standard Kalman filter, with its associated recursive solver.


We empirically evaluate the performance of \ac{FF} using the \ac{ZNCC} score, \cite{sun2002fast}, between a frontalized face and its ground-truth frontal counterpart. We embed \ac{FF} into a state of the art deep lip reading model \cite{ma2021towards}. We also use \ac{FF} in combination with a recently proposed deep \ac{AVSE} model \cite{sadeghi2020audio,sadeghi2021mixture}. We use three different datasets associated with these three sets of experiments and we compare our method with two traditional frontalization methods \cite{hassner2015effective}, \cite{banerjee2018frontalize}, and with two methods based on \acp{GAN}, \cite{zhou2020rotate}, \cite{yin2020dual}. We show that the proposed expression-preserving face frontalization method outperforms all the other methods -- either based on traditional computer vision or based on \acp{DNN} -- by a considerable margin. 
\addnote[gan-better]{1}{
A prominent result is that robust estimation of the rigid transformation underlying \ac{FF} outperforms \ac{GAN}-based frontalization. Indeed, the latter estimates millions of parameters of a non-linear image-to-image mapping, which cannot guarantee that the non-rigid facial deformations, e.g. expressions and lip movements, are preserved.}

The remainder of this article is organized as follows. Section~\ref{sec:related-work} summarizes the related work. Section~\ref{sec:face-frontalization} describes in detail the proposed expression-preserving frontalization framework. Section~\ref{sec:method-implementation-details} provides algorithm implementation details. Section~\ref{sec:ff-benchmark} describes a benchmark based on the \ac{ZNCC} score. Section~\ref{sec:lip-reading} describes experiments with a lip-reading dataset. Section~\ref{sec:avse} combines face frontalization with audio-visual speech enhancement. Finally Section
\ref{sec:conclusions} draws some conclusions.

\section{Related Work}
\label{sec:related-work}

As already mentioned, face frontalization consists of synthesizing a frontally-viewed face from an arbitrarily-viewed face. 
Recently, a successful approach has been to train \acp{DNN} in order to learn a non-linear 2D-to-2D mapping between an arbitrary view and a frontal view.
Some of the best performing \ac{DNN}-based frontalization methods use CNN-GAN architectures, e.g.  \cite{yin2017towards,huang2017beyond,tran2017disentangled,zhao2018towards,zhang2019face,rong2020feature,zhang2021semi,yin2020dual}, which outperform CNN-only models, e.g. \cite{yim2015rotating}. 
These methods necessitate large collections of input/output pairs of face images. For that purpose  \cite{zhang2019face,yin2020dual,zhang2021semi} use two datasets that contain multiple-camera recordings in a controlled setup, i.e. \cite{gao2007cas,gross2010multi}. 
\cite{zhang2019face} proposed to learn dense pixel-to-pixel correspondences between the input-output faces. Subsequently, \cite{zhang2021semi} proposed a semi-supervised GAN-based method that augments the paired face images of \cite{gross2010multi} with unpaired in-the-wild faces with large variations in identity, e.g. \cite{huang2008labeled}; their adversarial and identity-preserving losses enhance face recognition performance.  \cite{yin2020dual} proposed a dual-attention GAN architecture that captures long-term dependencies in image space, thus providing a mean to preserve identity. These DNN-based methods are designed to predict as-neutral-as-possible frontal faces, i.e. expression-free faces, in order to improve the performance of face recognition. On the one side, the profile/frontal pairs of  \cite{gao2007cas,gross2010multi} are collected in controlled settings in terms of illumination and expression.  On the other side, the non-frontal images from in the wild datasets do not have their frontal counterparts to allow frontalization training. 

Another way to estimate the non-linear 2D-to-2D mapping between a profile image and a canonical image of a face is to use a rectification network that learns local homographies between a deformed grid, that corresponds to a profile view, and a regular grid, that supposedly corresponds to a frontal view \cite{zhou2018gridface}. While this method is well suited for improving the performance of face recognition, it is unable to take into consideration off-the-image-plane rotations, to guarantee a frontal image and to separate rigid head pose from non-rigid facial deformations.

Other methods estimate the pose of an input face with respect to a frontal 3D face model, then use the pose parameters to warp the facial pixels from the input image onto a frontal one. These methods capitalize on pose estimation from 2D-to-3D point correspondences, e.g. \cite{Zhu_2015_CVPR,hassner2015effective,ferrari2016effective,banerjee2018frontalize}. 
\addnote[hassner-problem]{2}{
In \cite{hassner2015effective} it was proposed to use a 3D generic model of a face from which a frontal face is generated: 48 facial landmarks (2D) are extracted from the input face and from the neutral and frontal face model (3D), thus providing 2D-to-3D correspondences between the input face and the generic 3D model. This amounts to estimate the intrinsic camera parameters as well as the rigid pose. Similar methods were proposed by \cite{Zhu_2015_CVPR}, \cite{ferrari2016effective} and \cite{banerjee2018frontalize}. Note that with this setup there is an inherent large discrepancy between the expressive input face and the neutral model face. Hence, these methods lack a built-in robust statistical model that enables accurate inference in the presence of large errors in landmark localization and of non-rigid facial deformations. To mitigate this issue,  \cite{hassner2015effective} manually removes jaw landmarks and \cite{Zhu_2015_CVPR} purposely removes expressions in order to favour identity features. }

Recently, \cite{zhou2020rotate} proposed to synthesize profile views from a collection of frontal views in order to create input/target pairs for the purpose of training image-to-image translation \acp{GAN}. Their method starts by fitting a 3D face model to a frontal view, using the 2D-to-3D alignment technique of \cite{zhu2019face}, followed by rotating and rendering the fitted 3D model to obtain a profile view, and finally rotating and rendering it back to reconstruct a frontal view. To summarize,  \cite{zhou2020rotate}  uses \cite{zhu2019face} to estimate the rigid pose and the face deformation parameters in order to frontalize the face, and \cite{zhu2017unpaired} to fill in the occluded regions caused by frontalization. Although this method yields state-of-the-art results for the task of face recognition, there is no guarantee that non-rigid facial deformations are preserved by the profile-to-frontal mapping process.


Interestingly, there has only been a handful of attempts to combine dynamic models with facial shape deformation. \cite{baumberg1998hierarchical,lee2007hierarchical} use a Kalman filter to track a face in an image sequence and to initialize the parameters of a deformable shape model. In \cite{prabhu2010automatic} a Kalman filter is used to predict the location of individual landmarks, from the previous frame to the current frame, and to use these predictions to initialize the parameters of a deformable shape model. The dynamical model that we propose in this paper is totally different because it dynamically updates a deformable model with two interconnected latent variables -- this dynamic face-deformation model is inferred in alternance with landmark frontalization.

\addnote[3dfa-features]{1}{
The proposed method requires 3D facial landmarks. Recently there has been a flourishing literature on this topic, yielding several \ac{DNN} \ac{3DFA} models and associated software packages, e.g. \cite{bulat2016two,zhu2016face,feng2018joint,deng2018cascade,zhu2019face,jiang2019dual,tu20203d,ning2020real}. We thoroughly analysed and benchmarked four publicly available \ac{3DFA} software packages. The results reported in this paper were obtained with the method of \cite{bulat2016two}. The latter is trained using a very large dataset \cite{bulat2017far} and it assumes a weak-perspective camera model. Recently, it has been shown that the perspective camera model is better suited for guaranteeing the separation of rigid and non-rigid facial deformations \cite{sariyanidi2020can}. We propose an alternative rigid/non-rigid separation formulation based on robust statistics.
}

This article is an extended version of \cite{kang2021robust} and it contains two extensions: a dynamic face-deformation model and its inference based on linear dynamical systems, Section~\ref{section:dff}, and an in depth investigation of the effect of face frontalization on the performance of audio-visual speech enhancement, Section~\ref{sec:avse}.

\section{Expression-preserving face frontalization}
\label{sec:face-frontalization}
\begin{figure*}[t]
\centering
\subfloat[Head-pose and face-deformation estimation]{\includegraphics[width=0.45\linewidth]{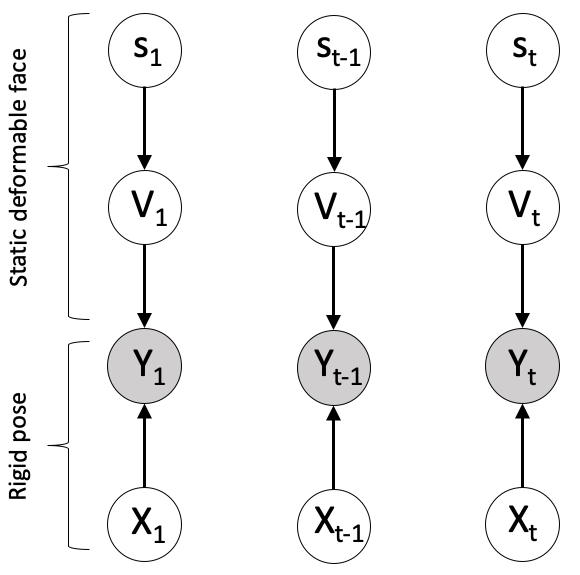}}
\subfloat[Dynamic face-deformation inference]{\includegraphics[width=0.45\linewidth]{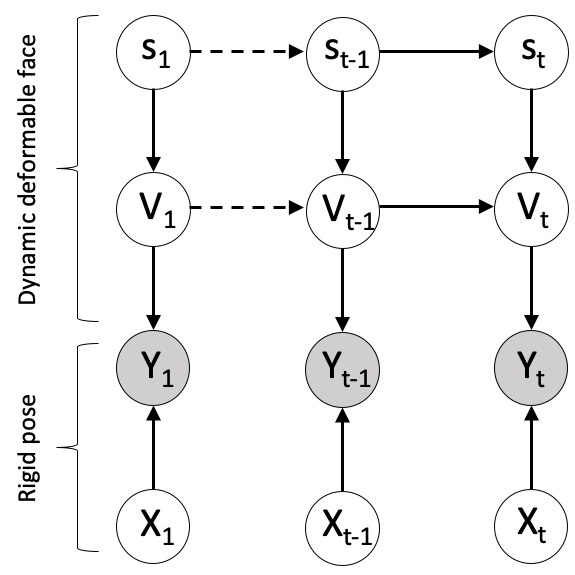}}
\caption{\label{fig:graphicalmodel} 
Both models alternate between the estimation of the rigid head-pose parameters and the estimation of the face-deformation parameters. The arrows illustrate stochastic generative models. In both cases, the frontalized landmarks $\Yvect$ are generated from the observed landmarks $\Xvect$.
The static model (a) uses the Student's t-distribution to estimate the rigid (head pose) and non rigid (face deformation) parameters. In addition, the temporal model (b) makes use of two stochastic dynamical regimes, one that governs the evolution of the shape embedding $\svect$ and a second one that governs the evolution of the shape vertices $\Vvect$. Note that these two latent variables are interconnected and that the frontalized landmarks are used as observations by the proposed doubly-latent LDS.}
\end{figure*}

In this section we describe in detail the static and dynamic models that reside at the core of the proposed expression-preserving \ac{FF} framework. These models are graphically represented in Figure~\ref{fig:graphicalmodel}. After briefly presenting the face deformation model, we describe in detail the estimation of the head-pose and face-deformation parameters, followed by a description of the dynamical formulation and the associated statistical inference. The final stage consist of warping the input face onto a frontal view in such a way that facial deformations remain invariant. 

\subsection{Face deformation model}
\label{sec:face-deformation-model}
In order to model non-rigid facial deformations, we consider a 3D deformable shape model \cite{blanz1999morphable}. Such a model is learnt from a training set of 3D faces, or meshes, $\mathcal{M}=\{\Mvect_m\}_{m=1}^M$. Each face $m$ in the training set is described by $N$ 3D vertices, namely $\Mvect_m=(\Mvect_{m1}, \dots, \Mvect_{mn}, \dots, \Mvect_{mN})\tp\in\mathbb{R}^{3N}$; moreover, the faces are registered: their vertices are in one-to-one correspondence. Let $\Cmat=1/M\sum_{m=1}^M(\Mvect_m - \overline{\Mvect})(\Mvect_m - \overline{\Mvect})\tp$ be the covariance matrix associated with this training set, where the \textit{mean shape} is defined by $\overline{\Mvect}=1/M\sum_{m=1}^M\Mvect_m$, and let
$(\Lambdamat, \Umat)$ be the $K$ principal eigenvalue-eigenvector pairs of $\Cmat$, where $\Lambdamat = \diag \begin{pmatrix} \lambda_1, \dots, \lambda_{K}\end{pmatrix}$, with $\lambda_1 \geq \dots \geq  \lambda_{K}\geq 0$ and $\Umat = \begin{pmatrix} \Uvect_1 & \dots & \Uvect_{K}\end{pmatrix}\in\mathbb{R}^{3N \times K}$ is a column-orthogonal matrix, i.e. $\Umat\tp\Umat = \Imat_K$, with $K\ll 3N$. 

The vertices of a face $\Vvect = (\Vvect_1, \dots, \Vvect_n, \dots, \Vvect_N)\tp\in\mathbb{R}^{3N}$ can be projected onto the low-dimensional space spanned by the principal eigenvectors, namely 
\begin{equation}
\label{eq:s-embedding}
\svect = \Umat\tp (\Vvect - \overline{\Mvect}),
\end{equation}
where $\svect\in\mathbb{R}^K$ is the face embedding (or encoding). 
Conversely, it is possible to reconstruct (or decode) the face $\Vvect$ from its embedding $\svect$, and this up to a \textit{decoding} error $\Fvect$:
\begin{equation}
\label{eq:mesh-rec}
\Vvect = \Umat\svect + \overline{\Mvect} + \Fvect, \; \mathrm{s.t.} \; \svect\tp\Lambdamat\inverse\svect \leq 1.
\end{equation}
The above inequality constrains the reconstructed mesh to correspond to an embedding $\svect$ that lies inside an ellipsoid with half axes equal to $\sqrt{\lambda_k}$. This guarantees with 99\% confidence that $\Vvect$ belongs to the space spanned by the training set.
Therefore, each vertex $\Vvect_n$ of $\Vvect$ can be reconstructed from $\svect$ with 
\begin{equation}
\label{eq:vertex-rec}
\Vvect_n = \Umat_n \svect + \overline{\Mvect}_n + \Fvect_n = \Wmat_n\Svect + \Fvect_n,
\end{equation}
where $\Umat_n\in\mathbb{R}^{3\times K}$ is such that $\Umat=(\Umat_1\dots\Umat_n\dots\Umat_N)$, $\Svect=[\svect ; 1]\in\mathbb{R}^{K+1}$ (vertical concatenation) and $\Wmat_n = (\Umat_n \; \overline{\Mvect}_n)\in\mathbb{R}^{3\times(K+1)}$.

\subsection{Head-pose and face-deformation estimation}

We now consider an image of a face with an unknown pose. Let $\Xvect=(\Xvect_1, \dots, \Xvect_j, \dots, \Xvect_J)\tp\in\mathbb{R}^{3J}$ be a vector of $J$ 3D landmarks extracted from this face. The pose is parameterized  by a rigid transformation, namely scale $\rho\in\mathbb{R}^{+}$, rotation $\Rmat\in\mathrm{SO(3)}$, and translation $\Tvect\in\mathbb{R}^3$,  between the observed landmarks $\Xvect$ and the frontalized landmarks $\Yvect\in\mathbb{R}^{3J}$:
\begin{equation}
\label{eq:rigidmapping}
\Yvect_j=\rho \Rmat \Xvect_j + \Tvect, \; \forall j\in\{1\dots J\}.
\end{equation}
These landmarks correspond to $J$ vertices,  annotated such that there is a one-to-one correspondence between $\{ \Vvect_j \}_{j=1}^J$ and $\{ \Yvect_j \}_{j=1}^J$ up to an error $\Dvect_j$. We have:
\begin{equation}
\label{eq:YfromV}
\Yvect_j = \Vvect_j + \Dvect_j = \Wmat_j\Svect + \Evect_j,
\end{equation}
where $\Evect_j = \Dvect_j + \Fvect_j$ is the total error. By combining the above equations, we obtain:
\begin{equation}
\label{eq:error-j}
\Evect_j = \rho \Rmat \Xvect_j + \Tvect - (\Umat_j\svect + \overline{\Mvect}_j),
\end{equation}
Assuming that these error vectors are random variables drawn from a probability distribution function (pdf) we can write a maximum likelihood estimator (MLE), or equivalently, the minimization of the following negative log-likelihood function:
\begin{equation}
\mathcal{L}(\thetavect | \Xvect) = - \sum_{j=1}^{J}  \log p (\Evect_j  ;  \thetavect),
\end{equation}
where $\thetavect$ is the vector of model parameters, i.e. the rigid parameters, the shape parameters and the pdf parameters.
From \eqref{eq:error-j} one may see that it is possible to alternate between the estimation of the shape parameters $\svect$ and the rigid (frontalization) parameters $\rho$, $\Rmat$, and $\Tvect$. Because the 3D landmarks are affected by noise and by non-rigid facial deformation, we opt for a robust pdf, namely the generalized Student's t-distribution:
\begin{align}
\label{eq:generalized-student}
p(\Evect_j; \thetavect)   = \int_{0}^{\infty} \mathcal{N} ( \Evect_j; 0, \omega_j \inverse \sigmavect) \mathcal{G}(\omega_j; \mu, 1) d\omega_j,
\end{align}
where $\mathcal{N}(\Evect;\zerovect,\omega\inverse\sigmavect)$ denotes a zero-centered normal distribution, $\omega\in\mathbb{R}^{+}$ is a precision and $\sigmavect\in\mathbb{R}^{3\times 3}$ is a covariance matrix. The precision $\omega$ is treated as a latent variable drawn from the Gamma distribution and it can be interpreted as an observation weight. Therefore the variables $\omega_{1:J}$ characterize the landmarks $\Xvect_{1:J}$: the higher the better. Unfortunately, direct minimization of \eqref{eq:error-j} using \eqref{eq:generalized-student} is intractable. Therefore one has to adopt a \ac{ECM} formalism: the negative log-likelihood is replaced with the expected complete-data negative log-likelihood conditioned by the observed data, $E_\omega[-\log P(\omega_{1:J}, \Xvect | \Xvect]$. \ac{ECM} alternates between an E-step and an several conditional M-steps, i.e. Algorithm~\ref{algo:rff}.
 \begin{algorithm}[h!]
 \caption{\label{algo:rff} Robust face frontalization (RFF).}

 \KwData{3D landmark coordinates $\Xvect_{1:J}$, 3D shape reconstruction matrix $\Umat$ and mean shape $\overline{\Mvect}$.}

 \textbf{Initialization}:  $\svect=\zerovect$, $\sigmavect=\Imat$, $\overline{\omega}_{1:J}=1_{1:J}$ and $\mu=1$\; 
 Compute $\Xvect\pri_{1:J}$ and $\Vvect\pri_{1:J}$  with \eqref{eq:X-centers}, \eqref{eq:V-centers}\; 
 Compute $\rho$ with \eqref{eq:opt-scale-robust}\;
 Use $\sigmavect=\Imat$ in \eqref{eq:opt-rotation-robust} to estimate $\Rmat$ in closed form\;
 Compute $\Tvect$, $\sigmavect$ and $\svect$ with \eqref{eq:opt-translation-robust}, \eqref{eq:covariance-student}, \eqref{eq:shape-student}\;
This yields $\thetavect=(\rho, \Rmat, \Tvect, \svect, \sigmavect)$.
 
 \While{$\|\thetavect - \thetavect^{\star}\|>\epsilon$}
 { 
 \textbf{E-step}: Evaluate $a$, $b_{1:J}$ and $\overline{\omega}_{1:J}$ with \eqref{eq:gamma-posterior}, \eqref{eq:weight-expectation}\; 
 Update $\Xvect\pri_{1:J}$ and $\Vvect\pri_{1:J}, \widetilde{\Xvect}, \widetilde{\Vvect}$ with \eqref{eq:X-centers}, \eqref{eq:V-centers}. \;
 
  \textbf{M-rigid-step}: Evaluate the new rigid parameters  and  covariance $\rho^{\star}, \Rmat^{\star}, \Tvect^{\star}, \sigmavect^{\star}$ with \eqref{eq:mu-digamma}-\eqref{eq:covariance-student}\;
    \textbf{M-non-rigid-step}: Evaluate the new non-rigid parameters $\svect^{\star}$ with \eqref{eq:shape-student}\;
  $\thetavect \leftarrow \thetavect^{\star}$\;

 } 

  \KwResult{\\Optimal model parameters $\thetavect^{\star}=(\rho^{\star}, \Rmat^{\star},\Tvect^{\star}, \svect^{\star}, \sigmavect^{\star})$\\ Frontalized landmarks $\Yvect_{1:J}=\rho^{\star} \Rmat^{\star} \Xvect_{1:J} + \Tvect^{\star}.$}

\end{algorithm}

 The E-step computes the parameters of the weights' posterior distributions $\mathcal{G}(\omega_j; a,b_j)$, namely
 \begin{equation}
\label{eq:gamma-posterior}
a= \mu + \frac{3}{2}, \quad b_j = 1+\frac{\|\Evect_j\|^2_{\sigmavect}}{2},
\end{equation}
from which the posterior means are evaluated:
\begin{equation}
\label{eq:weight-expectation}
\overline{\omega}_j = \mathrm{E}[\omega_j | \Evect_j ] = \frac{a}{b_j}.
\end{equation}
The M-step consists of the estimation of the model parameters, namely $\thetavect = (\rho, \Rmat, \Tvect, \sigmavect, \svect)$, via the minimization of:
\begin{align}
\label{eq:mstep}
Q(\rho, \Rmat, \Tvect, \svect, \sigmavect) & =  \sum_{j=1}^{J} \overline{\omega}_j \| \rho \Rmat \Xvect_j + \Tvect  - \Wmat_j\Svect \|^2_{\sigmavect} \nonumber \\
& + \log |\sigmavect| + \kappa \svect\tp\Lambdamat\inverse\svect,
\end{align}
and the computation of the parameter $\mu$ of the Gamma distribution, where $\Psi(a)\approx \log a - 1/2a$ is the digamma function:
\begin{equation}
\label{eq:mu-digamma}
\mu = \Psi^{-1}\left( \Psi(a) -\frac{1}{n} \sum_{j=1}^{J} \log b_j  \right).
\end{equation}
The last term of \eqref{eq:mstep} is a regularizer that forces $\svect$ to correspond to a \textit{valid} shape, i.e, the constraint of \eqref{eq:mesh-rec}.
The minimization  of \eqref{eq:mstep} with respect to the parameters yields the following conditional expressions:
\begin{align}
\label{eq:opt-scale-robust}
\rho^{\star}  =&  \left( \frac{\sum_{j=1}^{J} \overline{\omega}_j \Vvect_j^{\prime\top}  \sigmavect\inverse \Vvect\pri_j}{\sum_{j=1}^{J} \overline{\omega}_j (\Rmat \Xvect_j\pri )\tp  \sigmavect\inverse (\Rmat\Xvect\pri_j)} \right) ^{\frac{1}{2}}, \\
\label{eq:opt-rotation-robust}
\Rmat^{\star} =&  \argmin_{\Rmat} \sum_{j=1}^{J} ( \overline{\omega}_j \| \Vvect\pri_{j}- \rho^{\star} \Rmat \Xvect\pri_j \|^{2}_{\sigmavect}), \\
\label{eq:opt-translation-robust}
\Tvect^{\star} =& \widetilde{\Vvect} - \rho^{\star} \Rmat^{\star} \widetilde{\Xvect},
\end{align}
\begin{align}
\label{eq:covariance-student}
\sigmavect^{\star} &= \frac{1}{J} \sum_{j=1}^{J} \overline{\omega}_j (\Vvect\pri_j - \rho^{\star} \Rmat^{\star} \Xvect\pri_j) (\Vvect\pri_j - \rho^{\star} \Rmat^{\star} \Xvect\pri_j)\tp,
\end{align}
\begin{align}
\label{eq:shape-student}
\svect^{\star} =& \left(\sum_{j=1}^{J}\overline{\omega}_j \Umat_j\tp{\sigmavect^{\star}}\inverse\Umat + \kappa\Lambdamat\inverse\right)\inverse 
\nonumber \\ 
& \left(\sum_{j=1}^{J} \overline{\omega}_j \Umat_j\tp{\sigmavect^{\star}}\inverse (\rho^{\star} \Rmat^{\star} \Xvect_j + \Tvect^{\star}  - \overline{\Mvect}_j )\right), 
\end{align}
where $\Xvect\pri_j, \Vvect\pri_j, \widetilde{\Xvect}, \widetilde{\Vvect}$  are computed with:
\begin{align}
\label{eq:X-centers}
  \Xvect\pri_j & =\Xvect_j - \widetilde{\Xvect}, \; \; \widetilde{\Xvect} = \frac{ \sum_{j=1}^{J} \overline{\omega}_j \Xvect_j}{\sum_{j=1}^N \overline{\omega}_j},
  \\
  \label{eq:V-centers}
\Vvect\pri_j & = \Vvect_j  - \widetilde{\Vvect},  \; \; \widetilde{\Vvect}  = \frac{ \sum_{j=1}^{J} \overline{\omega}_j \Vvect_j }{\sum_{j=1}^N \overline{\omega}_j} \\
\label{eq:vertex-update}
\Vvect_j &= \Umat_j\svect^{\star} +\overline{\Mvect}_j.
\end{align}
The \ac{ECM}  procedure is summarized in Algorithm~\ref{algo:rff}.

\subsection{Dynamic face deformation inference}
\label{section:dff}

We now describe a dynamic model for estimating a time-varying deformable face. Let $\Yvect_{1:t}$ ($1:t$ is a shorthand for $1, 2, \dots, t$) be the sequence of frontalized landmarks obtained with Algorithm~\ref{algo:rff}, where $\Yvect_t\in\mathbb{R}^{3J}$ is the vector of frontalized landmarks at $t$. For the sake of clarity, we regroup \eqref{eq:vertex-rec}, \eqref{eq:rigidmapping} and \eqref{eq:YfromV}:
\begin{align}
\label{eq:decoding}
\Vvect_{t} &= \Wmat \Svect_t + \Fvect_{t}, \\
\label{eq:detecting}
\Yvect_{t} &= \Vvect_{t} + \Dvect_{t}, \\
\Yvect_{tj} &=\rho_t \Rmat_t \Xvect_{tj} + \Tvect_t,\; j\in\{1\dots J\},
\end{align}

We assume that the sequences $\Svect_{1:t}$ and $\Vvect_{1:t}$ are Markovian stochastic processes, each one with its own dynamic regime and interconnected via \eqref{eq:decoding}. Moreover, $\Vvect_{1:t}$ and $\Yvect_{1:t}$ are interconnected via \eqref{eq:detecting}. The graphical model shown on Figure~\ref{fig:graphicalmodel} describes the proposed \ac{DL-LDS}. Probabilistically, this system can be described with the following conditional distributions:
\begin{align}
\label{eq:transition-distribution-shape}
& p(\Svect_t | \Svect_{t-1}) = \mathcal{N} (\Svect_t;  \Svect_{t-1}, \Gammamat_S), \\
\label{eq:transition-distribution-vertices}
& p(\Vvect_t | \Vvect_{t-1}, \Svect_t) = \mathcal{N} (\Vvect_t; \alpha \Vvect_{t-1}
 + (1-\alpha) \Wmat \Svect_t, \Gammamat_V), \\
\label{eq:emission-distribution}
& p(\Yvect_t | \Vvect_{t}) = \mathcal{N} (\Yvect_t; \Vvect_{t}, \Sigmamat_t), 
\end{align}
where
$\Gammamat_S\in\mathbb{R}^{(K+1)\times(K+1)}$, $\Gammamat_V\in\mathbb{R}^{3J\times 3J}$, $\Sigmamat_t \in\mathbb{R}^{3J\times 3J}$ are covariance matrices, and where $\Sigmamat_t= \Imat_{J\times J} \otimes \sigmavect_t$ is obtained from \eqref{eq:covariance-student} ($\otimes$ denotes the Kronecker product).
The main difference between standard
LDSs and the proposed DL-LDS resides in the fact that there are two interconnected latent variables, having their own dynamical regimes. Consequently, the transition probability of $\Vvect_t$, \eqref{eq:transition-distribution-vertices} is conditioned both by $\Vvect_{t-1}$ and by $\Svect_t$. The scalar $\alpha\in[0,1]$ weights the relative importance of the vertex dynamics and of the vertex reconstruction from the current shape embedding.

We now show that the above DL-LDS can be cast into a standard LDS, i.e.\ the Kalman filter. Let the latent variable $\Zvect=[\Svect ; \Vvect]\in\mathbb{R}^{K+1+3J}$ be the concatenation of the two latent variables. In the particular case of our graphical model, we have:
\begin{align}
\label{eq:transition-pdf-conc}
p (\Zvect_t | \Zvect_{t-1} ) &= p(\Svect_t, \Vvect_t | \Svect_{t-1}, \Vvect_{t-1}) \nonumber \\
& = p(\Vvect_t | \Svect_{t}, \Vvect_{t-1} ) \; p (\Svect_t | \Svect_{t-1}),
\end{align}
where
the pdfs on the second row are given by \eqref{eq:transition-distribution-shape} and \eqref{eq:transition-distribution-vertices}. Let
$p (\Zvect_t | \Zvect_{t-1} ) = \mathcal{N} (\Zvect_t ; \muvect_t, \Gammamat)$.
By taking the logarithm of both sides of \eqref{eq:transition-pdf-conc} and by identifying the quadratic and linear terms, we obtain:
\begin{align}
 \muvect_t &= \Gammamat \Amat \Zvect_{t-1} \\
 \Gammamat^{-1} &= \begin{pmatrix}
\Gammamat_S^{-1}+(1-\alpha)^2\Wmat^\top\Gammamat_V^{-1}\Wmat & -(1-\alpha)\Wmat^\top\Gammamat_V^{-1} \\
-(1-\alpha)\Gammamat_V^{-1}\Wmat & \Gammamat_V^{-1}
 \end{pmatrix}\\
 \Amat &= 
 \begin{pmatrix}
 \Gammamat_S^{-1} & -\alpha(1-\alpha)\Wmat^\top\Gammamat_V^{-1} \\
 0 & \alpha\Gammamat_V^{-1}
\end{pmatrix}
\end{align}
To summarize, \eqref{eq:transition-distribution-shape}, \eqref{eq:transition-distribution-vertices} and \eqref{eq:emission-distribution} can be rewritten as:
\begin{align}
p(\Zvect_t | \Zvect_{t-1} ) &= \mathcal{N} (\Zvect_t ; \Gammamat \Amat \Zvect_{t-1}, \Gammamat) \\
p(\Yvect_t | \Zvect_{t}) &= \mathcal{N} (\Yvect_t; \Cmat \Zvect_{t}, \Sigmamat_t),
\end{align}
where matrix $\Cmat\in\mathbb{R}^{3J\times(K+1+3J)}$ projects the concatenated latent-variable space onto the space of observed variables. Similarly, matrix $\overline{\Cmat}\in\mathbb{R}^{K\times(K+1+3J)}$ projects the concatenated latent-variable space onto the space of the shape embedding. These two matrices write:
\begin{align}
\label{eq:projection-shape}
\Cmat & = \begin{pmatrix} \zerovect_{3J\times (K+1)} & \Imat_{3J\times 3J} \end{pmatrix}\\
\label{eq:projection-vertex}
\overline{\Cmat} &= \begin{pmatrix} \Imat_{K\times K} & \zerovect_{(3J+1)\times K} \end{pmatrix}
\end{align}
We now follow the standard Bayesian derivation of the Kalman filter.
For this purpose, we need to evaluate the following posterior and prior distributions:
\begin{align}
\label{eq:posterior-lds}
p(\Zvect_t | \Yvect_{1:t} ) &= \mathcal{N} (\Zvect_t ;  \nuvect_t, \Psimat_t), \\
\label{eq:x0-prior}
p(\Zvect_1) &= \mathcal{N} (\Zvect_1; \nuvect_1, \Psimat_1),
\end{align}
where $\nuvect_t\in\mathbb{R}^{K+1+3N}$ and $\Psimat_t\in\mathbb{R}^{(K+1+3N)\times(K+1+3N)}$ are the mean and covariance, respectively.  Applying the standard derivation of an LDS we have:
\begin{align}
\label{eq:Bayes-Kalman}
p(\Zvect_t | \Yvect_{1:t} ) \; p(\Yvect_t | \Yvect_{1:t-1}) = p(\Yvect_t | \Cmat \Zvect_t) \; p (\Zvect_t  | \Yvect_{1:t-1}),
\end{align}
as well as the marginalization:
\begin{align}
\label{eq:Bayes-marginalization}
p (\Zvect_t  | \Yvect_{1:t-1}) 
=  \int p (\Zvect_t | \Zvect_{t-1} ) 
p (\Zvect_{t-1} |  \Yvect_{1:t-1})
d\Zvect_{t-1}.
\end{align}
The integral can then be evaluated making use of the results of \cite{bishop2006pattern}:
\begin{align}
\int  \mathcal{N} (\Zvect_t ;  &\Gammamat \Amat  \Zvect_{t-1}, \Gammamat)  \mathcal{N} (\Zvect_{t-1}; \nuvect_{t-1}, \Psimat_{t-1}) d\Zvect_{t-1} \nonumber \\
& = \mathcal{N} (\Zvect_t ; \Gammamat \Amat \nuvect_{t-1}, \Pmat_{t-1})\\
\label{eq:Pmat}
 \mathrm{with:} \quad  \Pmat_{t-1} & = \Gammamat \Amat \Psimat_{t-1} \Amat\tp \Gammamat\tp + \Gammamat
\end{align}
We can now write \eqref{eq:Bayes-Kalman} as:
\begin{align}
\mathcal{N}  & (\Zvect_t ;  \nuvect_t, \Psimat_t) \; p(\Yvect_t | \Yvect_{1:t-1}) \\
& = \mathcal{N} (\Yvect_t; \Cmat \Zvect_t, \Sigmamat_t) \; \mathcal{N} (\Zvect_t ; \Gammamat \Amat \nuvect_{t-1}, \Pmat_{t-1}), \nonumber
\end{align}
from which we obtain the following recursive formulas:
\begin{align}
\label{eq:mean-rec}
\nuvect_t &= \Gammamat \Amat \nuvect_{t-1} + \Kmat_t (\Yvect_t - \Cmat \Gammamat \Amat \nuvect_{t-1} ) \\
\label{eq:cov-rec}
\Psimat_t &= (\Imat - \Kmat_t \Cmat) \Pmat_{t-1}\\
\label{eq:kalman-rec}
\Kmat_t &= \Pmat_{t-1}  \Cmat\tp (\Cmat  \Pmat_{t-1}  \Cmat\tp + \Sigmamat_t)\inverse\\
p(\Yvect_t | \Yvect_{1:t-1}) &= \mathcal{N} (\Yvect_t ; \Cmat \Gammamat \Amat \nuvect_{t-1}, \Cmat  \Pmat_{t-1}  \Cmat\tp + \Sigmamat_t)
\end{align}

In order to initialize the above recursion, one needs to provide the mean and covariance of the prior distribution, $\nuvect_1$ and $\Psimat_1$, as well as the covariances $\Gammamat_S$ and $\Gammamat_V$ associated with the dynamics of $\Svect$ and of $\Vvect$, respectively, i.e. \eqref{eq:transition-distribution-shape} and \eqref{eq:transition-distribution-vertices}. Let $\Svect_1$ be the shape embedding at $t=1$, which is provided by Algorithm~\ref{algo:rff}. The face vector of vertex coordinates is therefore estimated with $\Vvect_1=\Wvect\Svect_1$. We have:
\begin{align}
\label{eq:DLDS-init}
\nuvect_1=\begin{pmatrix} \Svect_1 \\ \Vvect_1 \end{pmatrix}, \; \Psimat_1=\Imat,\; \Pmat_1 = \Imat.
 \end{align}
Algorithm~\ref{algo:trff} describes an implementation of the proposed doubly-latent LDS combined with the robust estimation of the rigid transformation required by face frontalization. The output of Algorithm~\ref{algo:trff} is a temporal sequence of estimated embedding and vertices:
\begin{align}
\label{eq:s-inference}
\hat{\svect}_t &= \overline{\Cmat} \nuvect_t  \quad t\in\{1\dots T\}, \\
\label{eq:V-inference}
\hat{\Vvect}_t &= \Cmat\nuvect_t \quad t\in\{1\dots T\}.
\end{align}
\begin{algorithm}[th!]
 \caption{\label{algo:trff} Dynamic face frontalization (DFF).}

 \KwData{Temporal sequence of input landmark coordinates $\Xvect_{1:T} = (\Xvect_{1} \dots \Xvect_{T})$, with $\Xvect_t=(\Xvect_{t1} \dots \Xvect_{tJ})$, 3D shape reconstruction matrix $\Umat$ and mean shape $\overline{\Mvect}$, covariance matrices $\Gammamat_S$, $\Gammamat_V$ and scalar $\alpha$.}
 \textbf{Initialization:}  Use Algorithm~\ref{algo:rff} to Initialize the DL-LDS parameters $\nuvect_1$ and $\Psimat_1$ with \eqref{eq:DLDS-init}\;
\While{$t = 2 \dots T$}
 { 
 \textbf{Rigid-pose}: Use Algorithm~\ref{algo:rff} to compute $\Yvect_t$\;
 Evaluate $\Sigmamat_t= \Imat_{J\times J} \otimes \sigmavect_t^{\star}$\;
 \textbf{DL-LDS-recursion}:
Apply the recursive formulas \eqref{eq:Pmat}, \eqref{eq:mean-rec}, \eqref{eq:cov-rec} and \eqref{eq:kalman-rec} to compute parameters $\nuvect_t, \Psimat_t$ and the gain matrix $\Kmat_t$\;
 Evaluate $\svect_t$, $\Vvect_t$ with \eqref{eq:s-inference}, \eqref{eq:V-inference}\;
}
 
  \KwResult{Temporal sequence of shape embedding $\hat{\svect}_{1:T}$, shape vertices $\hat{\Vvect}_{1:T}$, and covariances $\Psimat_{1:T}$.}

\end{algorithm}

\subsection{Face warping}
\label{sec:depth-and-warp}
A frontal view of the face is computed in the following way. For convenience, the temporal index $t$ is dropped. A frontal 3D shape is first computed with \eqref{eq:vertex-rec}. 
The points $\Vvect_{1:N}$ are the vertices of a 3D triangulated mesh and therefore the projection of this mesh onto the frontal image $I_f$ form a 2D triangulated mesh; assuming orthographic projection, the image coordinates of a vertex $\Vvect_n$ are $(V_{n1}, V_{n2})$. Let $k_1$, $k_2$ and $k_3$ be the vertex indexes of a mesh triangle. We now compute the barycentric coordinates, $(\beta_1, \beta_2, \beta_3) \in \mathbb{R}^3$ of a pixel $(a_1, a_2)\in \mathbb{N}^2$ that lies inside that triangle, i.e. $0 \leq \beta_1, \beta_2, \beta_3 \leq 1$. These barycentric coordinates correspond to the solution of the following set of linear equations:
\begin{align}
\label{eq:barycenter}
 \begin{pmatrix}
 a_1 \\ a_2
 \end{pmatrix}  & = \beta_1 \begin{pmatrix}V_{k_11} \\V_{k_12} \end{pmatrix} + \beta_2 \begin{pmatrix}V_{k_21}\\V_{k_22}\end{pmatrix} + \beta_3 \begin{pmatrix}V_{k_31} \\V_{k_32}\end{pmatrix} \\
 1 &= \beta_1+\beta_2+\beta_3
\end{align}
Once the barycentric coordinates are computed, the depth $A_3\in \mathbb{R}$ associated with pixel  $(a_1 \; a_2)\tp$ is computed by linear interpolation, namely:
\begin{equation}
\label{eq:weighted-interpolation}
A_3 = \beta_1V_{k_13} + \beta_2V_{k_23} + \beta_3V_{k_33}
\end{equation}
The above procedure is repeated for all the triangles and for all the points inside each triangle, thus obtaining a frontal dense depth map for each face pixel. Let $\Avect=(a_1\; a_2\; A_3)\tp$ be the current point of the frontal dense depth map thus obtained.

The final face frontalization step consists of \textit{warping} the face's pixel colors from the input-image $I_p$ onto a synthesized frontal image $I_f$.
The rigid transformation that maps the 3D face, from a frontal centered coordinate frame
back onto the input view, is the inverse of the pose, namely $\rho\pri = \rho\inverse$, $\Rmat\pri = \Rmat\tp$, and $\Tvect\pri = -\rho\inverse\Rmat\tp \Tvect$. The dense depth map of the face can therefore be mapped back with
\begin{align}
\label{eq:pixel-pred}
\begin{pmatrix}
B_{1} \\ B_{2} \\ B_{3}
\end{pmatrix} =
\rho\pri \Rmat\pri 
\begin{pmatrix}
a_{1} \\ a_{2}  \\ A_{3} 
\end{pmatrix}
+
\Tvect\pri
\end{align}

\begin{figure}[t!h!]
\centering
 \includegraphics[width=\linewidth]{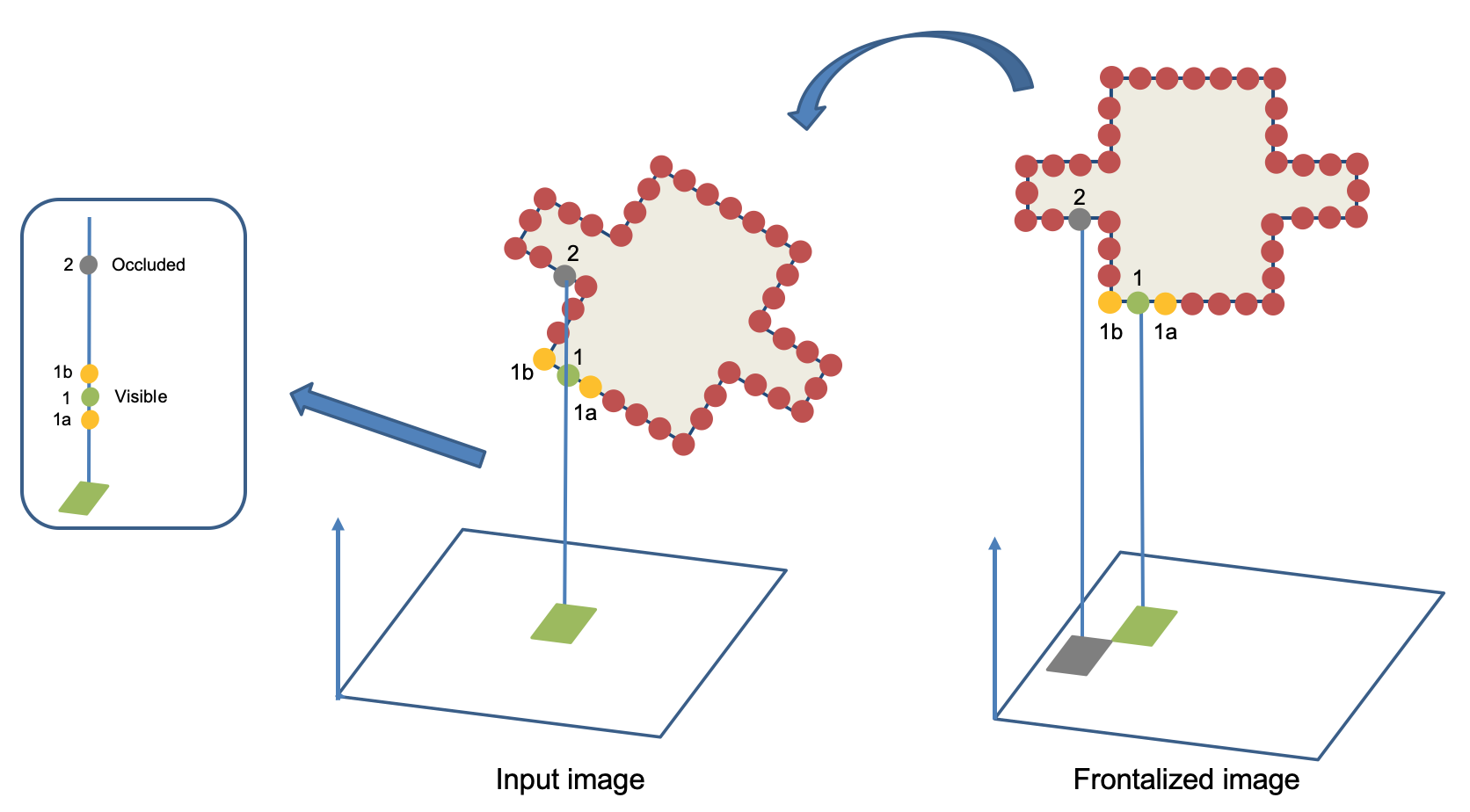}
\caption{\label{fig:warping} When an object is rotated to appear frontal, some of its vertices have no associated photometric information in the synthesized frontal image, because they are not visible in the input image. In this example, vertices 1 and 2 are both visible in the frontalized image, but only 1 is visible in the input image. Because of quantization noise, the one-ring neighbors of 1, 1a and 1b, lie on the same line of sight as 1. We disregard this quantization effect and mark 1 as visible.}
\end{figure}

Assuming scaled orthographic projection, the 2D pixel location $(b_1,b_2) \in I_p$ is computed from the real-valued coordinates $(b_1,b_2) = ([B_1], [B_2])$, where $\lbrack \cdot \rbrack$ is the \textit{round} operator.
Because of self occlusions and of quantization, \eqref{eq:pixel-pred}  maps several points, $\Avect^{1:Q}$, at the same pixel location, but with different depth values $(b_1, b_2, B_3^{1:Q})$. Notice that only the depth-map point with the smallest depth value should visible in the input image. Consequently, vertices that are not visible in the input image don't have any photometric information associated with them and hence, they give rise to blank areas in the frontalized image.
The final face frontalization step consists of synthesizing a frontal image:
\begin{equation}
\label{eq:face-warping}
I_f (a_1, a_2) =
\begin{cases}
 I_p (b_1, b_2)  & \mathrm{if} \; B_3 = \min_q \{ B_3^q\}_{q=1}^Q\\
\emptyset & \mathrm{otherwise},
\end{cases}
\end{equation}
where $\emptyset$ means that there is no photometric information available with that pixel. This is illustrated on Figure~\ref{fig:warping}. 

\section{Implementation details}
\label{sec:method-implementation-details}
\begin{figure*}[t!]
\begin{center}
\begin{tabular}{ccc}
\includegraphics[width=.33\textwidth,trim={0.cm, 0.cm, 0cm, 0cm},clip]{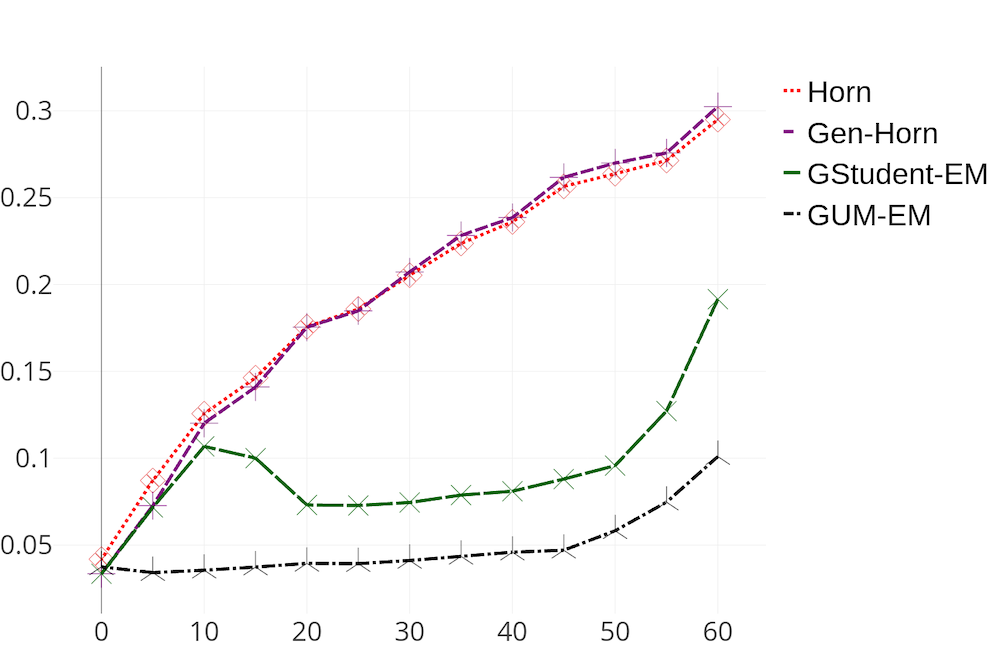} &
\includegraphics[width=.33\textwidth,trim={0.cm, 0.cm, 0cm, 0cm},clip]{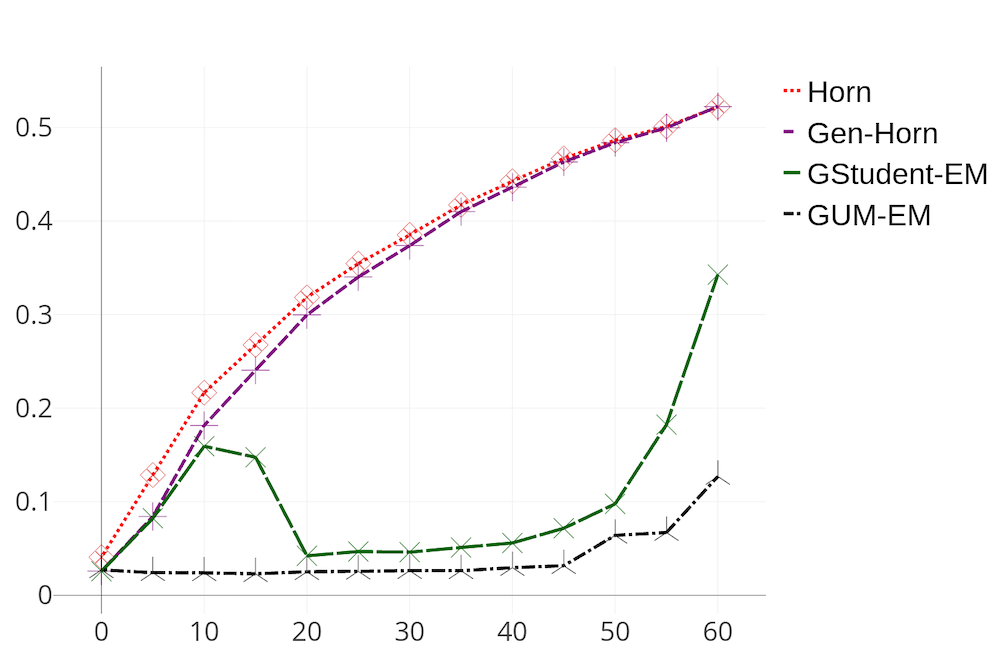} &
\includegraphics[width=.33\textwidth,trim={0.cm, 0.cm, 0cm, 0cm},clip]{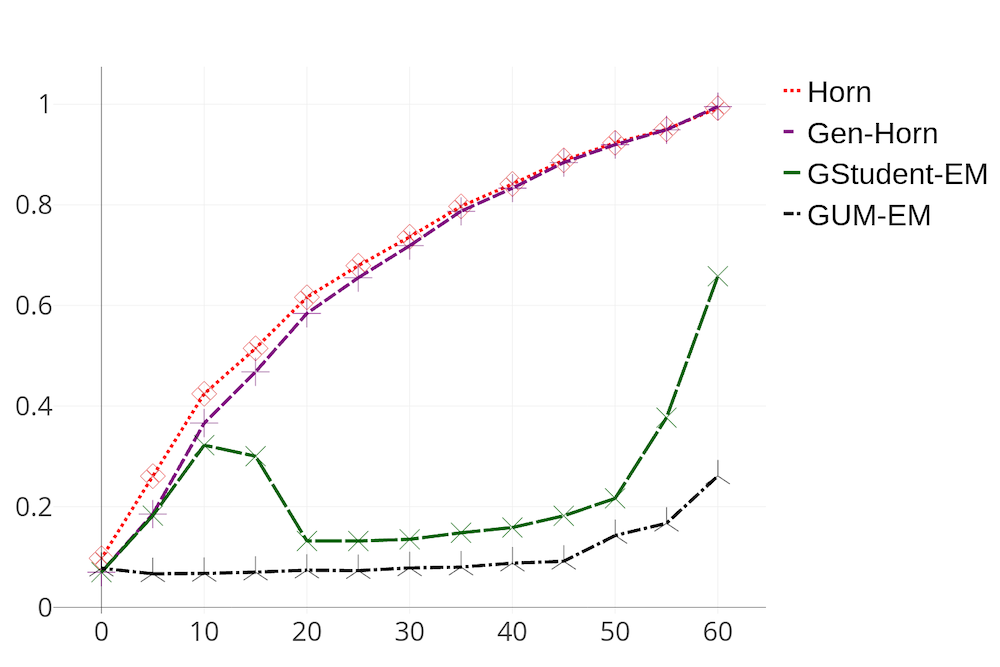}\\
Rotation & Scale & Translation
\end{tabular}
\end{center}
\caption{\label{fig:simulations} The root mean-square error as a function of the percentage of outliers (i.e. landmark localization errors) averaged over 500 trials.}
\end{figure*}

All the computations inside Algorithm~\ref{algo:rff} are in closed-form, with the notable exception of the estimation of the rotation matrix. The latter is parameterized with a unit quaternion \cite{horn1987closed}, which allows one to reduce the number of rotation parameters, from nine to four, and to express the orthogonality constraints inside the rotation matrix in a much simpler way. The minimization \eqref{eq:opt-rotation-robust} is carried out using a sequential least squares programming (SLSQP) solver\footnote{\url{https://docs.scipy.org/doc/scipy/reference/optimize.html}} in combination with a root-finding software package \cite{kraft1988software}.  The SLSQP minimizer found at the previous EM iteration is used to initialize the current EM iteration. At the start of EM, the closed-form method of \cite{horn1987closed} is used to initialize the rotation.

Algorithm~\ref{algo:trff} uses Algorithm~\ref{algo:rff} for intialization at $t=1$. Then, at the following time steps, Algorithm~\ref{algo:rff} is used to compute the frontalized landmarks, which are then used to recursively estimate the parameters of the posterior \eqref{eq:posterior-lds} and the Kalman gain matrix \eqref{eq:kalman-rec}. Note however, that the \textbf{M-non-rigid-step} of Algorithm~\ref{algo:rff} is not necessary because the shape embedding $\svect$ is treated as a Gaussian variable. Instead, the \textbf{rigid-pose} step of Algorithm~\ref{algo:trff} uses the shape parameters evaluated at the previous time step. The value of $\alpha$ was empirically estimated and set to 0.06 in all the experiments. The covariance matrices $\Gammamat_S$ and $\Gammamat_V$ were experimentally evaluated from the results obtained with Algorithm~\ref{algo:rff} on a large dataset of faces.

\addnote[simulations]{2}{
In all the experiments we used the \ac{3DFA} method of \cite{bulat2016two}, as already mentioned in Section~\ref{sec:related-work}. \ac{3DFA} predicts $J=68$ landmarks that may be prone to localization errors, i.e. outliers, in particular for non-frontal faces. We conducted a simulated experiment to show the effectiveness of the Student's t-distribution in the presence of outliers caused by \ac{3DFA}. For this purpose we considered a set of predicted 3D landmarks and we randomly simulated 500 rigid transformations. We added different noise types to the landmarks, as follows. For each trial we randomly split the landmarks into an inlier set and an outlier set. The inliers are corrupted by noise drawn from an anisotropic Gaussian distribution with a total variance $\lambda=0.0025$ (the landmark coordinates are normalized to lie in the interval $[0,1]$). The outlier noise is drawn from a uniform distribution whose volume is $1.5^3$. We tested the following distributions: an isotropic Gaussian distribution (Horn), a full-covariance Gaussian distribution (Gen-Horn), a mixture distribution with a Gaussian component and a uniform component (GUM-EM), and the generalized Student's t-distribution used in the paper (GStudent-EM). Horn is named after the author of the well-known closed-form solution for estimating scale, rotation and translation between two 3D point sets \cite{horn1987closed}. The plots of Figure~\ref{fig:simulations} display the root mean-square error (RMSE) over 500 trials and for an increasing percentage of outliers (from 0\% to 60\%).
}

The proposed method also requires the parameters of an already trained deformable shape model, namely $\Umat,\overline{\Mvect}$ in \eqref{eq:mesh-rec}. For this purpose we combined two publicly available face models, Basel Shape Model (BSM) \cite{paysan20093d} and Facewarehouse \cite{cao2014facewarehouse}. 
BFM \cite{paysan20093d} consists of 
a training set $\mathcal{M}^I= \{\Mvect_m^I\}_{m=1}^{m=M^I}$ of $M^I=200$ face scans of different identities. Each face in the dataset is frontally viewed and with a neutral expression. Each scan consists of a triangulated mesh composed of $N=53490$ vertices. Both the vertices and the edges of the meshes are registered. Facewarehouse \cite{cao2014facewarehouse} consists of a training set $\mathcal{M}^E= \{\Mvect_m^E\}_{m=1}^{m=M^E}$ of $M^E=7050$ face scans that correspond to 150 identities and 47 expressions, with the same number of vertices $N$ as BFM. The subjects were instructed to look frontally to the camera and to mimic 19 facial expressions as well as a neutral expression, from which 47 expressions were computed by linear blending. It should be noted that the 19 expressions correspond to emotions, e.g., mouth stretch, smile, anger, sadness, etc.

The identity and expression embeddings, $\svect^I$ and $\svect^E$, are of dimension $K^I=199$ and $K^E=29$, respectively. Therefore, a face mesh $\Vvect$ is reconstructed from a linear combination of identity and expression:
\begin{equation}
\Vvect = \Umat^I \svect^I + \overline{\Mvect}^I + \Umat^E \svect^E + \overline{\Mvect}^E
\end{equation}
The above formula can be plugged into \eqref{eq:mstep} whose minimization over $\svect^I$ and $\svect^E$ allows one to estimate the identity and expression embeddings of $\Vvect$. 

\addnote[identity-exp]{1}
{
In this paper we are interested in processing a face sequence. Since the identity remains unchanged during a sequence, the deformable face model just described is particularly interesting. Indeed, the identity embedding is estimating only once, at $t=1$, which yields the following formulas to be used for the subsequent faces, i.e. for $t=2\dots T$:
\begin{align}
\Umat &= \Umat^E \\
\overline{\Mvect} &= \Umat^I \svect^I_1 + \overline{\Mvect}^I + \overline{\Mvect}^E
\end{align}
}

\addnote[cpu-time]{1}{
The processing time for a $256\times 256$ face image is of $1.11$ seconds on an Intel(R), Xeon(R) W-2145,  3.70GHz CPU equipped with a Quadro RTX 4000 GPU. This time decomposes as follows: 3D landmark extraction (0.48~s), pose estimation (0.02~s), model fitting (0.23~s), depth map interpolation and face warping (0.38~s). 
}

\section{Face frontalization benchmark}
\label{sec:ff-benchmark}
We now evaluate and benchmark the proposed face frontalization formulation based on a score that measures the correlation between a frontalized face and a ground-truth frontal image of the same face.
For this purpose we use a dataset that contains 
pairs of frontal and profile videos of speaking participants for a large number of subjects. The evaluation consists of computing a metric between an image obtained by face frontalization of a profile view of a speaker, with an image containing a frontally-viewed face of the same speaker. It is important that the profile and frontal images are recorded with synchronized cameras in order to capture the same facial expression. Consequently, the proposed evaluation is based on image-to-image comparison. Several metrics were developed in the past for comparing two images, e.g. feature-based and pixel-based metrics. In this work we use the \ac{ZNCC} score between two image regions, a measure that has successfully been used for stereo matching, e.g. \cite{sun2002fast}. ZNCC is invariant to differences in brightness and contrast between the two images, due to the normalization with respect to mean and standard deviation. 

Let $R_f (h,v)\subset I_f$ be a region of size $H\times V$ whose center coincides with pixel location $(h,v)$ of a frontalized image $I_f$. Similarly, let $R_t (h,v)\subset I_t$ be a region of the same size and whose center coincides with pixel location $(h,v)$ of a ground-truth image $I_t$. The ZNCC score between these two regions writes:
\begin{align}
\label{eq:zncc}
\mathrm{ZNCC} & (h,v, \delta h\pri, \delta v\pri) = \\
  \max_{\delta h, \delta v} & \bigg\{ \frac{\cov [R_f (h,v),R_t (h+\delta h, v+ \delta v)]}{\var [R_f (h,v)]^{1/2} \var [R_t (h+\delta h,v+\delta v)]^{1/2}  } \bigg\}, \nonumber
\end{align}
where $\cov[\cdot, \cdot]$ is the centered covariance between the two regions, $\var [\cdot]$ is the centered variance of a region, $\delta h$ and $\delta v$ are horizontal and vertical shifts, and $\delta h\pri$ and $\delta v\pri$ are the horizontal and vertical shifts that maximize the ZNCC score. ZNCC lies in the interval $[0,1]$.

\begin{figure*}[t!]
   \begin{center}
\subfloat[Faces recorded with the $30^\circ$ camera]{
\includegraphics[trim = 14cm 6cm 26cm 8cm,clip=true,keepaspectratio=true,width=0.24\columnwidth]{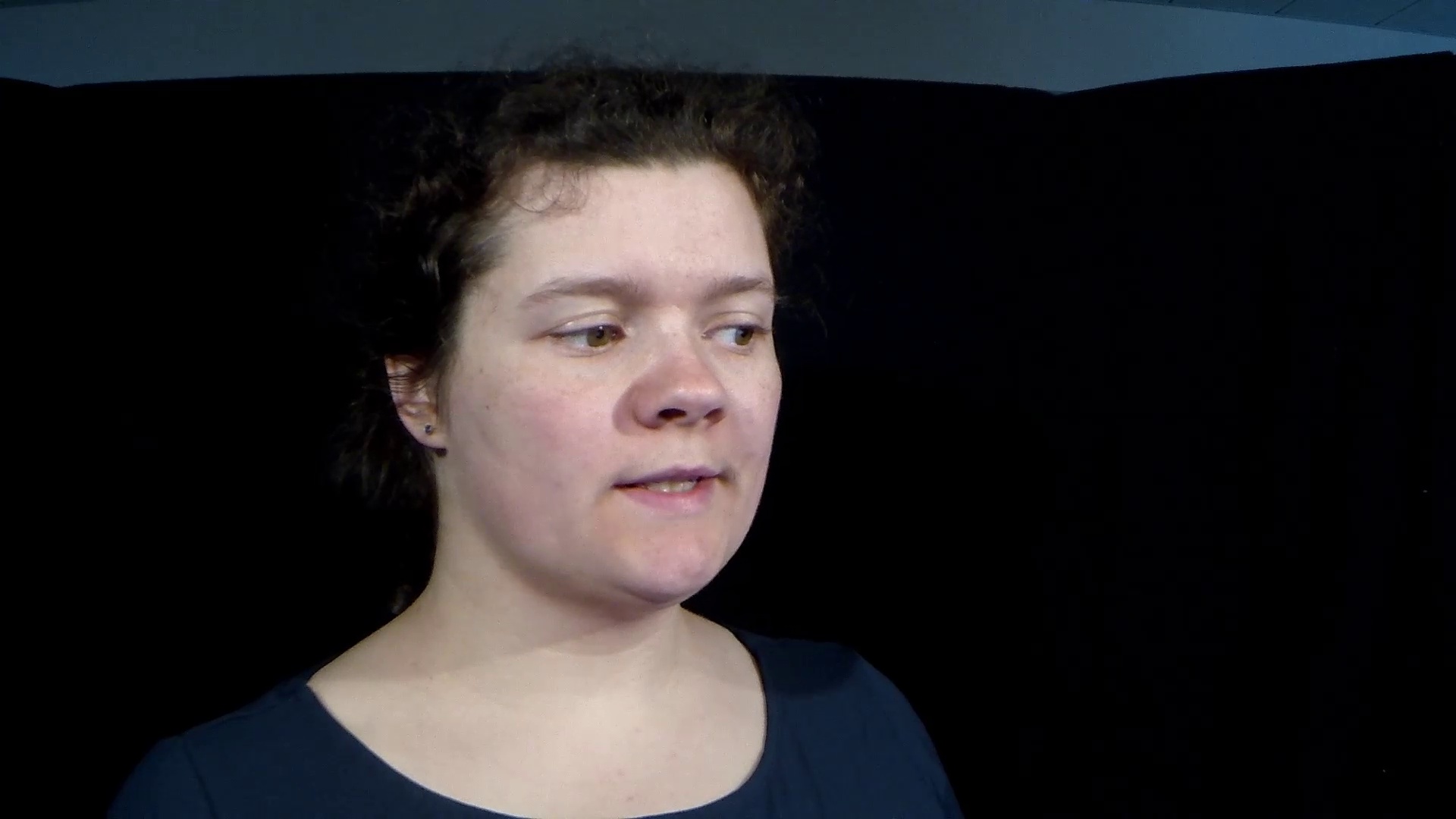}\hspace{1mm}
\includegraphics[trim = 14cm 6cm 26cm 8cm,clip=true,keepaspectratio=true,width=0.24\columnwidth]{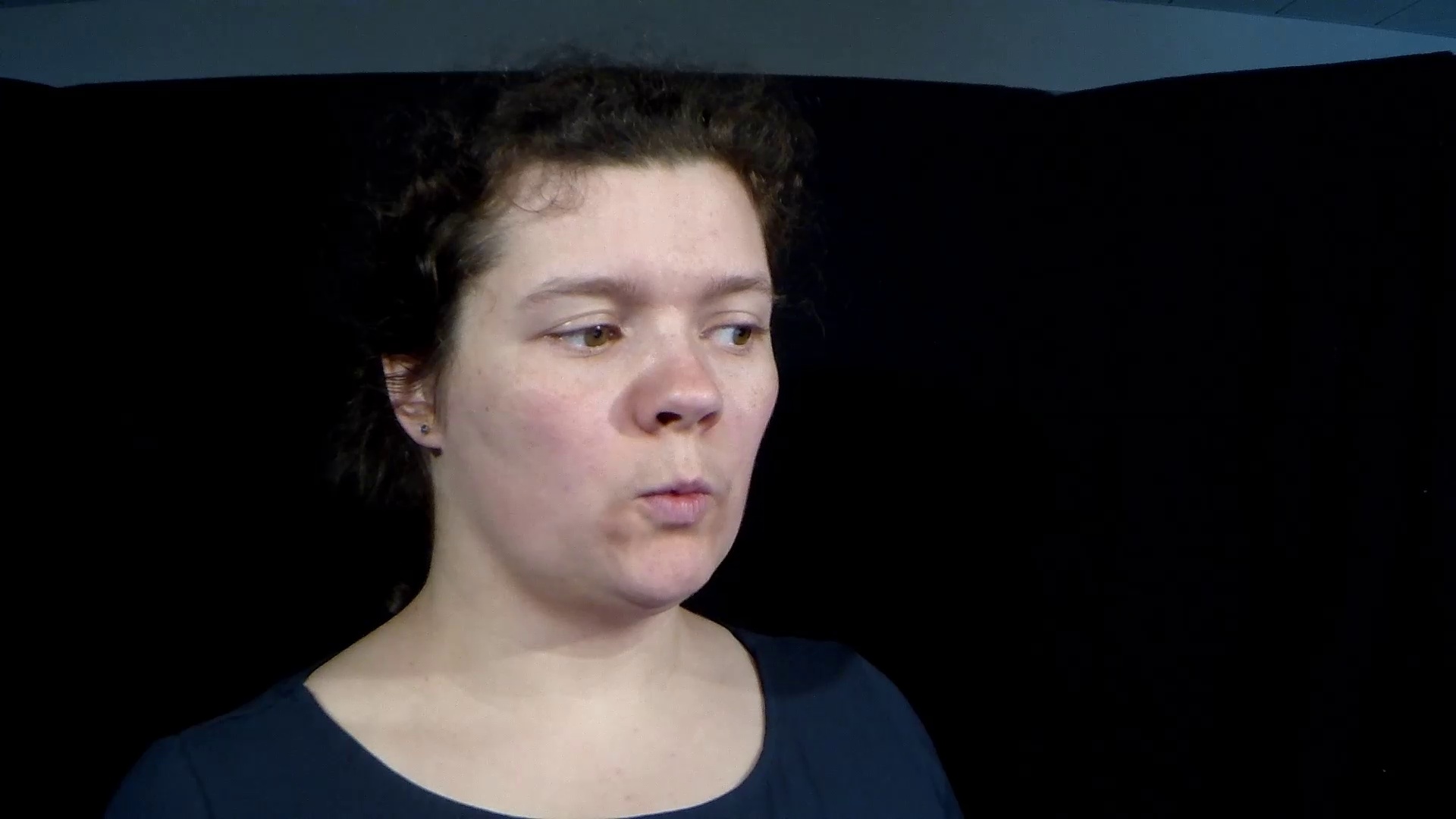}\hspace{1mm}
\includegraphics[trim = 14cm 6cm 26cm 8cm,clip=true,keepaspectratio=true,width=0.24\columnwidth]{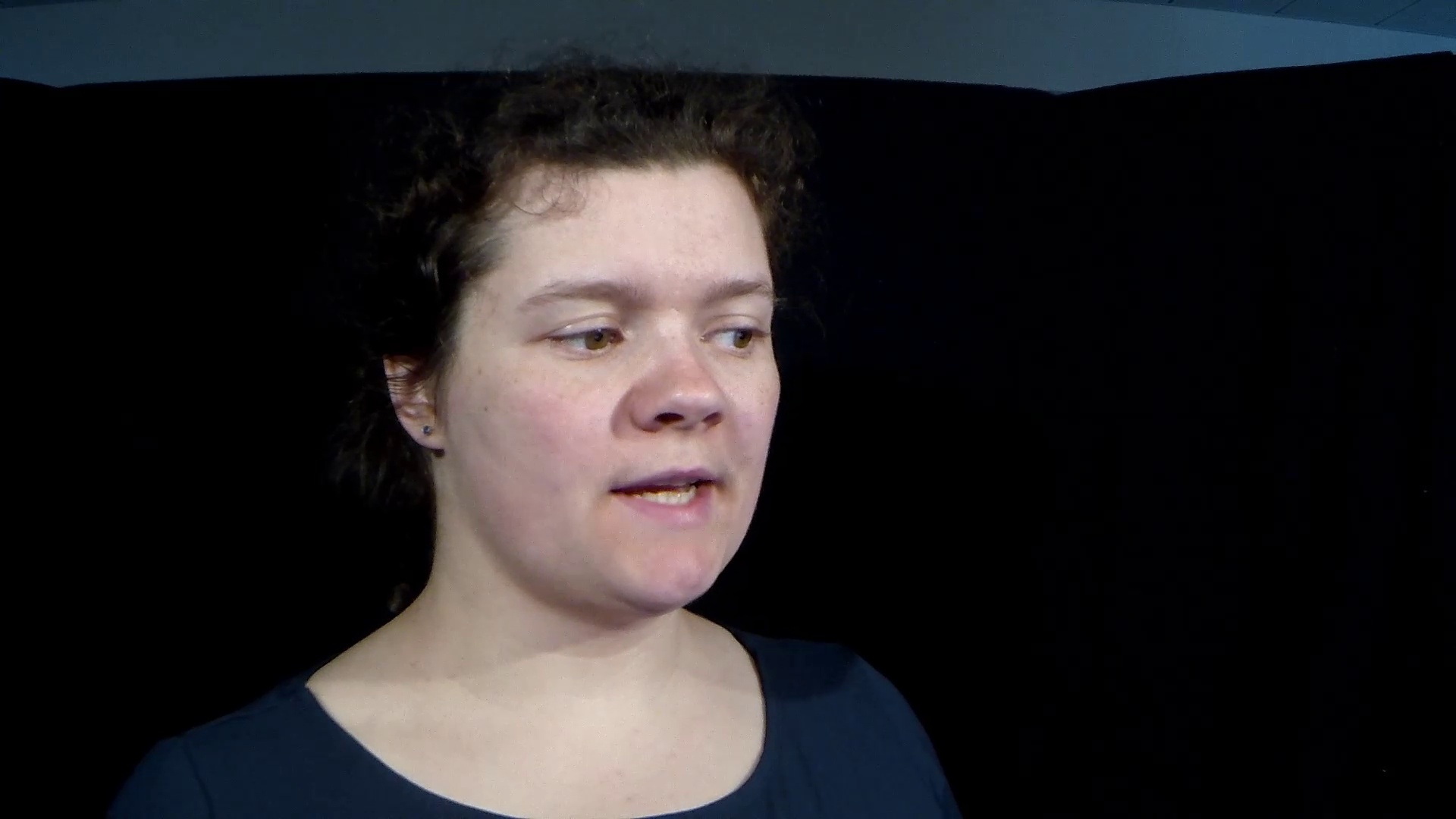}\hspace{1mm}
\includegraphics[trim = 14cm 6cm 26cm 8cm,clip=true,keepaspectratio=true,width=0.24\columnwidth]{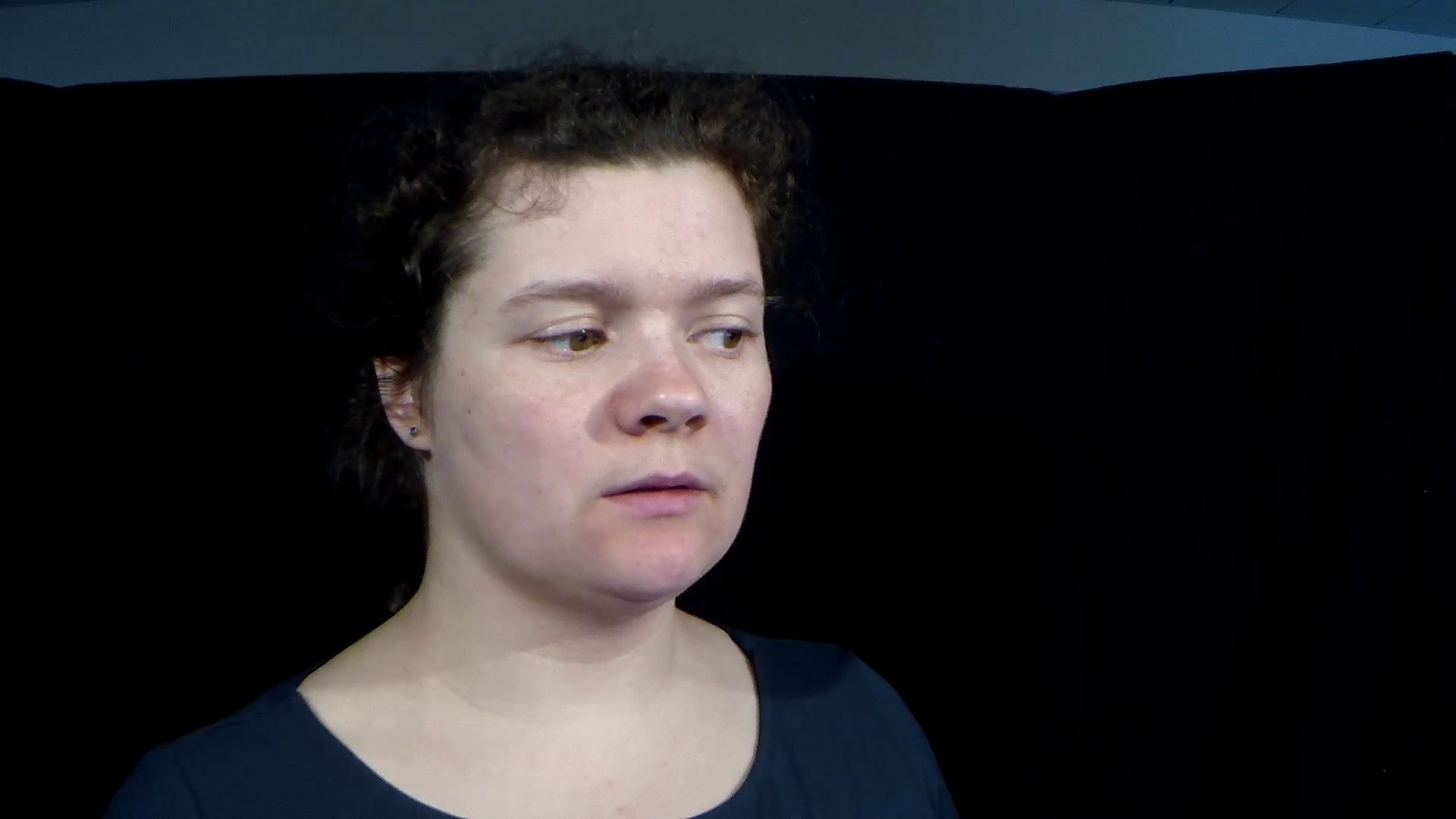}\hspace{1mm}
\includegraphics[trim = 9cm 2cm 27cm 8cm,clip,keepaspectratio=true,width=0.24\columnwidth]{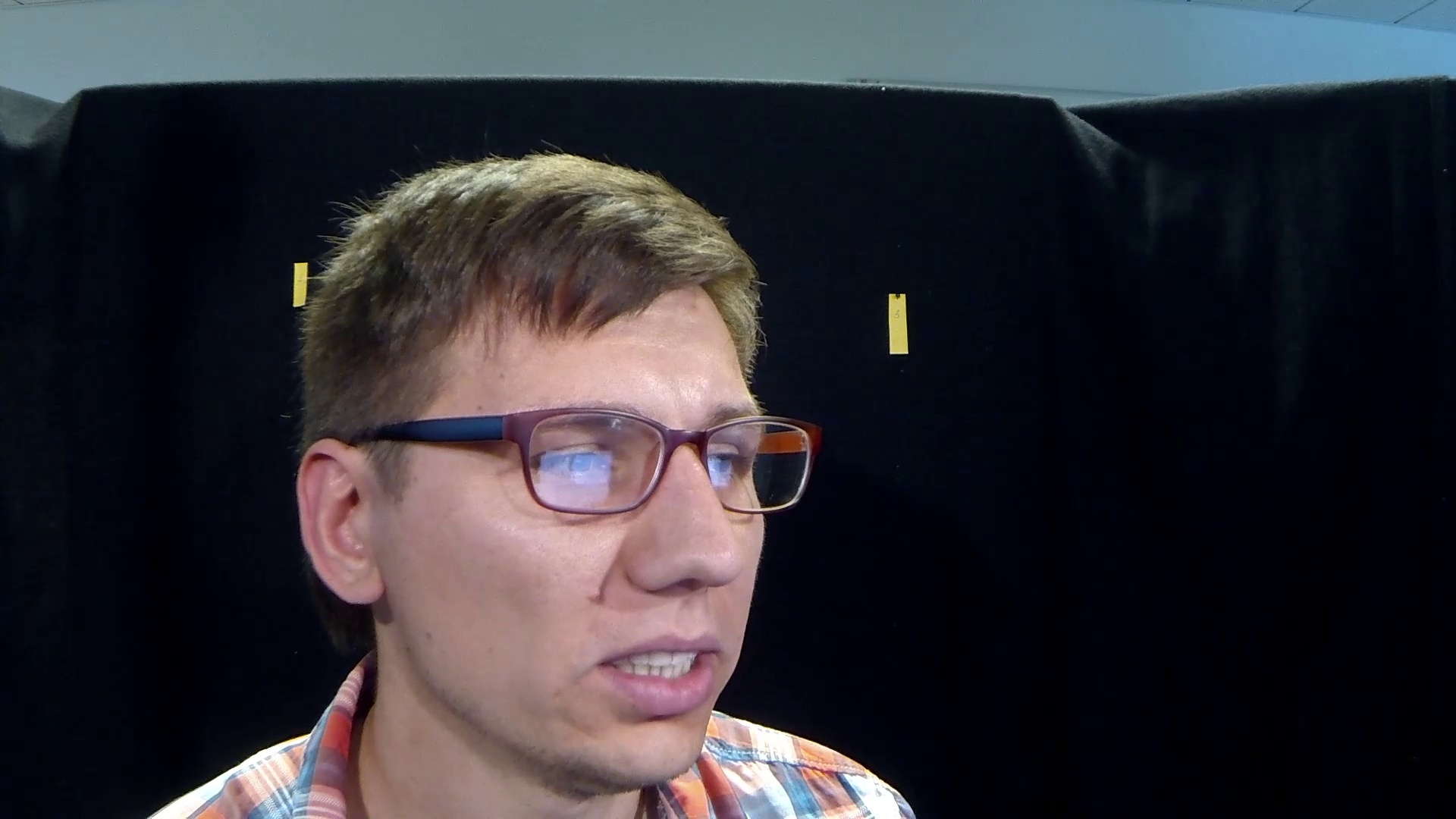}\hspace{1mm}
\includegraphics[trim = 9cm 2cm 27cm 8cm,clip,keepaspectratio=true,width=0.24\columnwidth]{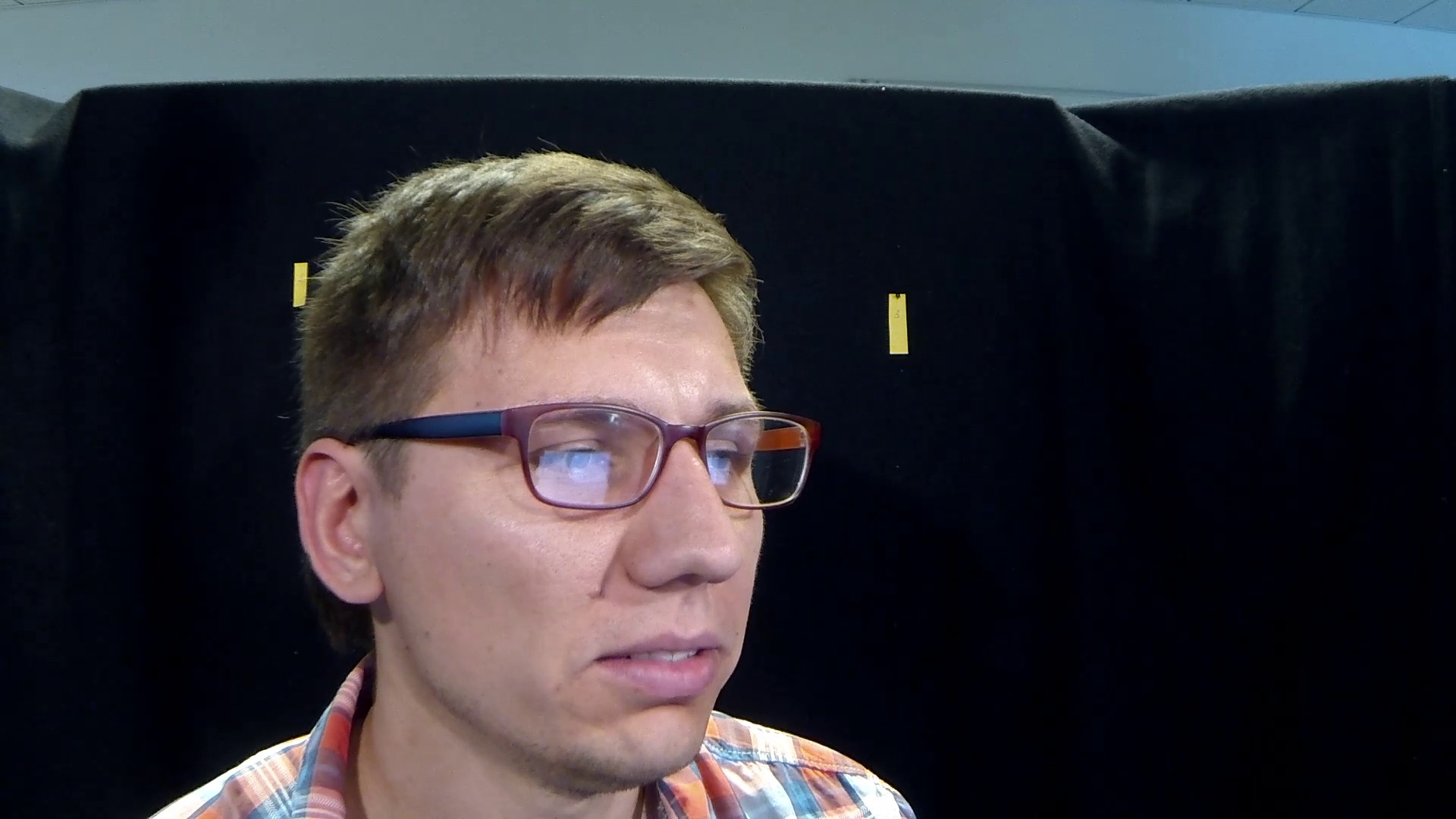}\hspace{1mm}
\includegraphics[trim = 9cm 2cm 27cm 8cm,clip,keepaspectratio=true,width=0.24\columnwidth]{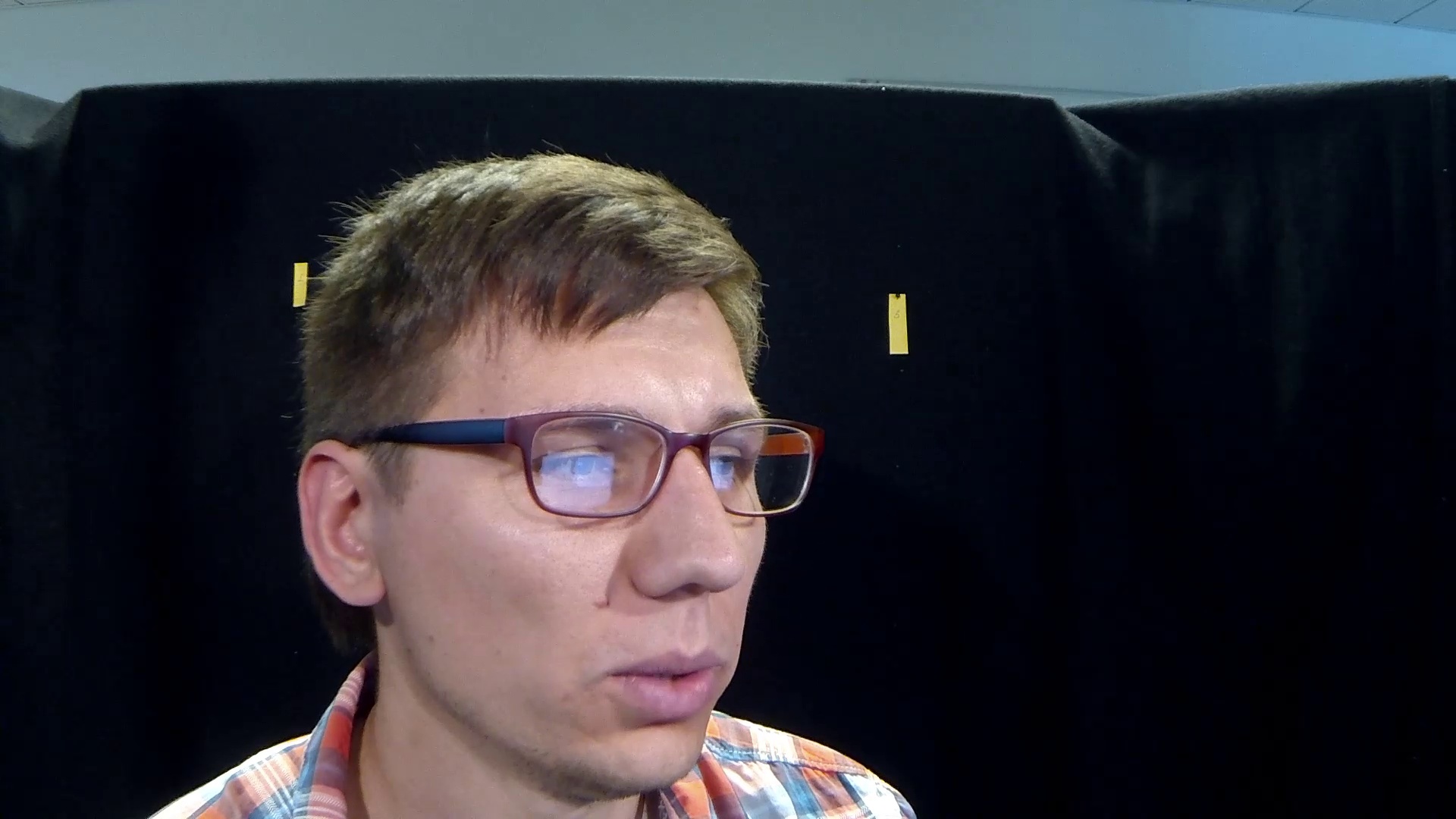}\hspace{1mm}
\includegraphics[trim = 9cm 2cm 27cm 8cm,clip,keepaspectratio=true,width=0.24\columnwidth]{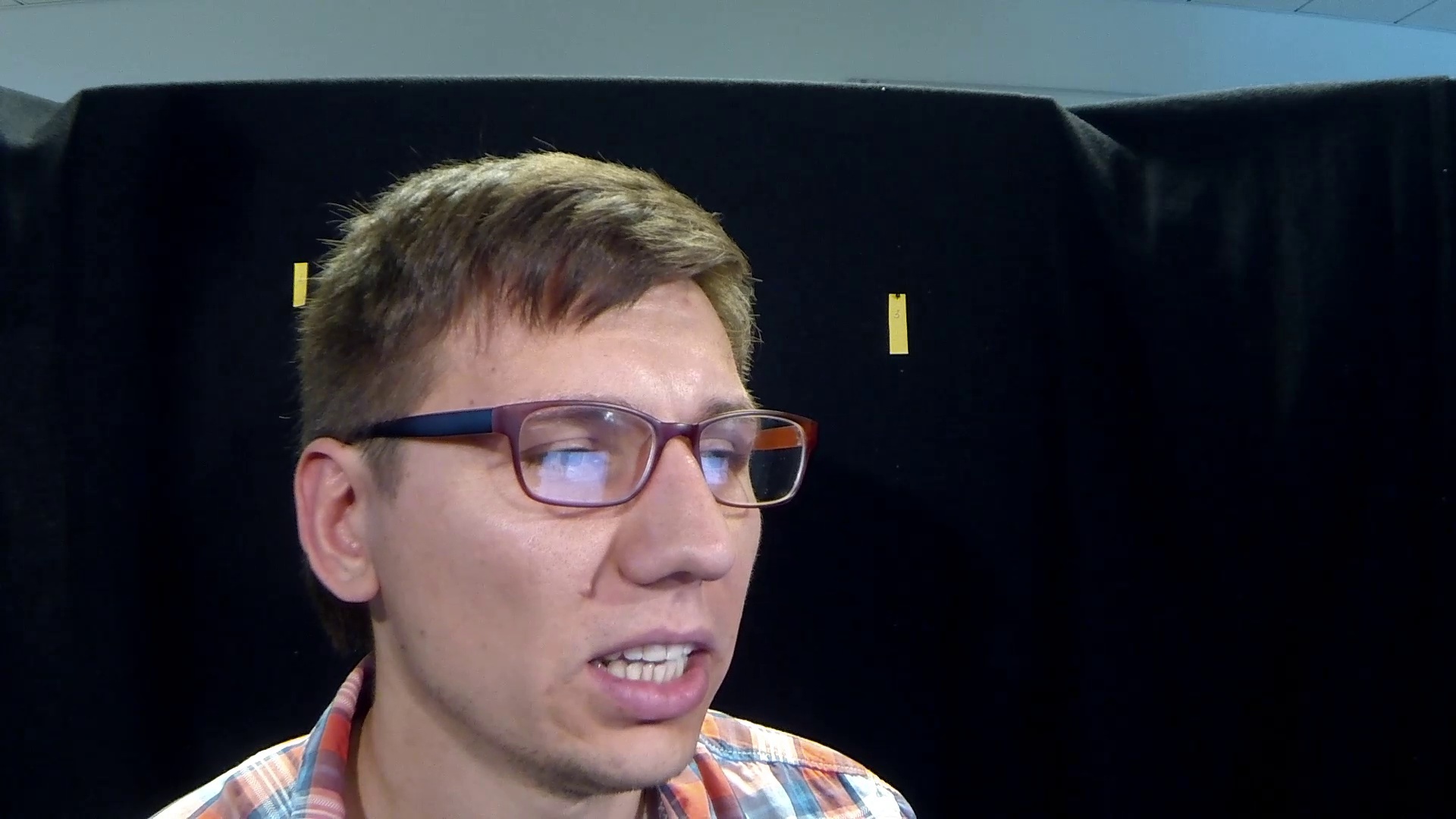}
}\\
\vspace{-3mm}
\subfloat[Faces recorded with the $0^\circ$ camera]{
\includegraphics[trim = 22cm 10cm 18cm 4cm,clip,keepaspectratio=true,width=0.24\columnwidth]{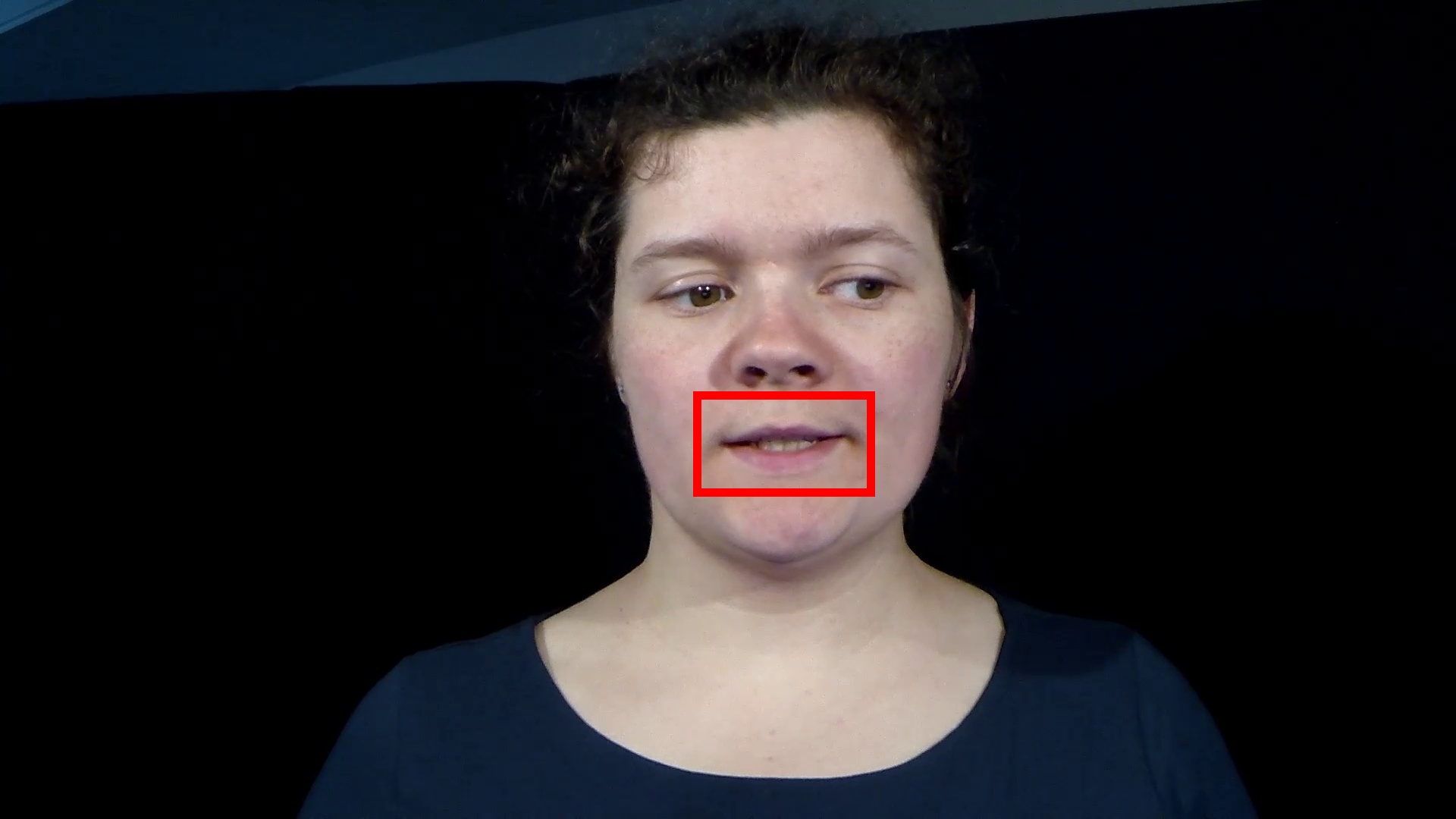}\hspace{1mm}
\includegraphics[trim = 22cm 10cm 18cm 4cm,clip,keepaspectratio=true,width=0.24\columnwidth]{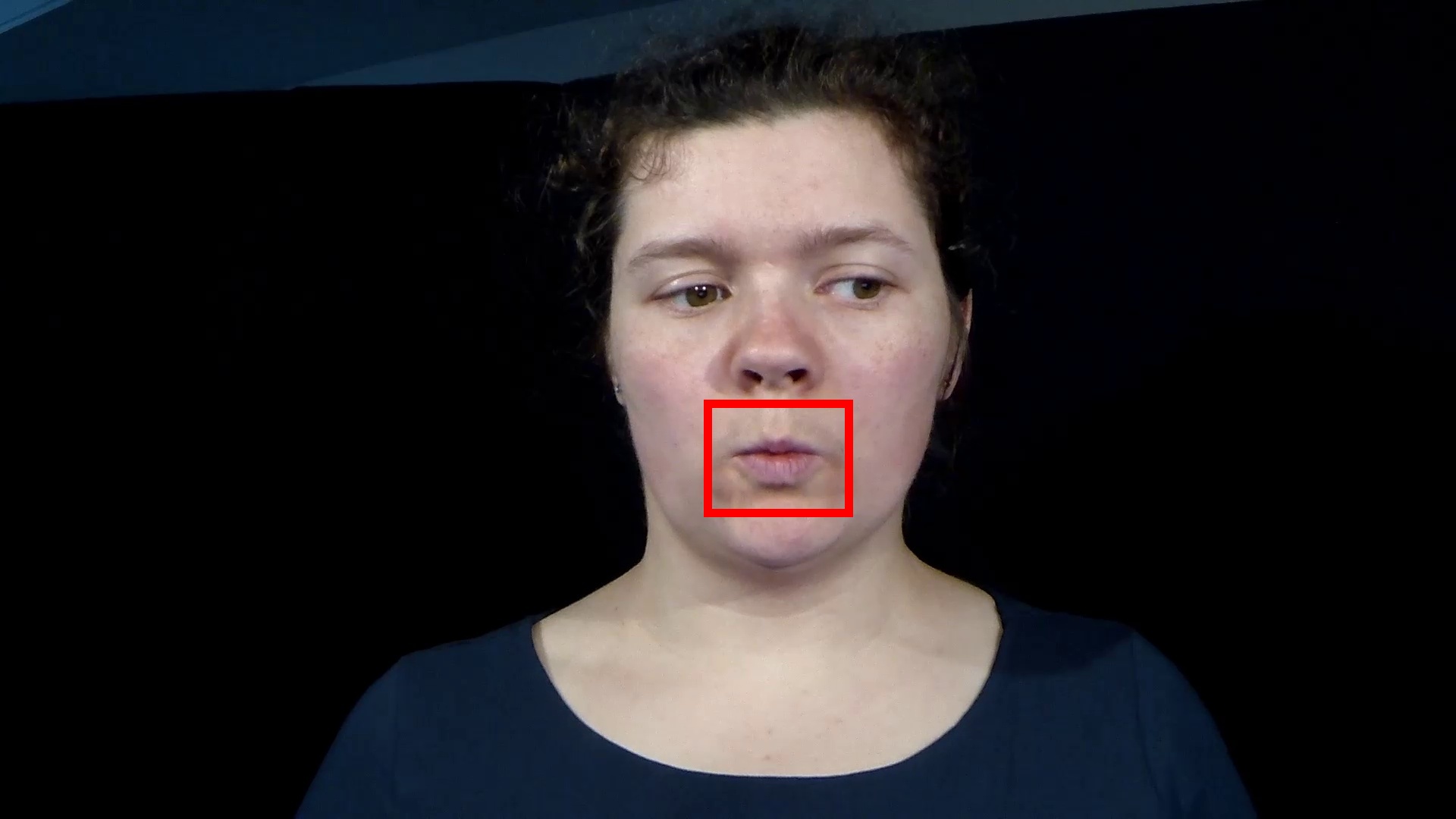}\hspace{1mm}
\includegraphics[trim = 22cm 10cm 18cm 4cm,clip,keepaspectratio=true,width=0.24\columnwidth]{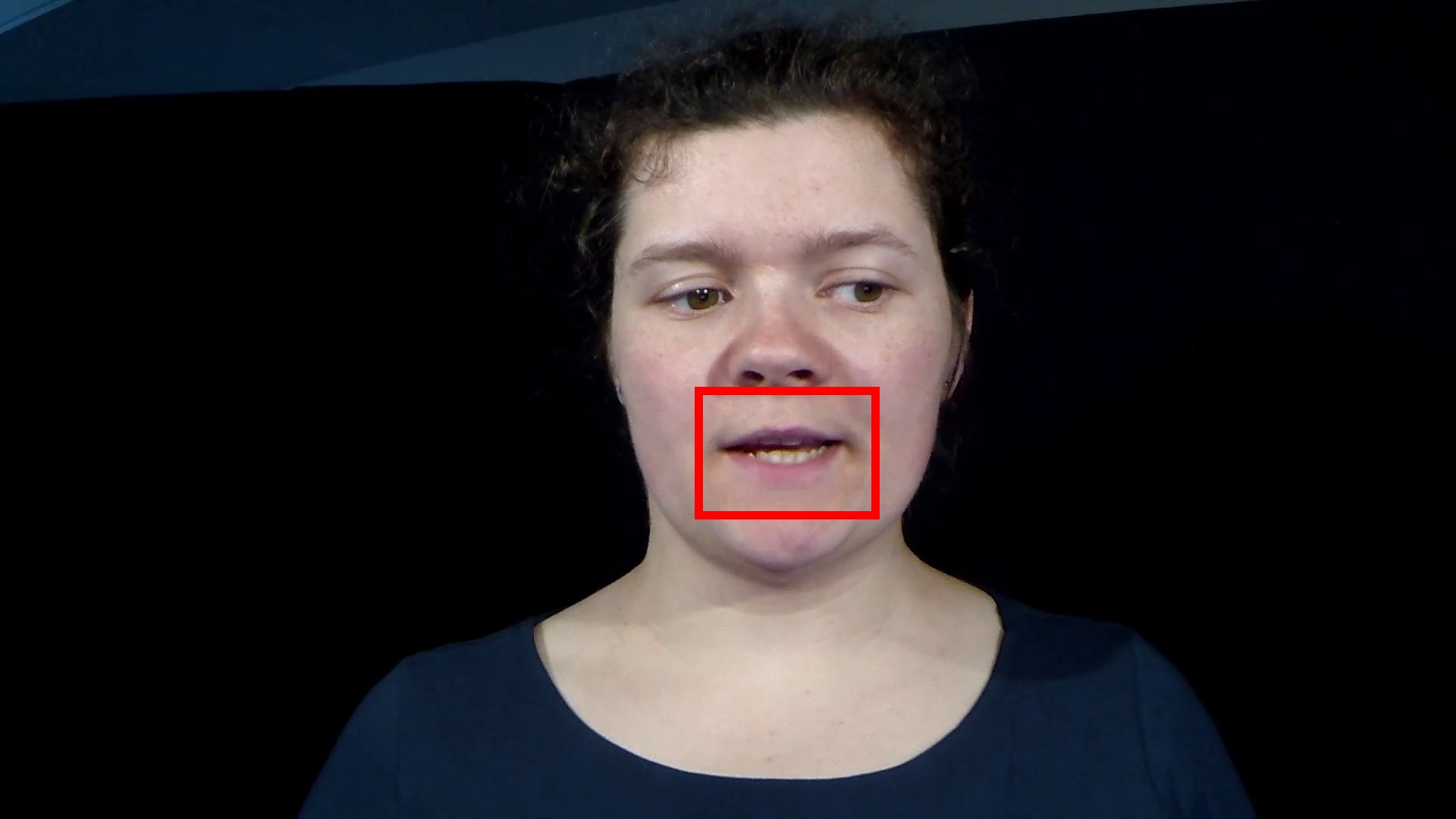}\hspace{1mm}
\includegraphics[trim = 22cm 10cm 18cm 4cm,clip,keepaspectratio=true,width=0.24\columnwidth]{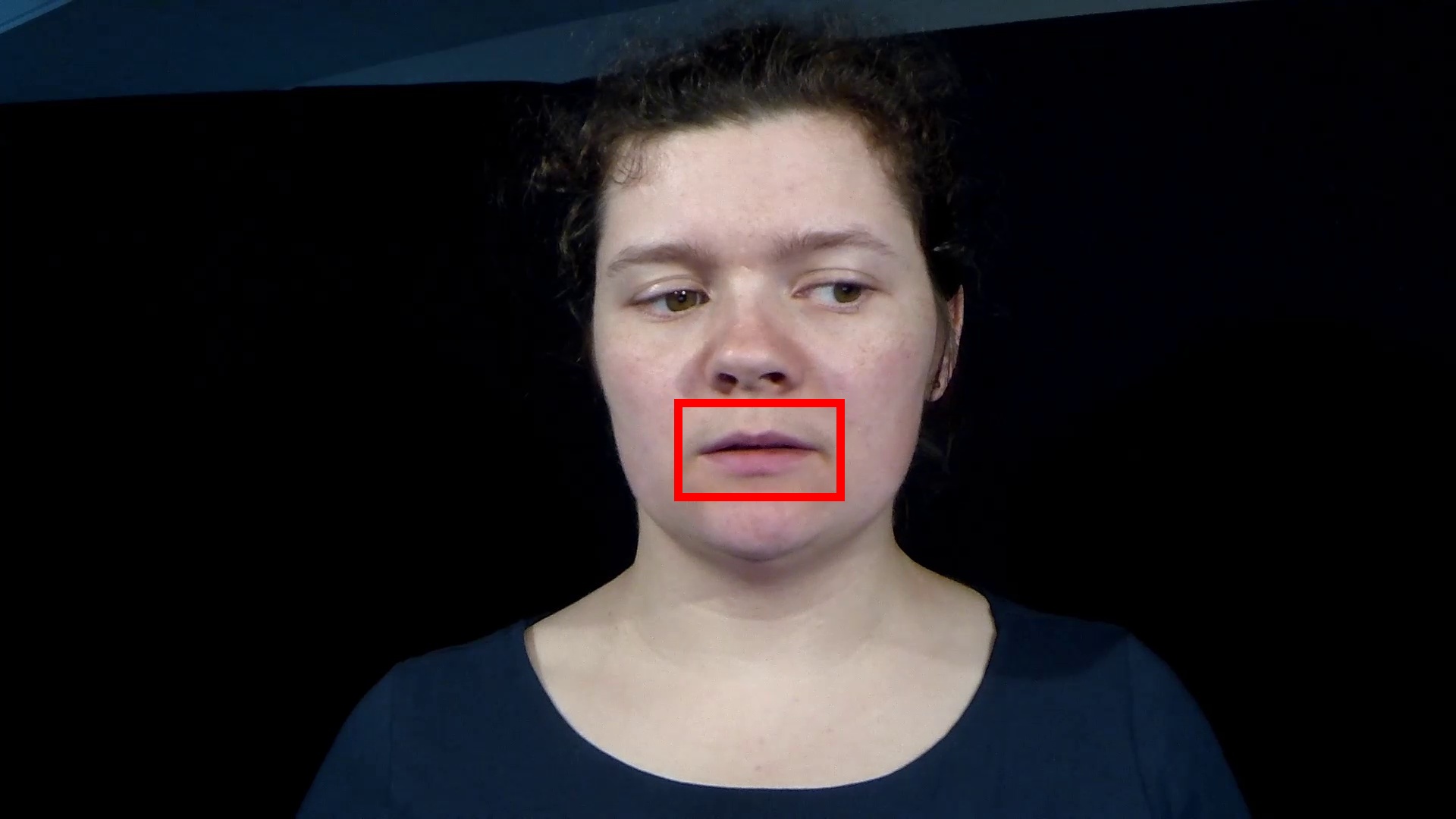}\hspace{1mm}
\includegraphics[trim = 19cm 3cm 19cm 9cm,clip,keepaspectratio=true,width=0.24\columnwidth]{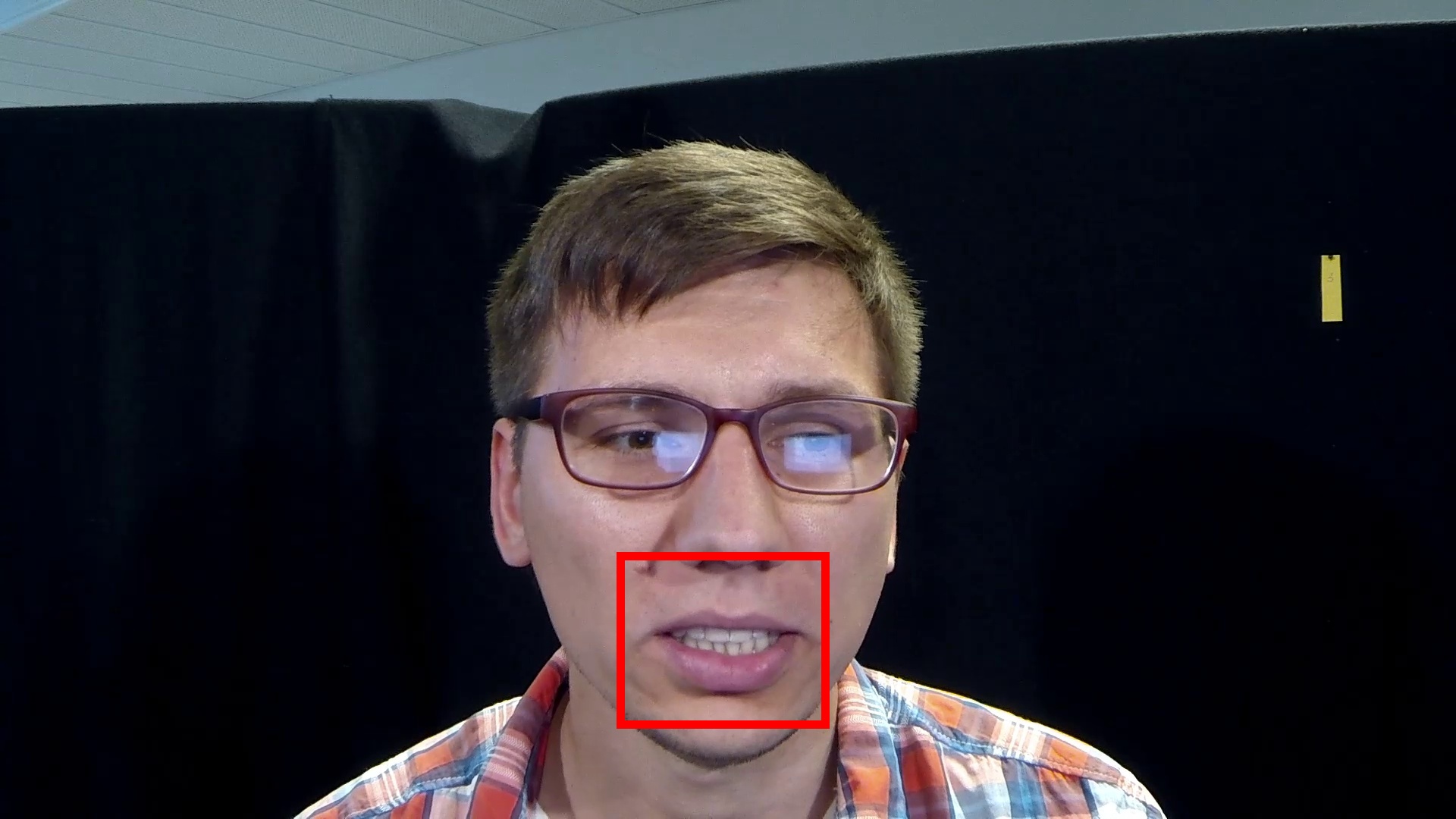}\hspace{1mm}
\includegraphics[trim = 19cm 3cm 19cm 9cm,clip,keepaspectratio=true,width=0.24\columnwidth]{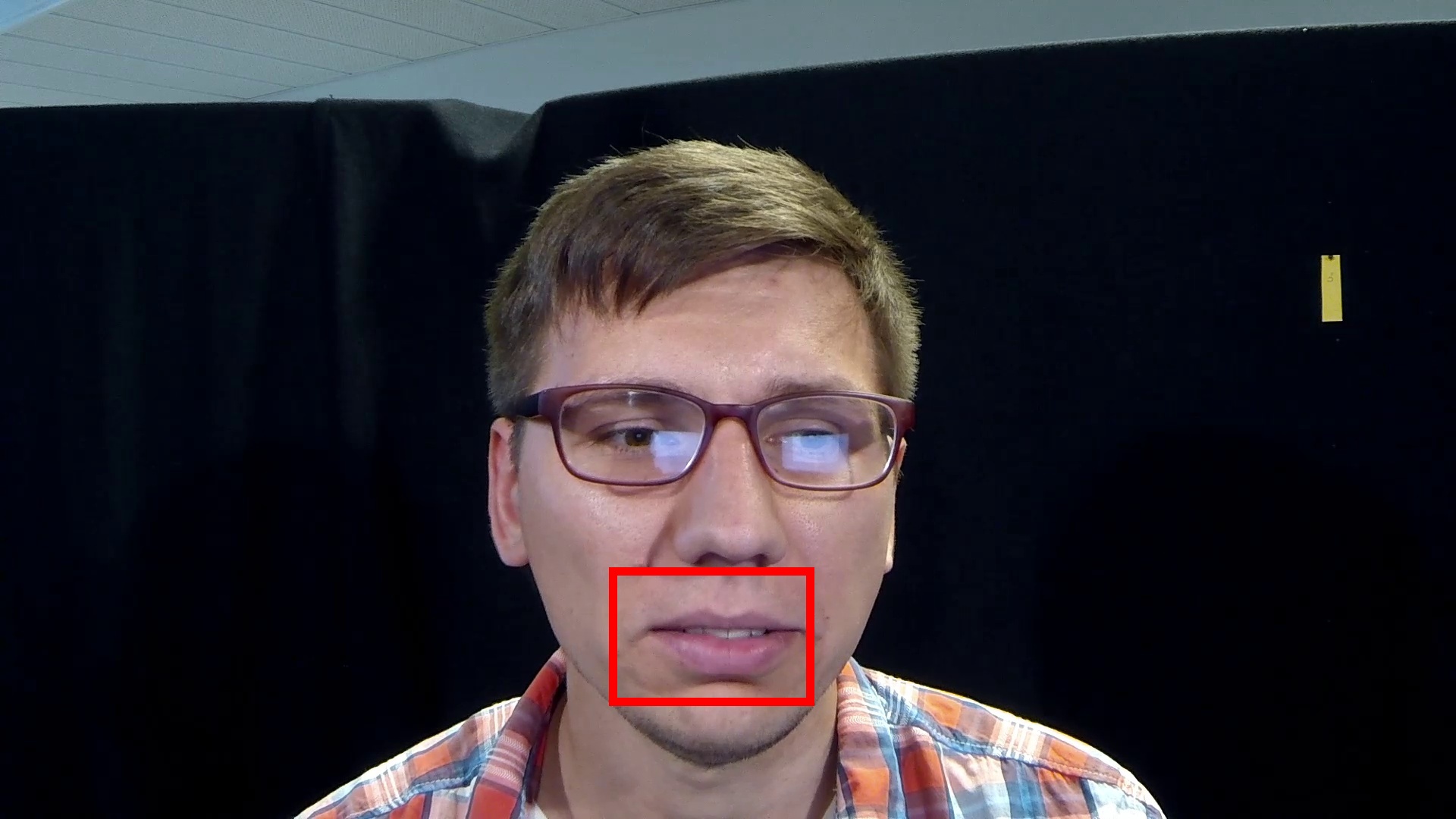}\hspace{1mm}
\includegraphics[trim = 19cm 3cm 19cm 9cm,clip,keepaspectratio=true,width=0.24\columnwidth]{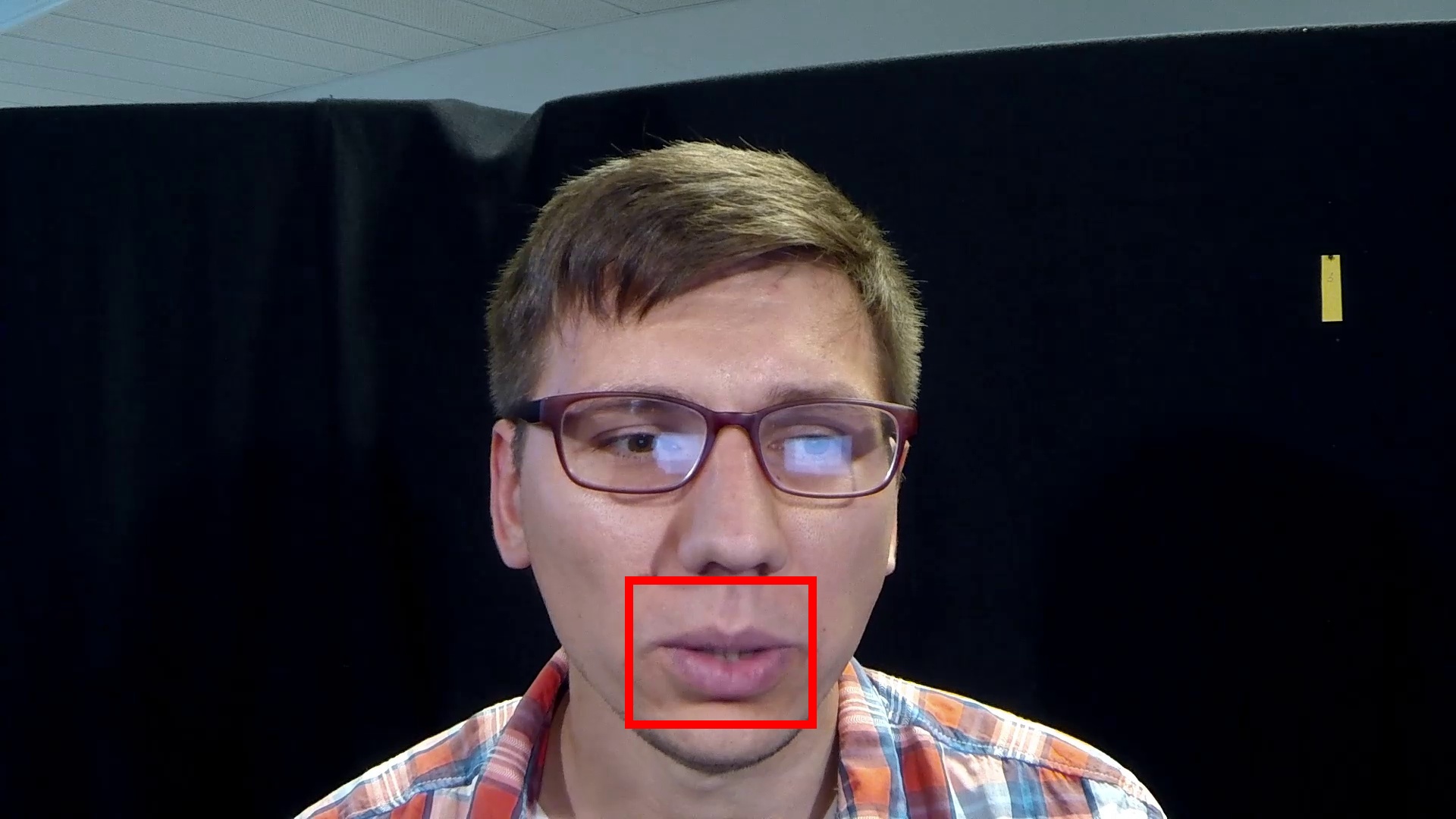}\hspace{1mm}
\includegraphics[trim = 19cm 3cm 19cm 9cm,clip,keepaspectratio=true,width=0.24\columnwidth]{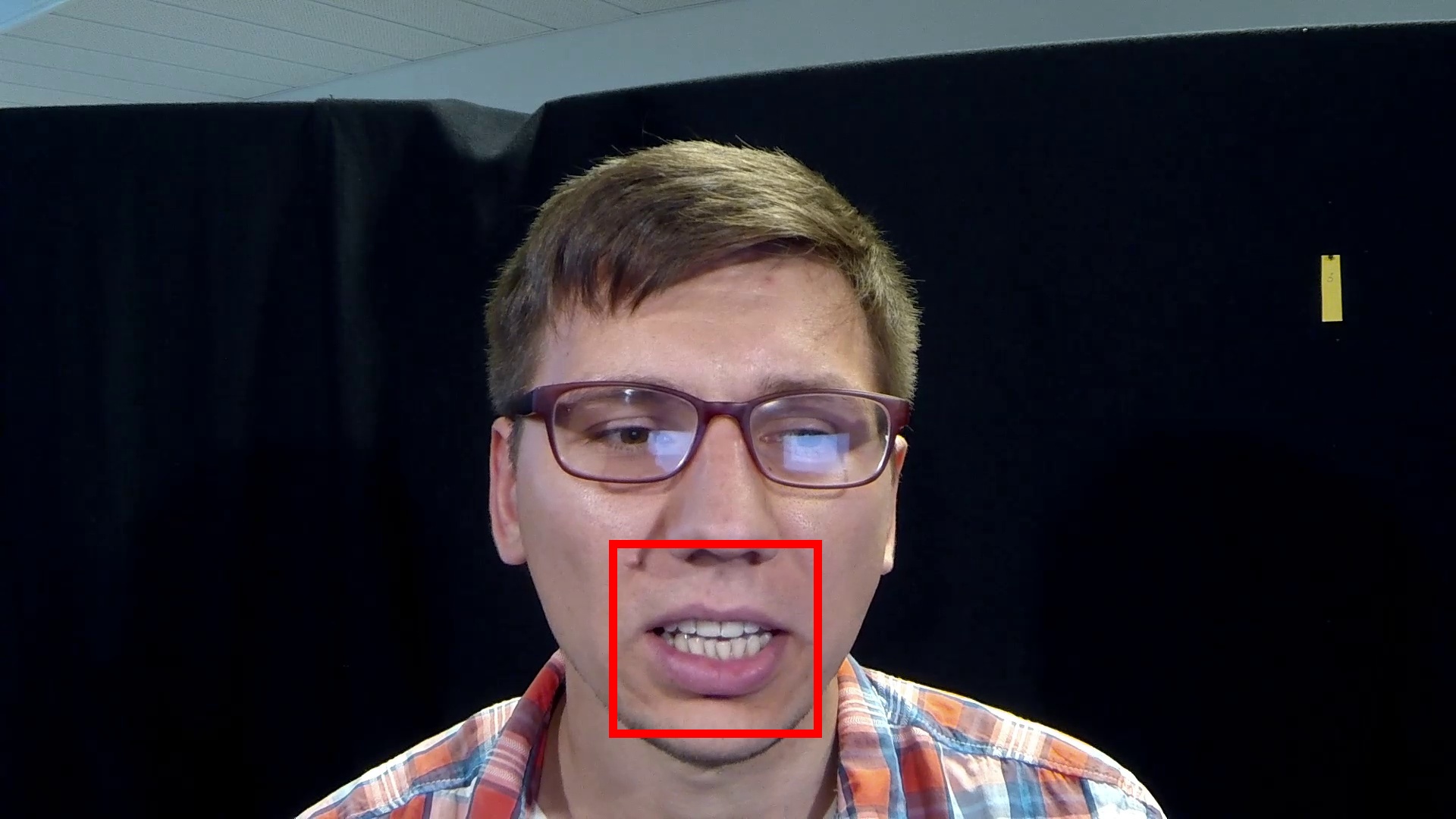}
}\\
\vspace{-3mm}
\subfloat[Proposed (self-occluded regions are displayed in white)]{
\includegraphics[trim = 20cm 6cm 20cm 8cm,clip,keepaspectratio=true,width=0.24\columnwidth]{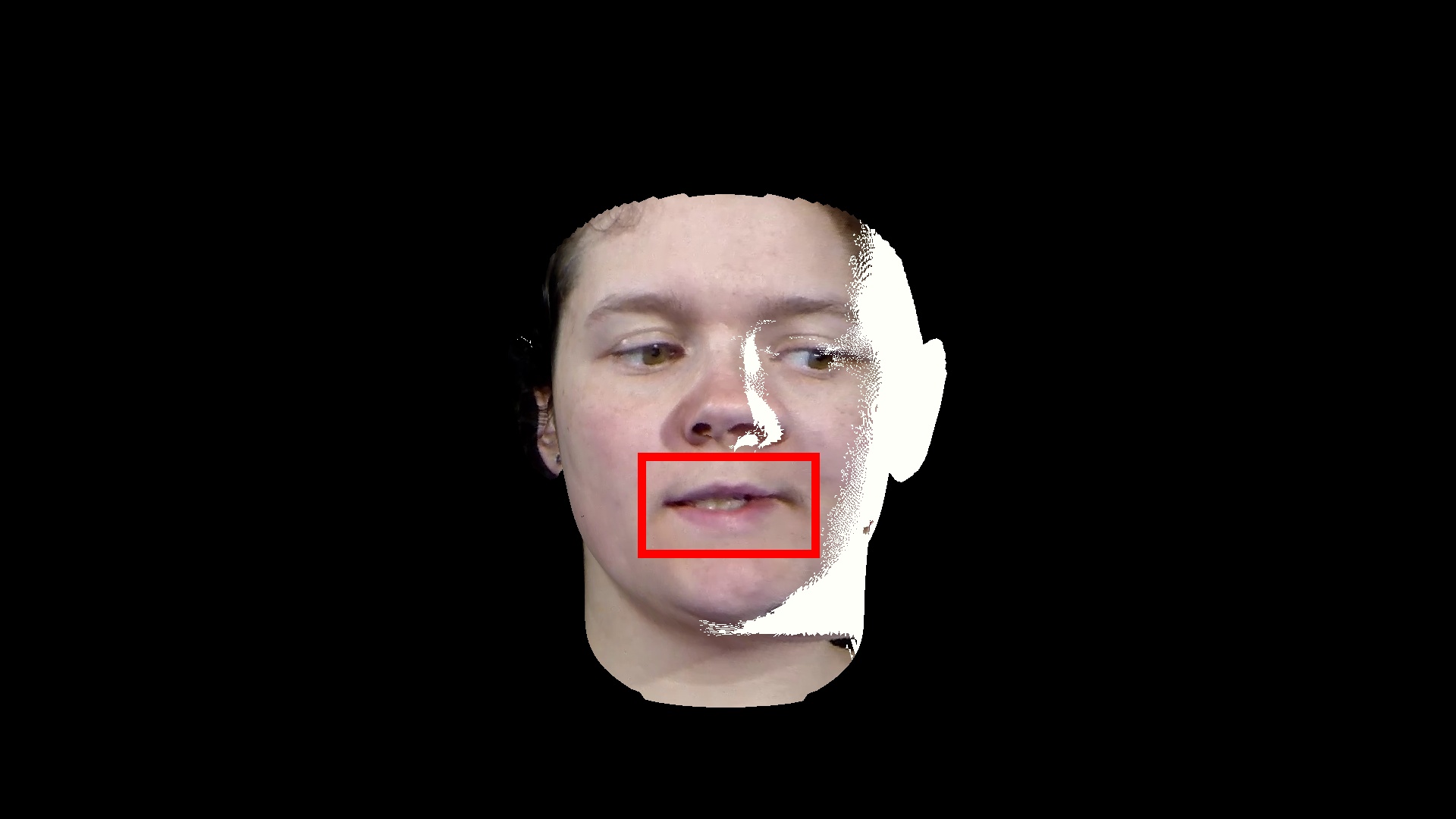}\hspace{1mm}
\includegraphics[trim = 20cm 6cm 20cm 8cm,clip,keepaspectratio=true,width=0.24\columnwidth]{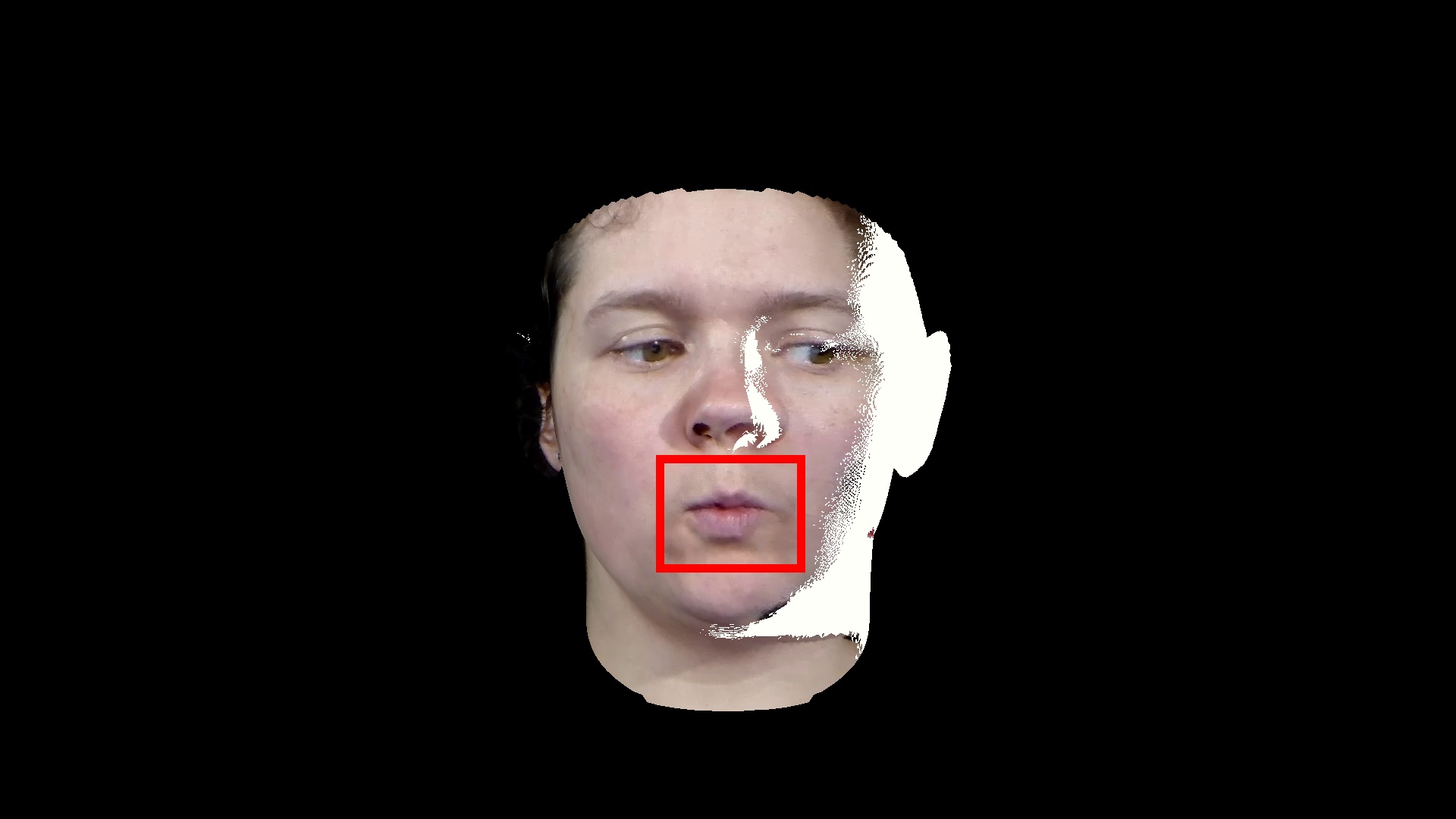}\hspace{1mm}
\includegraphics[trim = 20cm 6cm 20cm 8cm,clip,keepaspectratio=true,width=0.24\columnwidth]{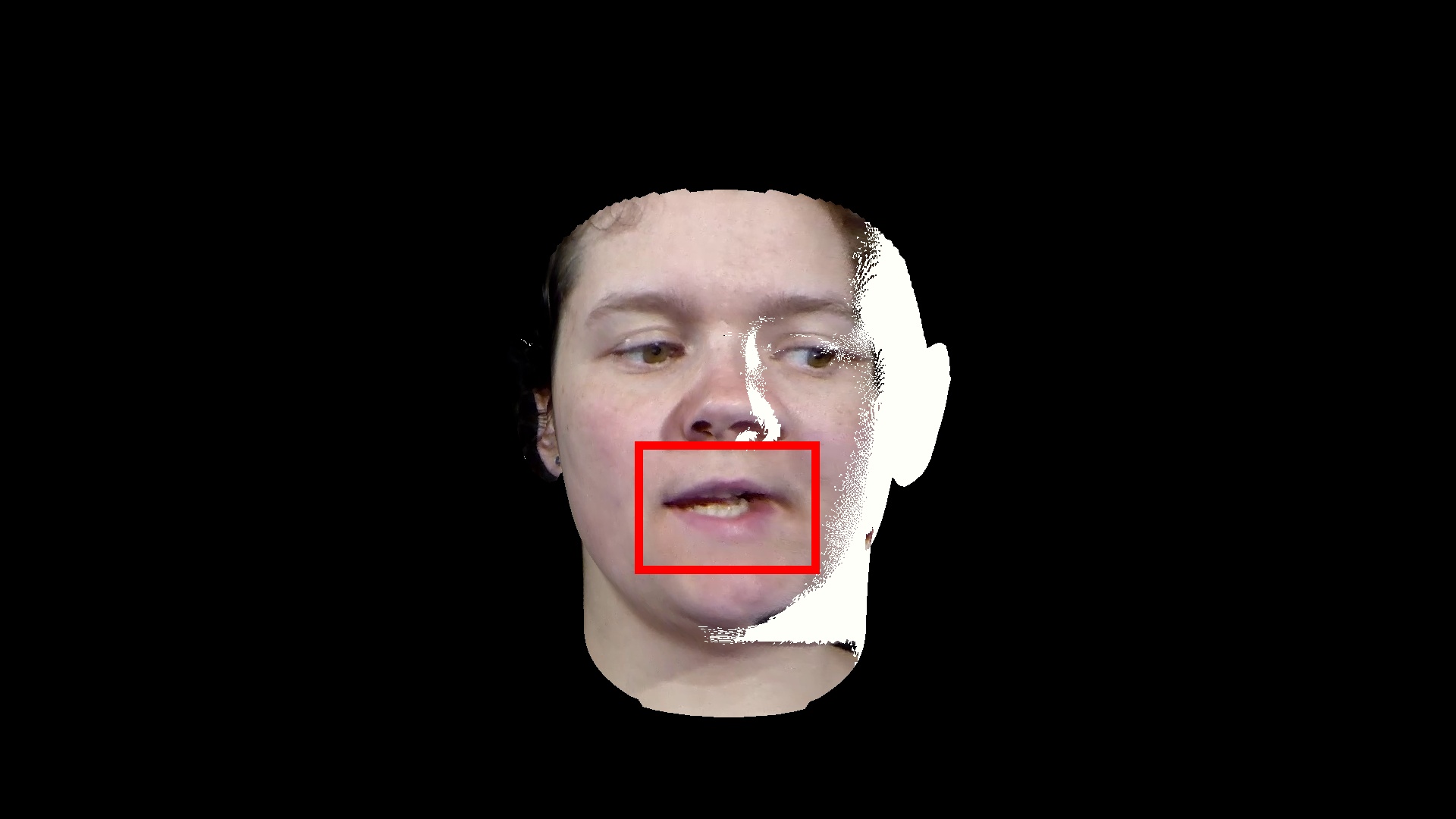}\hspace{1mm}
\includegraphics[trim = 20cm 6cm 20cm 8cm,clip,keepaspectratio=true,width=0.24\columnwidth]{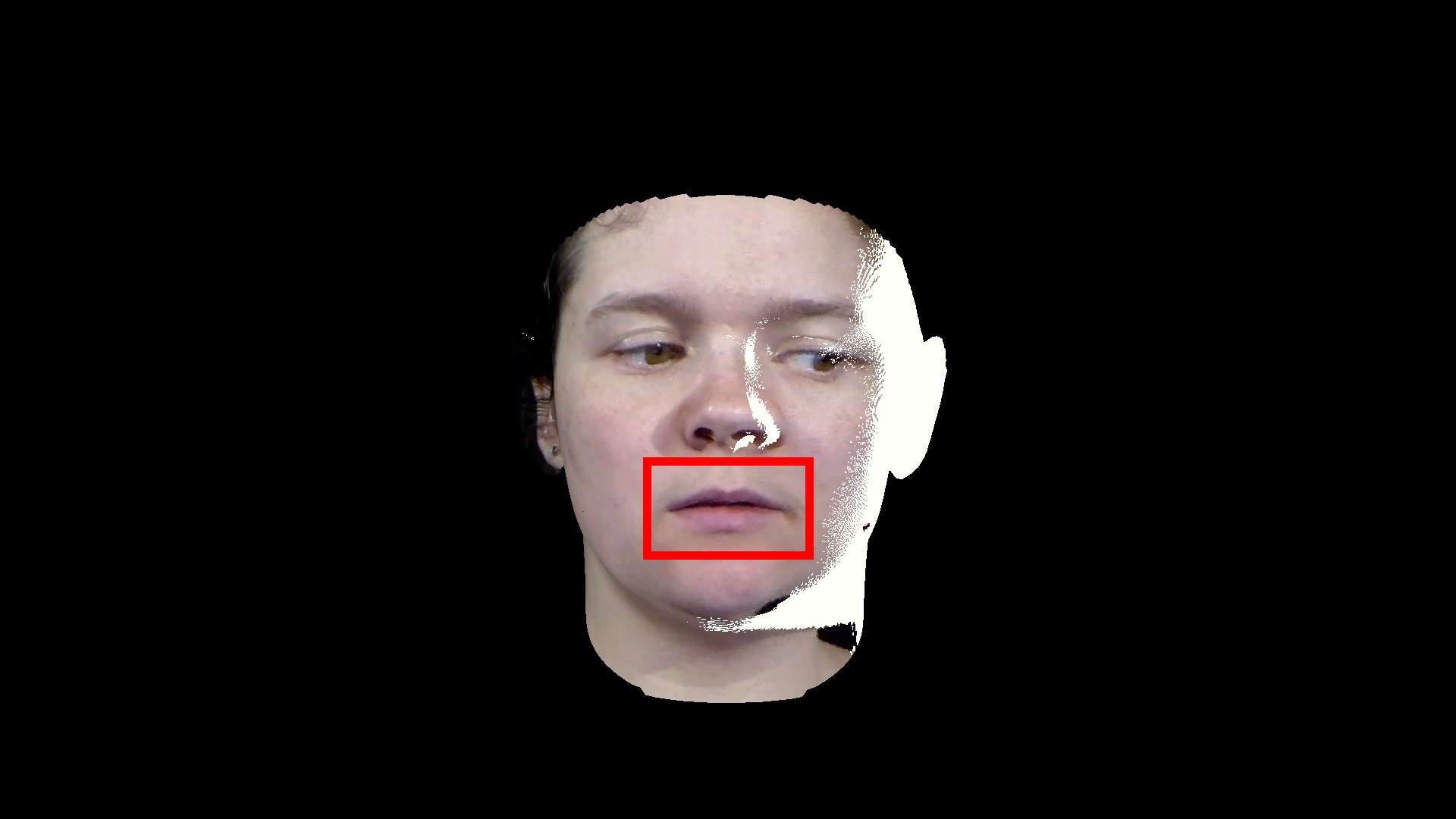}\hspace{1mm}
\includegraphics[trim = 20cm 6cm 20cm 8cm,clip,keepaspectratio=true,width=0.24\columnwidth]{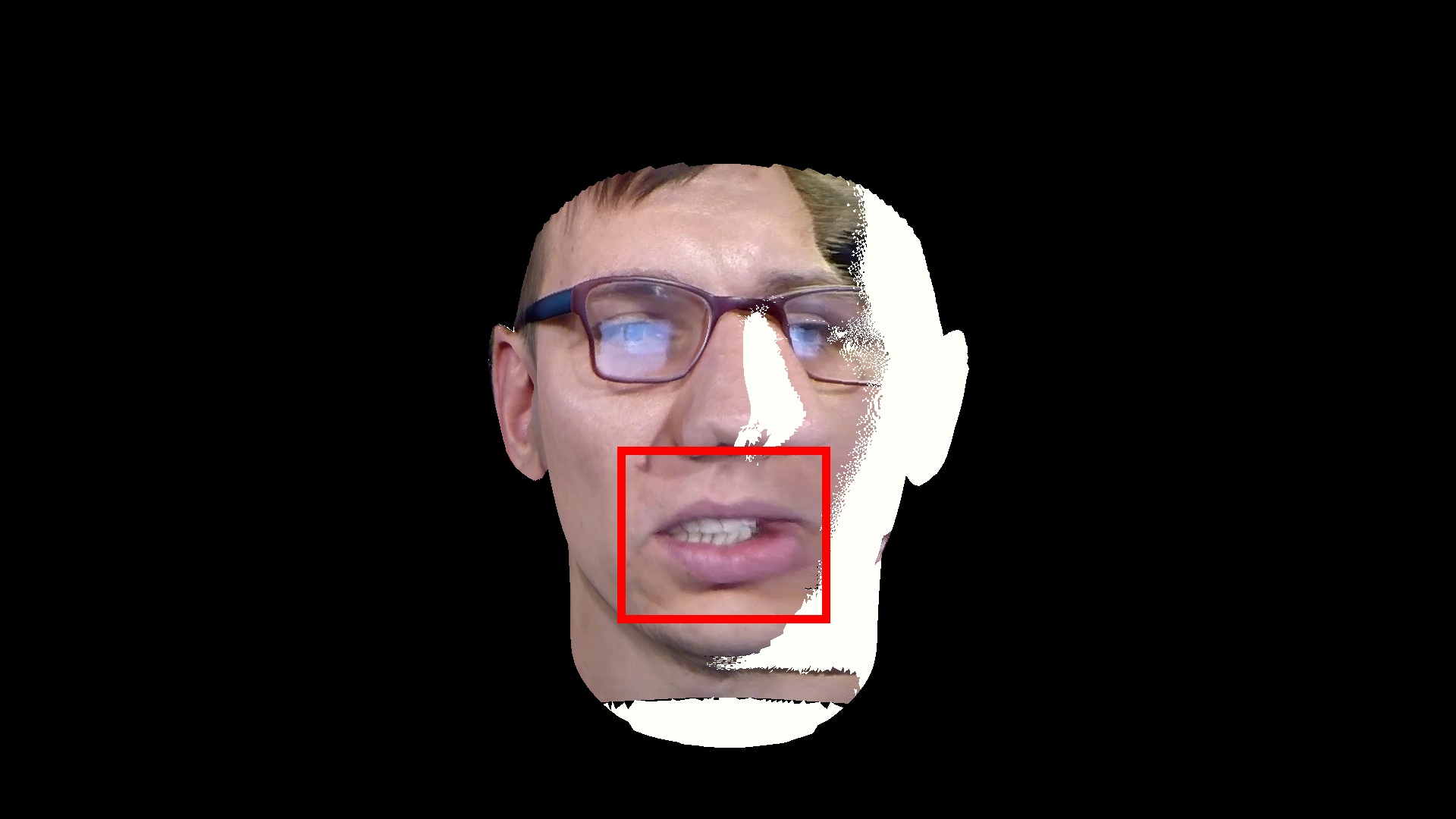}\hspace{1mm}
\includegraphics[trim = 20cm 6cm 20cm 8cm,clip,keepaspectratio=true,width=0.24\columnwidth]{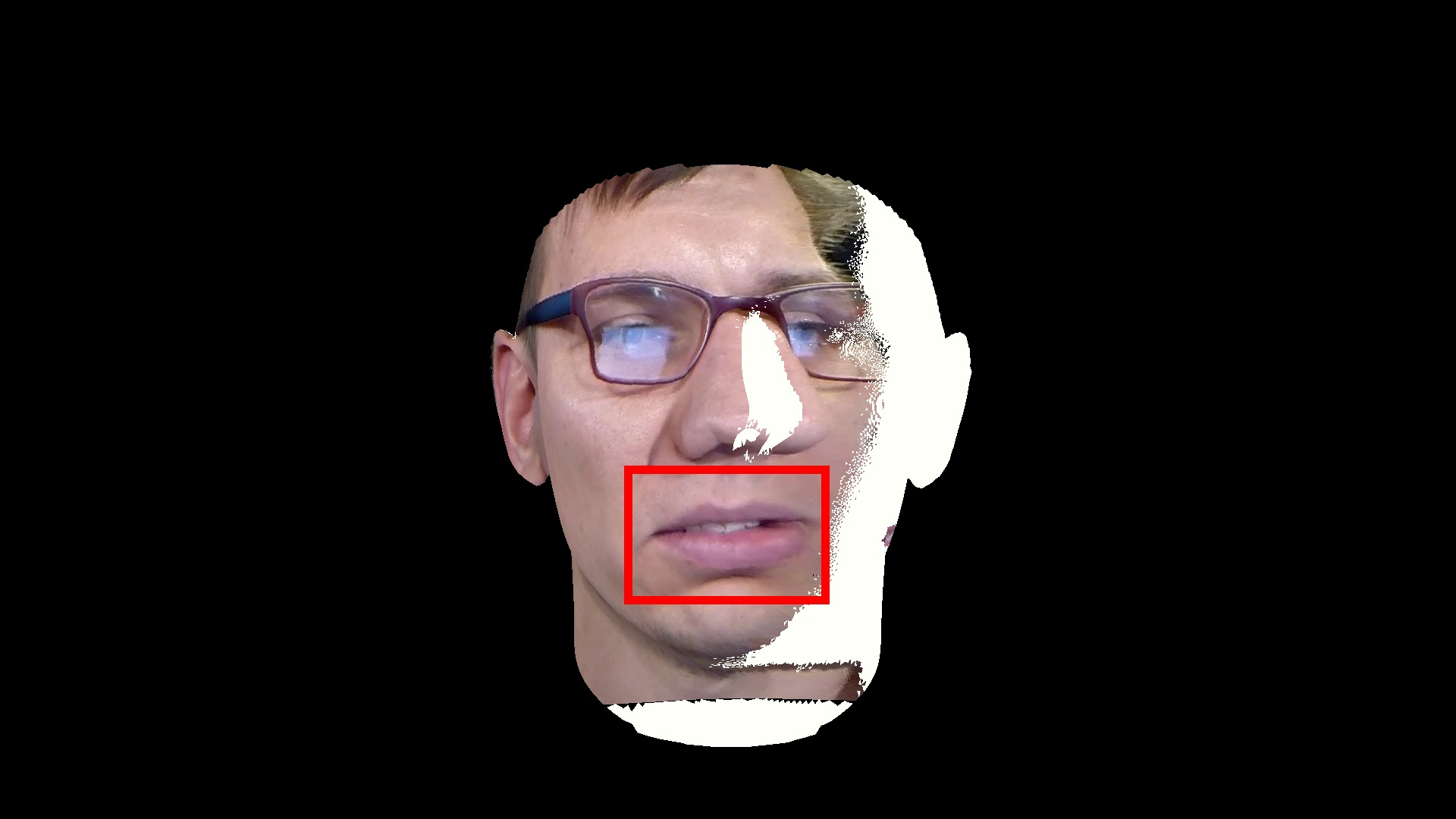}\hspace{1mm}
\includegraphics[trim = 20cm 6cm 20cm 8cm,clip,keepaspectratio=true,width=0.24\columnwidth]{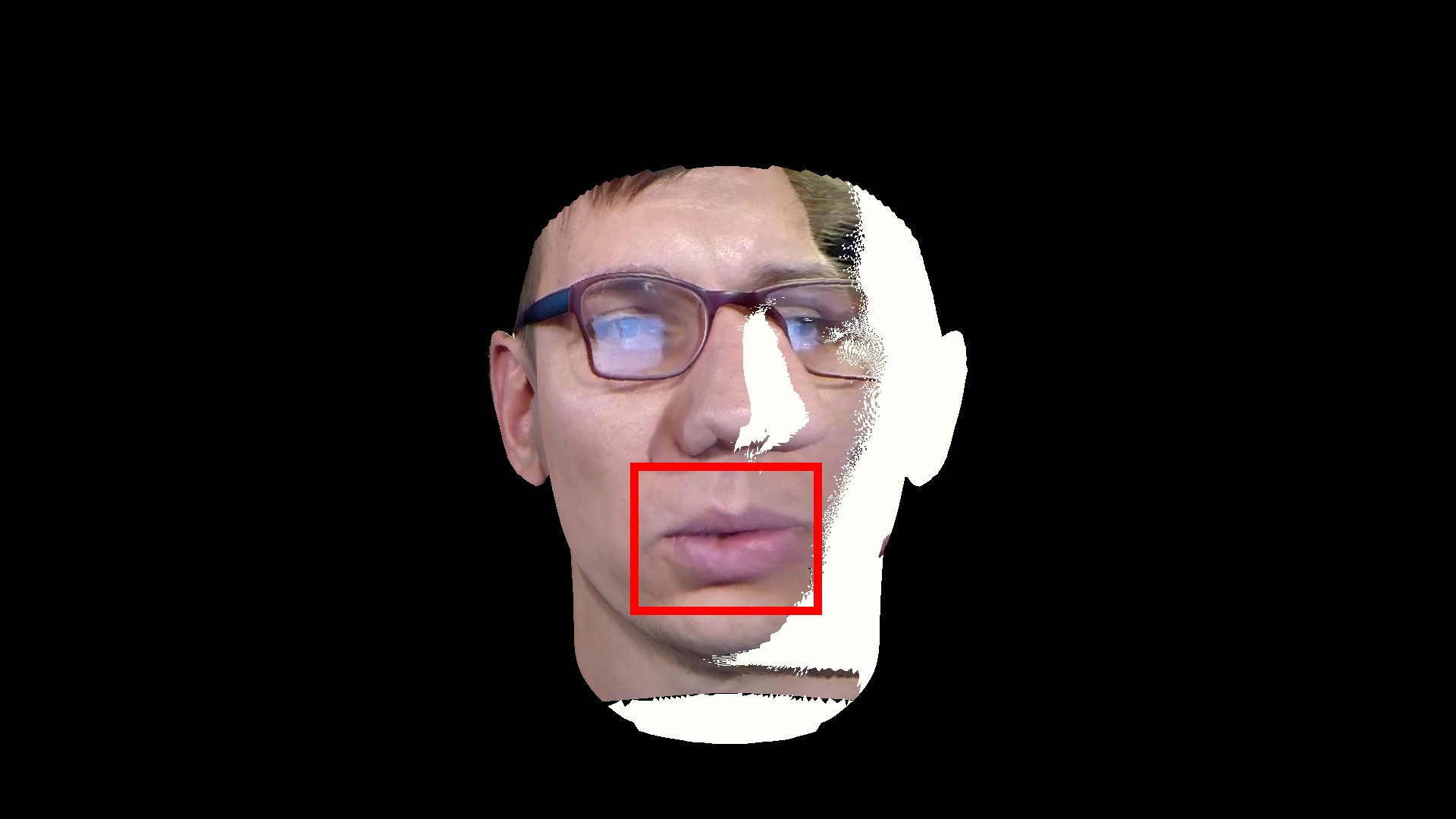}\hspace{1mm}
\includegraphics[trim = 20cm 6cm 20cm 8cm,clip,keepaspectratio=true,width=0.24\columnwidth]{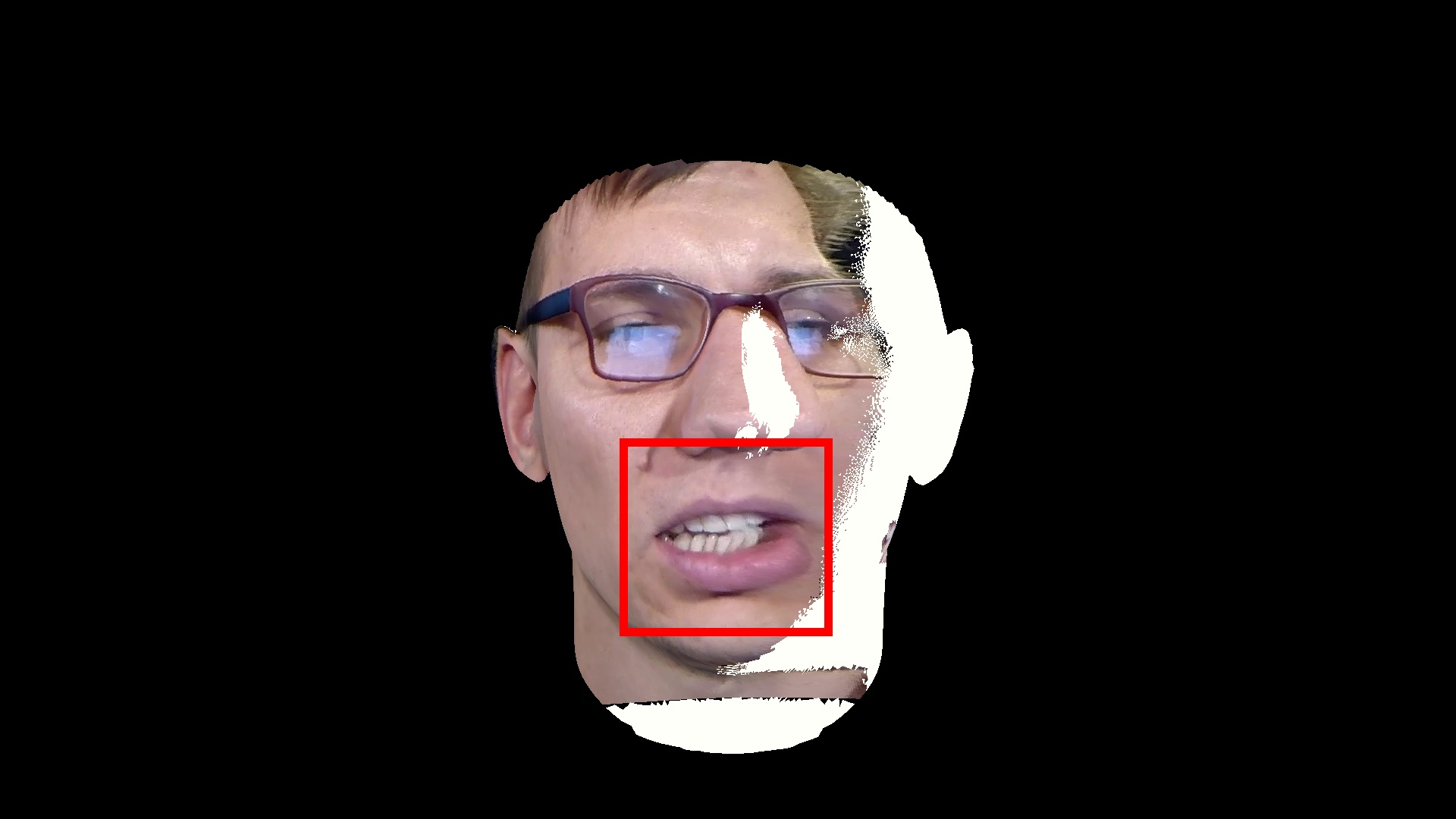}
}\\
\vspace{-3mm}
\subfloat[\cite{hassner2015effective}]{
\includegraphics[width=0.24\columnwidth]{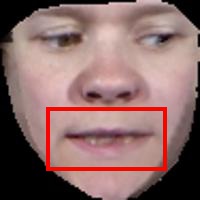}\hspace{1mm}
\includegraphics[width=0.24\columnwidth]{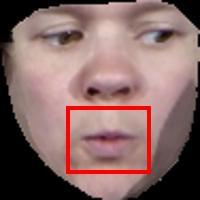}\hspace{1mm}
\includegraphics[width=0.24\columnwidth]{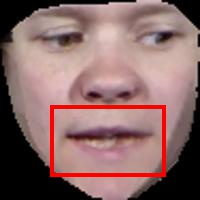}\hspace{1mm}
\includegraphics[width=0.24\columnwidth]{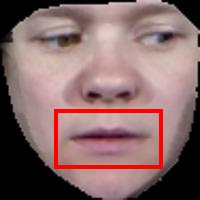} \hspace{1mm}
\includegraphics[width=0.24\columnwidth]{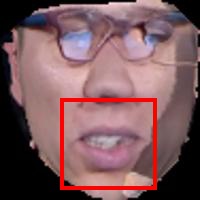}\hspace{1mm}
\includegraphics[width=0.24\columnwidth]{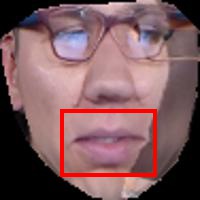}\hspace{1mm}
\includegraphics[width=0.24\columnwidth]{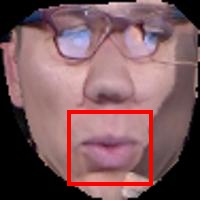}\hspace{1mm}
\includegraphics[width=0.24\columnwidth]{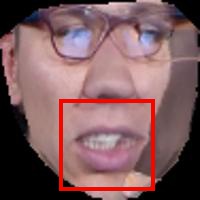}
}\\
\vspace{-3mm}
\subfloat[\cite{banerjee2018frontalize}]{
\includegraphics[width=0.24\columnwidth]{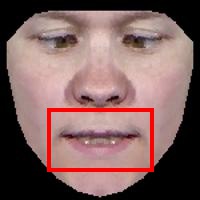}\hspace{1mm}
\includegraphics[width=0.24\columnwidth]{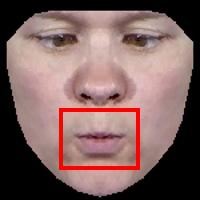}\hspace{1mm}
\includegraphics[width=0.24\columnwidth]{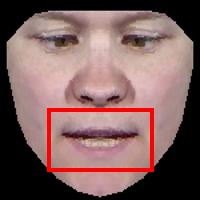}\hspace{1mm}
\includegraphics[width=0.24\columnwidth]{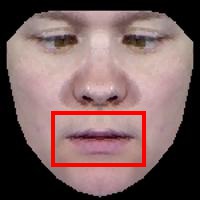}\hspace{1mm}
\includegraphics[width=0.24\columnwidth]{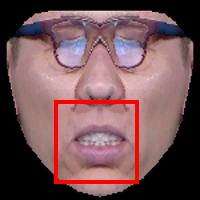}\hspace{1mm}
\includegraphics[width=0.24\columnwidth]{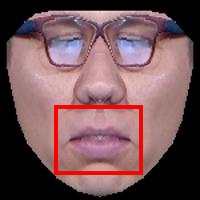}\hspace{1mm}
\includegraphics[width=0.24\columnwidth]{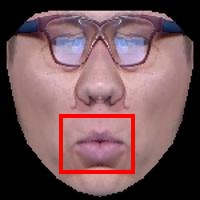}\hspace{1mm}
\includegraphics[width=0.24\columnwidth]{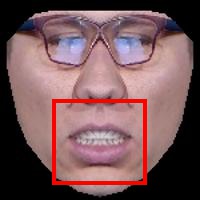}
}\\
\vspace{-3mm}
\subfloat[\cite{yin2020dual}]{
\includegraphics[width=0.24\columnwidth]{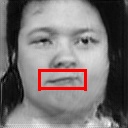}\hspace{1mm}
\includegraphics[width=0.24\columnwidth]{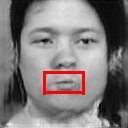}\hspace{1mm}
\includegraphics[width=0.24\columnwidth]{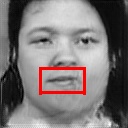}\hspace{1mm}
\includegraphics[width=0.24\columnwidth]{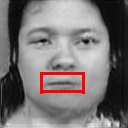}\hspace{1mm}
\includegraphics[width=0.24\columnwidth]{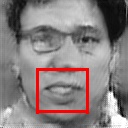}\hspace{1mm}
\includegraphics[width=0.24\columnwidth]{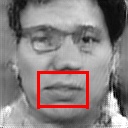}\hspace{1mm}
\includegraphics[width=0.24\columnwidth]{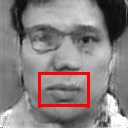}\hspace{1mm}
\includegraphics[width=0.24\columnwidth]{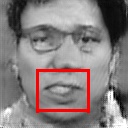}
}\\
\vspace{-3mm}
\subfloat[\cite{zhou2020rotate}]{
\includegraphics[width=0.24\columnwidth]{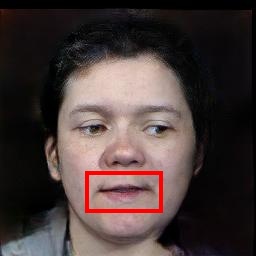}\hspace{1mm}
\includegraphics[width=0.24\columnwidth]{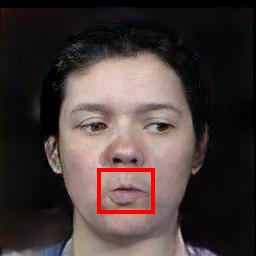}\hspace{1mm}
\includegraphics[width=0.24\columnwidth]{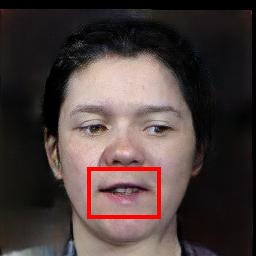}\hspace{1mm}
\includegraphics[width=0.24\columnwidth]{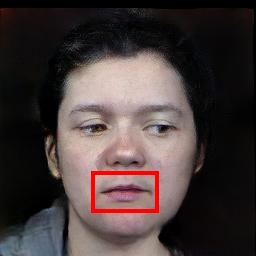}\hspace{1mm}
\includegraphics[width=0.24\columnwidth]{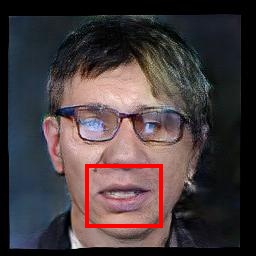}\hspace{1mm}
\includegraphics[width=0.24\columnwidth]{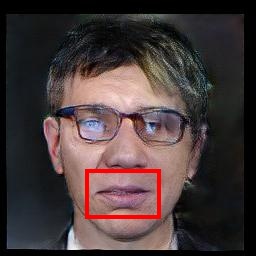}\hspace{1mm}
\includegraphics[width=0.24\columnwidth]{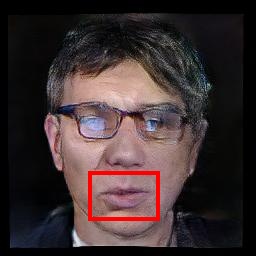}\hspace{1mm}
\includegraphics[width=0.24\columnwidth]{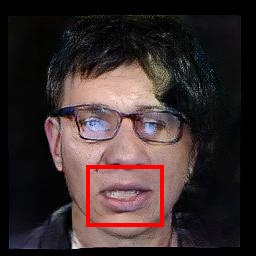}
}\\
\end{center}
\vspace{-4.5mm}
    \caption{Frontalization examples for participants \#02 (left) and \#21 (right) from the OuluVS2 dataset. The ZNCC scores correspond to the mouth bounding boxes shown in red. The estimated horizontal head orientation (yaw angle) is $24.9^\circ$ and $40.6^\circ$ for participant \#2 and \#21, respectively.}
    \label{fig:Oulu2}
\end{figure*}    

In order to evaluate the performance of the proposed frontalization method and to compare it with state-of-the-art methods, we used a publicly available dataset, namely the OuluVS2 dataset \cite{anina2015ouluvs2}. This dataset targets the understanding of speech perception, more precisely, the analysis of non-rigid lip motions that are associated with speech production. The dataset was recorded in an office with ordinary (artificial and natural) lighting conditions.  The recording setup consists of five synchronized cameras (2~MP, 30~FPS) placed at different points of view and with different orientations: $0^{\circ}$, $30^{\circ}$, $45^{\circ}$, $60^{\circ}$, $90^{\circ}$. 

The dataset contains $5 \times 106$ videos recorded with 53 participants. Each participant was instructed to read loudly several text sequences displayed on a computer monitor placed slightly to the left and behind the $0^{\circ}$ (frontal) camera. The displayed text consists of digit sequences, e.g. ``one, seven, three, zero, two, nine", of phrases, e.g. ``thank you", ``have a good time", and ``you are welcome", as well as of sequences from the TIMIT dataset, e.g. ``agricultural products are unevenly distributed". 
\addnote[pose-match]{1}{
While participants were asked to keep their heads still, natural uncontrolled head movements and body position changes were inevitable. As a consequence the actual head pose varies from one participant to another and there is no exact match between the head and camera orientations. 
}

  \begin{table}[b!]
    \begin{center}
    \begin{tabular}{ |l|l|c|  }
     \hline
     Method & Principle & ZNCC\\
     \hline
      \citeauthor{hassner2015effective}  &   2D-to-3D fitting + symmetry & 0.771\\
      \citeauthor{banerjee2018frontalize} & 2D-to-3D fitting + symmetry & 0.749\\
      \citeauthor{zhou2020rotate} & 2D-to-3D fitting + GAN& 0.793  \\
      \citeauthor{yin2020dual} & 2D-to-2D mapping using GAN &0.769 \\
       \citeauthor{ma2021towards} & 2D-to-2D affine fitting & 0.760\\
     \citeauthor{kang2021robust} & 3D-to-3D robust fitting &  0.831\\
     Proposed & 3D-to-3D robust/dynamic inference & \textbf{0.839} \\
     \hline
    \end{tabular}
       \end{center}
        \caption{ \label{table:zncc} Mean ZNCC scores for all 53 participants of the OuluVS2 dataset. ZNCC lies in the interval $[0,1]$.}
    \end{table}
    
\begin{table*}[h]
\begin{center}
\begin{tabular}{|c|c|c|c|c|c|c|c|}
 \hline
 Part. & Yaw  & \citeauthor{hassner2015effective}  & \citeauthor{banerjee2018frontalize} &  \citeauthor{zhou2020rotate} &  \citeauthor{yin2020dual} & \citeauthor{kang2021robust} & Proposed \\
 \hline
 \#31  & 19.1 & 0.905 & 0.856 & 0.822 & 0.875 & \textbf{0.927} & \textit{\textbf{0.925}}\\
 \#01   & 23.5   & 0.915 & 0.893&  0.884& \textbf{0.921} & 0.909 & \textit{\textbf{0.918}}\\
 \#02  & 24.9   & 0.888& 0.878 & 0.929 & 0.881 & \textbf{0.956} & \textit{\textbf{0.952}}\\
 \#10 & 29.0 & 0.805 & 0.812 &  \textbf{0.873} & 0.792 & 0.812 & \textit{\textbf{0.829}} \\
 \#23 & 30.0 & 0.810 & \textbf{0.857} &  0.819& 0.817 & \textit{\textbf{0.847}} & 0.843 \\
 \#27 & 32.9 & 0.685 & \textbf{0.852} &  \textit{\textbf{0.824}}& 0.772& 0.787& 0.805 \\
 \#19 & 37.8 &\textit{\textbf{0.752}} & 0.650 & 0.662 & 0.677& \textbf{0.755} &  \textbf{0.755} \\
 \#12 & 38.5& 0.731 & 0.713 & 0.755 & 0.683 &  \textbf{0.770} & \textit{\textbf{0.766}}\\
 \#21 & 40.6 & 0.632 & 0.743  & 0.653 & 0.673 & \textbf{0.766} & \textit{\textbf{0.751}} \\
 \hline
  Mean & & 0.791 & 0.801& 0.802  & 0.787 & \textit{\textbf{0.836}} & \textbf{0.838}  \\
\hline
\end{tabular}
\end{center}
\caption{\label{table:yawangles} ZNCC scores for nine participants as a function of estimated yaw angle (in degrees) that corresponds to the horizontal head orientation computed with the proposed 3D head-pose estimator. For each participant, the best scores are in \textbf{bold} and the second best are in \textit{\textbf{slanted bold}}.}
\end{table*}

In practice, we evaluated the performance of the proposed method and we compared it with four state-of-the-art methods for which the code is publicly available, \cite{hassner2015effective,banerjee2018frontalize,zhou2020rotate,yin2020dual}. We applied the frontalization to images extracted from the videos recorded with the $30^{\circ}$ camera ($I_p$) and compared the results with the ``ground-truth", namely the corresponding images extracted from the videos recorded with the $0^{\circ}$ camera ($I_t$). Notice that videos recorded with higher viewing angles, i.e. $45^{\circ}$,  $60^{\circ}$ and $90^{\circ}$, can be hardly exploited by a frontalization algorithm because half of the face is occluded. For each frontalized image $I_f$ we extract the mouth region $R_f$ and we search in the associated ground-truth image $I_t$ for the best-matching region  $R_t$. This provides a ZNCC score \eqref{eq:zncc} for each query image $I_p$. 
 \addnote[scale-frontal]{1}{
Notice that \eqref{eq:zncc} only cares about the horizontal and vertical shifts in the image plane and assumes that the frontalized face and the corresponding ground-truth frontal face share the same scale. In practice, different frontalization algorithms output faces at different scales. For this reason and for the sake of fairness, prior to applying  \eqref{eq:zncc}, we extract facial landmarks from both the frontalized and ground-truth faces and we use a subset of this set of landmarks to estimate the scale factor between the two faces. We do this for all the frontalization methods used in the comparison.}

We used the 106 video pairs recorded with the $30^\circ$ and $0^\circ$ cameras, respectively, associated with the 53 participants of the OuluVS2 dataset.
Each video contains 160 images, hence there are $106 \times 160 = 16,900$ image pairs in our benchmark. The mean ZNCC scores obtained with four methods, with  \cite{kang2021robust}, and with the proposed extension are shown in Table~\ref{table:zncc}. 
We noticed that there were important discrepancies in method performance across participants. In order to better understand this phenomenon, we computed the mean ZNCC scores for nine participants and displayed these means as a function of the yaw angle, i.e. horizontal head orientation estimated with the proposed method, Table~\ref{table:yawangles}. \addnote[yaw-diffs]{1}{One may notice that there is a wide range of yaw angles, from $19^\circ$ to $40^\circ$, and that the performance gracefully decreases as the yaw angle increases}. The proposed method yields results that are more consistent than the other methods, as the yaw angle increases.

\addnote[observed-shape]{2}{
The best performing methods are \cite{kang2021robust} and its dynamic extension. One remarks that the improvement of the dynamic model over \cite{kang2021robust} is minor, and this for the following reason. The dynamic \ac{FF} uses 68 observed landmarks in order to update the deformable model. However the latter is composed of thousands of vertices: consequently, the vast majority of these vertices are not observed. This means that the innovation term in \eqref{eq:mean-rec} affects a handful of the shape's vertices. 
}


Examples of face frontalization obtained with our method and with four other methods, \cite{hassner2015effective,banerjee2018frontalize,zhou2020rotate,yin2020dual}, are shown on Figure~\ref{fig:Oulu2}: (a) input mages recorded with the $30^\circ$ camera, (b) ground-truth images recorded with the $0^\circ$ camera, (c)-(g) frontalization results.
The ZNCC correlation scores correspond to the mouth region, shown in red.  As already mentioned, both \cite{hassner2015effective} and \cite{banerjee2018frontalize} enforce facial symmetry as a post-processing frontalization step to compensate for the gaps caused by self occlusions. 
It is interesting to note that the more recent GAN-based methods, \cite{zhou2020rotate,yin2020dual}, yield results comparable with the traditional computer vision methods. 


\section{Lip reading benchmark}
\label{sec:lip-reading}

\begin{table*}[t!]
\begin{center}
\begin{tabular}{| p{5cm} | c | c | c | c | c |}
\hline
\diagbox[innerwidth=5cm]{Training}{Testing} & \citeauthor{hassner2015effective}  & \citeauthor{zhou2020rotate} & \citeauthor{yin2020dual} & \citeauthor{ma2021towards} & Proposed \\
\hline
Training with \cite{ma2021towards} & 60 &59 &20 &87 & 82\\
\hline
Fine tuning with \cite{zhou2020rotate} & 60 & 72 & 20 & 84 & 80 \\
\hline
Fine tuning with proposed & 64 & 66 & 24 & 88 &85 \\
\hline
\end{tabular}
\caption{\label{table:lipreading} 
Word classification scores (WCSs) in \%. \textit{First row:} The lip-reading model is trained with the built-in \ac{FF} of \cite{ma2021towards}; \textit{Second row:} the lip-reading model is fine tuned with the \ac{FF} of \cite{zhou2020rotate}; \textit{Third row:} The lip-reading model is fine tuned with the proposed \ac{FF} method. For testing, we preprocessed the test images with the \ac{FF} methods included in the comparison. 
}
\end{center}
\end{table*}

We also evaluated the ability of our method to improve the performance of lip reading and we compared it with other methods. For this purpose, we considered an isolated word recognition (IWR) task. The LRW (lip reading in the wild) dataset \cite{chung2016lip} consists of $500,000$ videos of $500$ English words uttered by $1,000$ different speakers. Each video contains 29 frames and each target word is surrounded by context words. There are large inter-speaker variations in terms of head motions. To date, the best performing method for this 500-IWR task is based on the temporal convolutional network (TCN) model of \cite{martinez2020lipreading,ma2021towards,ma2020lip} which achieves a word classification score (WCS) of 87\%. This lip-reading model and its variants use their own \ac{FF} method which estimates a 3D affine mapping between the input face and a generic face model, \cite{martinez2020lipreading}. Their \ac{FF} is used as a preprocessing stage for training, validation and test. The authors don't provide a detailed description of the frontalization method that they use. 

\addnote[lip-read-details]{1}
{We performed the following 500-IWR experiments. In the first experiment we used the lip-reading model provided by the publicly available software packages of \cite{ma2021towards}. This model is trained with their \ac{FF}.  In the second experiment we preprocessed a subset of the training dataset with the proposed dynamic \ac{FF} method and we fine tuned the lip-reading model of \cite{ma2021towards} on the 500-IWR task. For the purpose of fine tuning, for each one of the 500 words, we used 200 videos for training and 20 videos for validation, hence 100,000 training videos and 10,000 validation videos. Finally, we repeated the second experiment using \cite{zhou2020rotate} for \ac{FF}. In order to test these three models, we used the entire test dataset of LRW, namely 20 test videos for each one of the 500 words. The test videos were then preprocessed with each one of the \ac{FF} models included in the benchmark: Table~\ref{table:lipreading} shows the results obtained with  $3\times 5$ configurations corresponding to different train/test combinations. The proposed/\citeauthor{ma2021towards} combination yields the best results:  for this train/test combination, the WCS score is slightly increased, from 87 to 88, while the \citeauthor{zhou2020rotate}/\citeauthor{ma2021towards} combination decreases the WCS score from 87 to 84.
}

\addnote[catastrophe]{1}
{The proposed frontalization model preserves facial expressions, hence the statistical properties of the training dataset are preserved. On the contrary, the GAN-based method of \cite{zhou2020rotate} doesn't enjoy this Euclidean invariance. Consequently, the statistical distribution of the data used for fine tuning is modified. The model tends to overfit to the new distribution thus leading to a performance drop, as if the finely tuned model forgets what it was learned before. This phenomenon is referred to as catastrophic learning \cite{mccloskey1989catastrophic}, and is extensively investigated in continual learning.
}

\section{Audio-visual speech enhancement}
\label{sec:avse}

In this section we report experiments with using the proposed method in conjunction with \ac{AVSE}. We start by summarizing the \ac{AVSE} method based on a conditional \ac{VAE} model~\cite{sadeghi2020audio}, which we denote AV-CVAE. The whole framework consists of two steps: training and testing (inference). At training, a prior distribution of clean speech is learned from the concatenation of a clean audio signal with an embedding of the associated lip images. At inference, clean speech is extracted from a noisy-speech signal and from a sequence of lip images: the learned prior distribution is combined with a noise model, whose parameters together with the parameters of the clean speech that were previously learned, are estimated following a variational expectation-maxi\-mi\-zation (VEM) procedure.

\begin{figure*}[t!]
    \centering
    \includegraphics[width=0.9\linewidth]{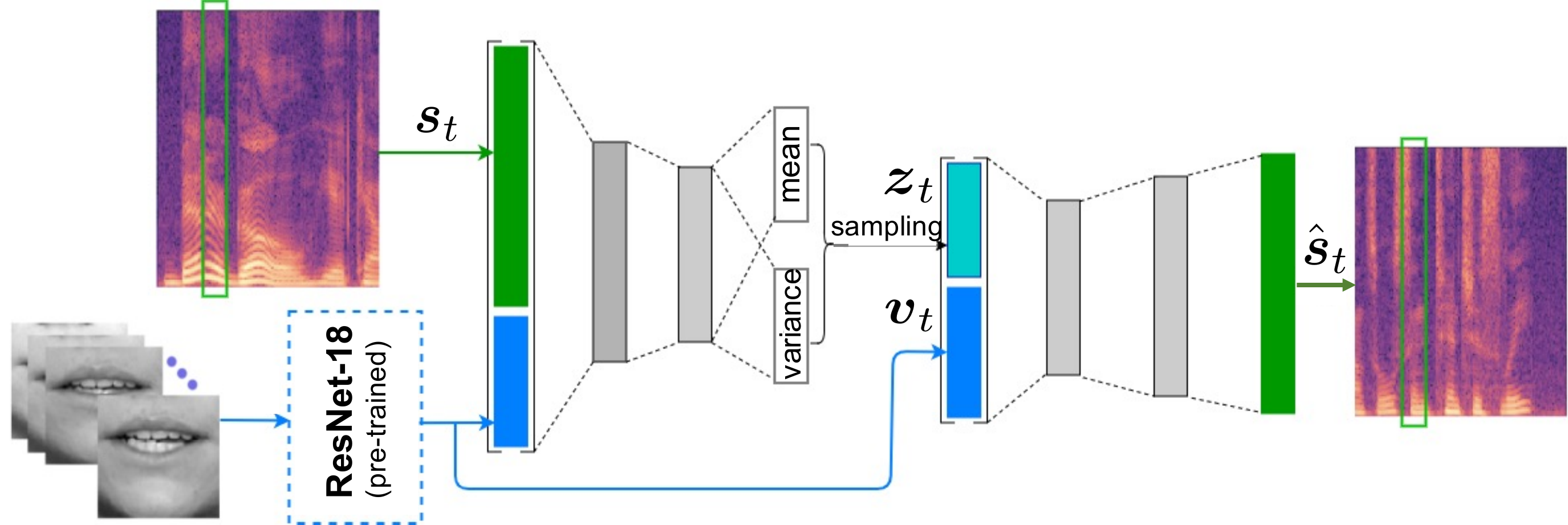}
    \caption{AV-CVAE and ResNet-AV-CVAE architectures used in our speech enhancement experiments.}\label{fig:av-cvae}
  \end{figure*}

Given a dataset of complex-valued short-time Fourier transform (STFT) frames of a clean-speech signal, denoted $\svect_{t}\in\mathbb{C}^F$, and the corresponding lip embedding obtained from a lip bounding-box cropped from the image of a speaker face, denoted $\vvect_t \in \mathbb{R}^M$, a latent-variable generative model is trained using the VAE framework. This involves defining a parametric distribution for the likelihood $p_{\Theta}(\svect_{t}|\zvect_t, \vvect_t)$, and a parametric prior distribution for the latent code $\zvect_{t}\in\mathbb{R}^L$, $L\ll F$, $p_{\Gamma}(\zvect_t|\vvect_t)$. These distributions are implemented by some deep neural network architectures, whose parameters, \{$\Theta, \Gamma$\}, are learned following an amortized variational inference \cite{KingW14}, where an encoder network is introduced to approximate the intractable posterior distribution of the latent codes. Fig.~\ref{fig:av-cvae} illustrates the AV-CVAE architecture. The main difference between this architecture and the one proposed in \cite{sadeghi2020audio,sadeghi2021mixture} is the presence of a ResNet backbone from a pretrained model specialized for lip reading~\cite{martinez2020lipreading}.

With the parametric prior distribution for clean speech being learned, we consider an observation model as $\ovect_{t} = \svect_{t}+\bvect_{t}$, in which $\ovect_{t}\in\mathbb{C}^F$ and  $\bvect_{t}\in\mathbb{C}^F$ denote observed speech and noise, respectively. Considering an NMF-based model for noise, and combining it with the speech model, the set of NMF parameters are then learned by a variational inference procedure. Once learned, the clean speech estimate $\hat{\svect}_{t}$ is obtained via a probabilistic Wiener filtering. More details can be found in \cite{sadeghi2020audio}.

\begin{table*}[h]
\centering
	\resizebox{\textwidth}{!}{
\begin{tabular}{|l|c|c|c|c|c||c|c|c|c|c||c|c|c|c|c|}
\hline
  Measure & \multicolumn{5}{c||}{STOI $[0,1]\uparrow$} & \multicolumn{5}{c||}{PESQ $[-0.5,4.5]\uparrow$} & \multicolumn{5}{c|}{SI-SDR (dB) $\uparrow$} \\
\hline
{SNR (dB)} & {-10} & {-5} & {0} & {5} & {10} & {-10} & {-5} & {0} & {5} & {10} & {-10} & {-5} & {0} & {5} & {10}  \\ \hline\hline
Noisy audio input & 0.40 & 0.53 & 0.66 & 0.78 & \textbf{0.86} & 0.90        & 1.24 & 1.67      & 2.05       & 2.42   & -15.92 & -10.62 & -5.44 & -0.40 & 4.60                   \\ \hline
A-VAE \citeauthor{Leglaive_MLSP18} & 0.41 & 0.56& 0.70 & \textbf{0.79} & 0.85 & 0.93         & 1.51        & 2.02       & 2.43       & 2.73 & -7.01 & -0.29 & 5.08 & 9.41 & 12.74  \\ \hline
AV-CVAE \citeauthor{sadeghi2020audio}  & 0.42 & 0.57 & 0.69 & \textbf{0.79} & 0.84  & 1.02 & 1.56 & 2.06 & 2.42 & 2.73  &      -6.96  & -0.04                  & 5.01                     &  9.06                    & 12.25 \\ \hline
Res-AV-CVAE-W/O-FF & 0.41 & 0.55 & 0.67 & 0.77 & 0.83 &   1.02     &   1.53                 &  1.99                    &  2.35                    &  2.70 & -7.84 &  -0.60 & 4.68 & 8.81 & 12.30 \\ \hline
Res-AV-CVAE-DA-ST-GAN \citeauthor{zhou2020rotate}  & 0.40 &  0.55 & 0.68 & 0.78 & 0.84 & 1.01  & 1.54  & 2.01 & 2.39 & 2.72 &     -7.92   &   -1.14                & 4.13                    &        9.27              & 11.77\\ \hline
Res-AV-CVAE-DA-GAN \citeauthor{yin2020dual}  & 0.39  & 0.55 & 0.66 & 0.68 & 0.72 & 0.76  &  1.42 & 1.87 & 1.66 & 1.96 & -9.08        &  -0.45                  & 3.88                     &  4.55            & 5.23 \\ \hline
Res-AV-CVAE-RFF \citeauthor{kang2021robust} & \textbf{0.43} & 0.58 & 0.71 & \textbf{0.79} & 0.85 & 1.12 & 1.69 & 2.13  & \textbf{2.48}   & \textbf{2.77}  &\textbf{-6.30} & 0.10 & 5.24& 9.30 & 12.60 \\ \hline
Res-AV-CVAE-DFF & \textbf{0.43} & \textbf{0.60} &  \textbf{0.73} & \textbf{0.79} & 0.85 & \textbf{1.13}  & \textbf{1.71}  & \textbf{2.20}  & \textbf{2.48}   & \textbf{2.77}  &
 -6.35 &  \textbf{0.28} & \textbf{5.87} & \textbf{9.42} & \textbf{12.77} \\ \hline
\end{tabular}}
\caption{\label{tab:measures} Average STOI, PESQ, SI-SDR values.}
\end{table*}

All the experiments reported below use the
MEAD dataset~\cite{wang2020mead} which contains short videos of talking faces with large-scale facial expressions. For all 46 publicly available participants, there are recordings of eight different emotions at three different intensity levels and seven camera viewpoints. Many participants have natural head motions, which challenges state-of-the-art AVSE. Among all videos, we select the videos of all emotion categories taken at the frontal view and at the level 3 (the highest) of emotion intensity. These high-intensity emotions are associated with large head movements and exaggerated lip motions, thus allowing to assess the effect of head movements on the performance of speech enhancement. In total, there are around 5 hours of videos for training, 0.7 hours for validation and 0.7 hours for testing.

We process the input videos with four different \ac{FF} methods in order to compare their effectiveness of removing head movements and hence of improving the quality of the speech output: the GAN based methods \cite{zhou2020rotate} and \cite{yin2020dual}, denoted ST-GAN and DA-GAN, respectively, the method of  \cite{kang2021robust} that corresponds to Algorithm~\ref{algo:rff}, denoted RFF, and the dynamic method that corresponds to Algorithm~\ref{algo:trff}, denoted DFF.
Additionally, we consider the case of directly using the raw input without any form of face frontalization, denoted W/O-FF.
For all these cases we crop the lip region, yielding 67$\times$67 images, which are then converted to gray scale and normalized to facilitate the downstream processing.

We consider three speech enhancement pipelines, all based on VAEs. The Audio-only VAE (A-VAE), \cite{Leglaive_MLSP18} has an encoder and a decoder composed of fully-connected layers. The extracted audio feature vector is of size $F=513$ whereas the latent space is of size $L = 32$. The  AV-CVAE model \cite{sadeghi2020audio} shares a similar encoder-decoder architecture as A-VAE, with the additional fully-connected layers to encode the visual information. Furthermore, we propose to use a ResNet backbone specially trained for lip reading~\cite{martinez2020lipreading} (shown in a dashed box in Figure~\ref{fig:av-cvae}) for visual feature extraction. The backbone follows the standard design of ResNet-18~\cite{he2016deep} except for the first convolutional layer, which is replaced by a 3D convolutional layer to incorporate temporal information from neighbouring frames. This variant is denoted as Res-AV-CVAE. In practice, the dimension of the visual embedding is $M = 128$.

All the VAEs are trained in an end-to-end manner. A-VAE is trained on audio data. AV-CVAE \cite{sadeghi2020audio} is fine-tuned with the MEAD dataset, whereas Res-AV-CVAE is trained from scratch using MEAD. Note that the ResNet backbone is frozen without requiring the gradients. Hence, it is a static feature extractor. We set $5e^{-5}$ as the learning rate for the fine-tuning model and $1e^{-4}$ for the training from scratch. The Adam optimizer was used with a batch size of 128. We also applied early stopping with a patience of 10 epochs. Note that we trained and tested one model with one specific lip preprocessing method at a time. 
At test time, noise from the DEMAND dataset~\cite{thiemann2013demand} is combined with the clean speech to construct the audio input. There are five noise levels for each type of noise, namely $-10$~dB, $-5$~dB, 0~dB, 5~dB and 10~dB. Three standard speech enhancement metrics are used for quantitative evaluation: the \textit{scale-invariant signal-to-distortion ratio} (SI-SDR)~\cite{le2019sdr}, the \textit{short-time objective intelligibility} (STOI)~\cite{taal2011algorithm} and the \textit{perceptual evaluation of speech quality} (PESQ)~\cite{rix2001perceptual}. SI-SDR is measured in decibels (dB), while STOI and PESQ values are in the range $\left[0,1\right]$ and $\left[-0.5,4.5\right]$, respectively (the higher the better).

\begin{figure}[t]
    \centering
    \includegraphics[width=\linewidth]{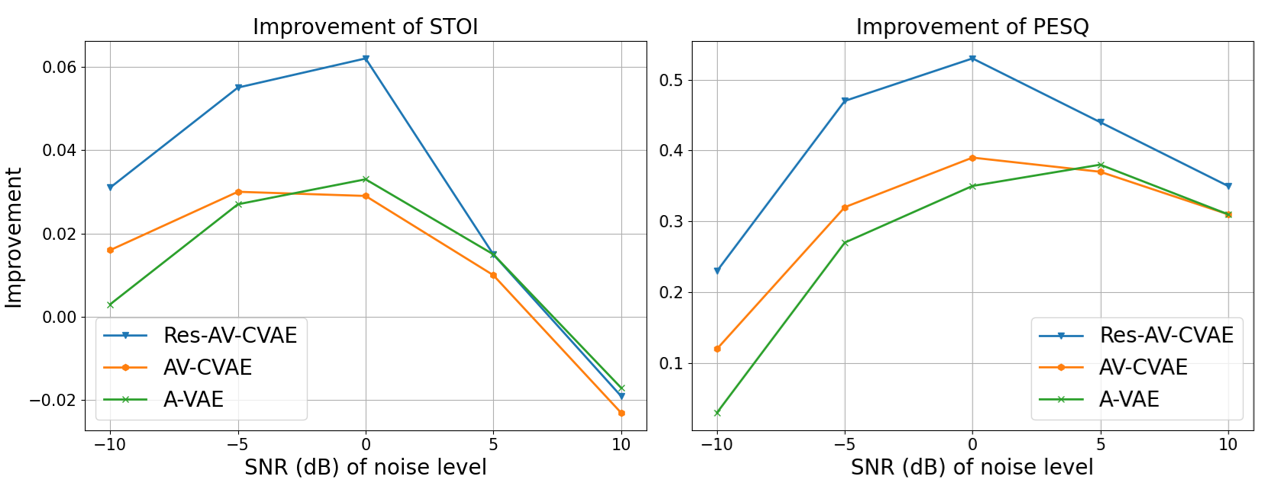}
    
    \caption{Performance comparison of A-VAE, AV-CVAE and Res-AV-CVAE based on STOI (left) and PESQ (right). }\label{fig:vae_compare}
    
 \end{figure}

We start with evaluating the impact of different frontalization methods on AVSE performance, i.e. Table~\ref{tab:measures}, where the average scores for different levels of noise (SNR) are presented. Selecting RFF and DFF -- the best-performing methods -- as examples, we remark that the difference between with and without frontalization is significant. This confirms that the head motions interfere the processing of visual speech patterns. In other words, separating the rigid head movements from the non-rigid lip deformations allows the model to learn a better clean speech model. Moreover, the comparison between A-VAE and Res-AV-CVAE with RFF/DFF further validates the contribution of the visual modality. \addnote[avse-exp]{2}{In addition, DFF demonstrates a better performance than RFF, especially in terms of the speech intelligibility score STOI at high noise levels. However, one should note that the VAE models used in this paper do not incorporate the temporal dynamics of the audio and visual data, and rather process the data time frames independently. Using a dynamical VAE model would lead to even higher performance gain in DFF compared to RFF. }

The choice of the face frontalization method is important. While RFF/DFF offers significant improvements, ST-GAN  yields a minor difference compared to the case with head movements. Indeed, GAN-based image generation models have no theoretical guarantee for preserving the lip shape -- they add some form of visual noise, which neutralizes the gain of frontalization. This explanation is also supported by the results of DA-GAN: its performance is falling far behind the other methods. As the results of ST-GAN are conditioned on the transformation-based process, the model possesses a prior knowledge about the frontalized face. Moreover, the direct mapping from an arbitrary viewpoint to a frontal view of DA-GAN introduces even more dramatic modifications in the lip shape. Thus, the model has more difficulties to learn the correct speech patterns from lip movements. 

We then compare the performance of different VAE architectures in Figure~\ref{fig:vae_compare}, where the improvement of scores are shown as a function of different levels of noise (SNR). More precisely, the improvement refers to the difference between the score obtained by using the raw noisy speech and those obtained by using the enhanced speech. First, it is remarkable to see that Res-AV-CVAE significantly outperforms AV-CVAE, showing the gain of using a more powerful feature extractor. Second, we observe that with a noise level in the range $[-5, 0]$~dB, the Res-AV-CVAE model reaches an optimal stage (a peak in the curve) for fusing the audio-visual data. That is, with the noise level going higher (smaller SNR), the audio would be too corrupted to be enhanced and the visual contribution is significant. In contrast, with the noise level going lower (higher SNR), the importance of the visual data is decreasing and the already clean speech becomes harder to be enhanced. While the superior performance of the Res-AV-CVAE models is more significant at high noise levels, it is quite remarkable to observe that Res-AV-CVAE-RFF performs almost equally well as A-VAE for low noise levels. These experiments confirm the 
complementary roles of the visual and audio modalities for the task of speech enhancement.

To give an insight on the impact of removing head movements, Figure~\ref{fig:displacement} shows the horizontal and vertical displacements of a landmark located on the upper lip. Both the vertical and horizontal trajectories of this lip landmark are strongly affected by head motions. In the light of this experiment, one may interpret the process of separating rigid head movements and non-rigid lip movements, as a way of extracting clean visual-speech information from the raw videos.

\section{Conclusions}
\label{sec:conclusions}

\addnote[shape]{1}{
Shape as defined by \cite{kendall1989survey}, is the geometric information that remains once an object has been normalized with
respect to rotation, scaling and translation. 
}
The proposed face frontalization methodology follows this definition and hence it guarantees that face geometric information, i.e. non-rigid facial deformations, is preserved.
This stays in contrast with state-of-the-art DNN-based frontalization methods that learn millions of parameters without the theoretical guarantee that they faithfully preserve facial deformations.


We conducted several experiments in order to 
 analyze the effect of frontalization onto visual speech processing, whose success critically relies on the analysis of non-rigid mouth motions, e.g. lip reading. For this purpose, we used three datasets, OuluVS2, LRW, and MEAD.
 
We proposed an evaluation pipeline that consists of measuring the ZNCC score between a frontalized face and a frontal view of the same face. We compared our method with four state-of-the-art methods that use various geometric and DNN models. This benchmark reveals that the proposed method performs better than the other methods in preserving the shape of the mouth. 
 
The LRW and MEAD datasets contain videos of persons uttering speech. Unlike the OuluVS2 participants, who keep their heads in a fixed position and orientation, the LRW and MEAD participants perform head motions -- a natural human behavior. We combined our frontalization method with two speech processing tasks, lip reading and speech enhancement, and we thoroughly analyzed its effect onto two scores: word classification and speech intelligibility. These experiments reveal that these scores are improved significantly with respect with both classical geometric models and GAN models.

It is interesting to remark that the proposed formulation may well be viewed as a method for separating rigid head motions from non-rigid facial expressions. This is useful, not only for improving the performance of visual speech, but for a number of other tasks that involve the analysis of facial expressions in realistic scenarios.


\bibliographystyle{spbasic}   

\end{document}